\documentclass[11pt]{article}

\usepackage[final]{acl}

\usepackage{times}
\usepackage{latexsym}

\usepackage[T1]{fontenc}

\usepackage[utf8]{inputenc}

\usepackage{microtype}

\usepackage{inconsolata}

\usepackage{graphicx}

\usepackage{calc}
\usepackage{microtype}
\usepackage{hyperref}
\usepackage{xurl}
\usepackage{booktabs}
\usepackage{xspace}
\usepackage{lineno}
\usepackage{graphicx}
\usepackage{rotating}
\usepackage{booktabs}
\usepackage{tabularx}
\usepackage{array}
\usepackage{geometry}
\usepackage{adjustbox}
\usepackage{longtable}
\usepackage{tcolorbox}
\usepackage{float}
\usepackage{afterpage}
\usepackage{multirow}
\usepackage{subfigure}
\usepackage{xfrac}   %
\usepackage[table]{xcolor}
\usepackage{colortbl}

\definecolor{SoftGreen}{HTML}{B5E7B0}
\definecolor{DarkGreen}{HTML}{006400}

\newcommand{\cellcolorvalue}[1]{%
    \ifdim #1 pt > 95 pt \cellcolor{SoftGreen!100}\else%
    \ifdim #1 pt > 90 pt \cellcolor{SoftGreen!90}\else%
    \ifdim #1 pt > 85 pt \cellcolor{SoftGreen!80}\else%
    \ifdim #1 pt > 80 pt \cellcolor{SoftGreen!70}\else%
    \ifdim #1 pt > 75 pt \cellcolor{SoftGreen!60}\else%
    \ifdim #1 pt > 70 pt \cellcolor{SoftGreen!50}\else%
    \ifdim #1 pt > 65 pt \cellcolor{SoftGreen!45}\else%
    \ifdim #1 pt > 60 pt \cellcolor{SoftGreen!40}\else%
    \ifdim #1 pt > 55 pt \cellcolor{SoftGreen!35}\else%
    \ifdim #1 pt > 50 pt \cellcolor{SoftGreen!30}\else%
    \ifdim #1 pt > 45 pt \cellcolor{SoftGreen!25}\else%
    \ifdim #1 pt > 40 pt \cellcolor{SoftGreen!20}\else%
    \ifdim #1 pt > 35 pt \cellcolor{SoftGreen!18}\else%
    \ifdim #1 pt > 30 pt \cellcolor{SoftGreen!15}\else%
    \ifdim #1 pt > 25 pt \cellcolor{SoftGreen!12}\else%
    \ifdim #1 pt > 20 pt \cellcolor{SoftGreen!10}\else%
    \ifdim #1 pt > 15 pt \cellcolor{SoftGreen!8}\else%
    \ifdim #1 pt > 10 pt \cellcolor{SoftGreen!6}\else%
    \ifdim #1 pt > 5 pt \cellcolor{SoftGreen!4}\else%
    \cellcolor{white}\fi\fi\fi\fi\fi\fi\fi\fi\fi\fi\fi\fi\fi\fi\fi\fi\fi\fi\fi%
    #1%
}

\definecolor{SoftRed}{HTML}{F9B7B0}
\definecolor{DarkRed}{HTML}{8B0000}

\newcommand{\cellcolorvaluered}[1]{%
    \ifdim #1 pt > 95 pt \cellcolor{SoftRed!100}\else%
    \ifdim #1 pt > 90 pt \cellcolor{SoftRed!90}\else%
    \ifdim #1 pt > 85 pt \cellcolor{SoftRed!80}\else%
    \ifdim #1 pt > 80 pt \cellcolor{SoftRed!70}\else%
    \ifdim #1 pt > 75 pt \cellcolor{SoftRed!60}\else%
    \ifdim #1 pt > 70 pt \cellcolor{SoftRed!50}\else%
    \ifdim #1 pt > 65 pt \cellcolor{SoftRed!45}\else%
    \ifdim #1 pt > 60 pt \cellcolor{SoftRed!40}\else%
    \ifdim #1 pt > 55 pt \cellcolor{SoftRed!35}\else%
    \ifdim #1 pt > 50 pt \cellcolor{SoftRed!30}\else%
    \ifdim #1 pt > 45 pt \cellcolor{SoftRed!25}\else%
    \ifdim #1 pt > 40 pt \cellcolor{SoftRed!20}\else%
    \ifdim #1 pt > 35 pt \cellcolor{SoftRed!18}\else%
    \ifdim #1 pt > 30 pt \cellcolor{SoftRed!15}\else%
    \ifdim #1 pt > 25 pt \cellcolor{SoftRed!12}\else%
    \ifdim #1 pt > 20 pt \cellcolor{SoftRed!10}\else%
    \ifdim #1 pt > 15 pt \cellcolor{SoftRed!8}\else%
    \ifdim #1 pt > 10 pt \cellcolor{SoftRed!6}\else%
    \ifdim #1 pt > 5 pt \cellcolor{SoftRed!4}\else%
    \cellcolor{white}\fi\fi\fi\fi\fi\fi\fi\fi\fi\fi\fi\fi\fi\fi\fi\fi\fi\fi\fi%
    #1%
}

\definecolor{comment-red}{rgb}{0.8,0,0}
\definecolor{lightgray}{gray}{0.7}

\newcommand{\Emoji}{\raisebox{-0.11em}{\includegraphics[height=1.0em]{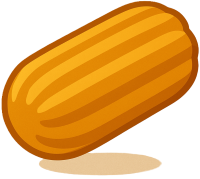}}}

\newcommand{\dataset}[0]{\textsc{Churro-DS}\xspace}

\newcommand{\model}[0]
{\textsc{Churro}\xspace}
\newcommand{\modelWithEmoji}{%
  \texorpdfstring{\Emoji~\model}{\model}%
}

\title{\modelWithEmoji: Making History Readable  \\ with an Open-Weight Large Vision-Language Model \\ for High-Accuracy, Low-Cost Historical Text Recognition}

\author{
  Sina J. Semnani \quad
  Han Zhang \quad
  Xinyan He \quad
  Merve Tekgürler \quad
Monica S. Lam \\
  Stanford University, Stanford, CA \\
  \texttt{\{sinaj, parisz, xinyanh, lam\}@cs.stanford.edu}, 
  \texttt{mtekgurl@stanford.edu}
}

\begin{document}
\maketitle

\begin{abstract}

Accurate text recognition for historical documents can greatly advance the study and preservation of cultural heritage. Existing vision-language models (VLMs), however, are designed for modern, standardized texts and are not equipped to read the diverse languages and scripts, irregular layouts, and frequent degradation found in historical materials.

This paper presents \modelWithEmoji, a 3B-parameter open-weight VLM specialized for historical text recognition. The model is trained on \dataset~\footnote{\model stands for \textbf{C}ollection of \textbf{H}istorical \textbf{U}nified \textbf{R}esources for \textbf{R}obust \textbf{O}CR.}, the largest historical text recognition dataset to date. \dataset unifies 155 historical corpora comprising 99,491 pages, spanning 22 centuries of textual heritage across 46 language clusters, including historical variants and dead languages.

We evaluate several open-weight and closed VLMs and optical character recognition (OCR) systems on \dataset and find that \model outperforms all other VLMs.
On the \dataset test set, \model achieves 82.3\% (printed) and 70.1\% (handwritten) normalized Levenshtein similarity, surpassing the second-best model, Gemini 2.5 Pro, by 1.4\% and 6.5\%, respectively, while being 15.5 times more cost-effective.

By releasing the model and dataset, we aim to enable community-driven research to improve the readability of historical texts and accelerate scholarship.~\footnote{Dataset, model, and code are available at \url{https://github.com/stanford-oval/Churro}.}

\end{abstract}

\section{Introduction}

\begin{figure}[th!]
    \centering
    \includegraphics[width=1.0\linewidth]{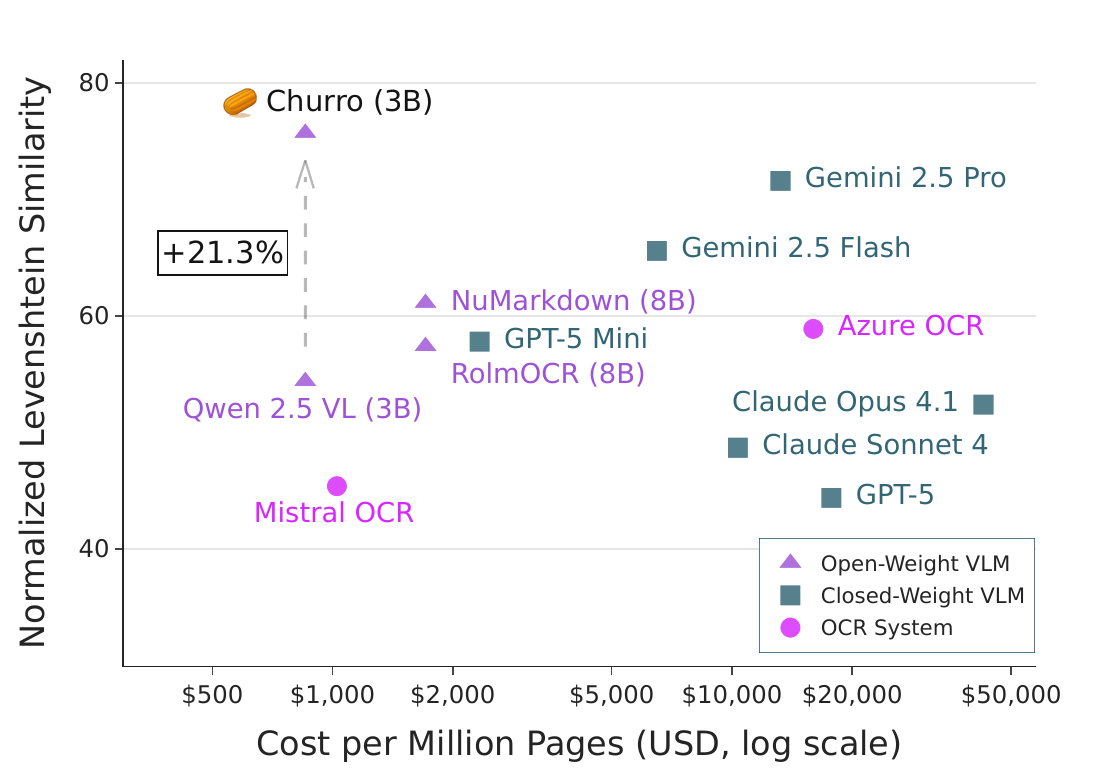}
     \caption{Summary of the performance–cost trade-off across different VLMs on historical documents, measured on the full \dataset test set.
    \model outperforms the 3B-parameter Qwen 2.5 VL by 21.3\%. It also outperforms the second-best model (Gemini 2.5 Pro) by 4.1\% while being 15.5 times more cost-effective. Costs for closed models are measured with the 50\% batching discount offered by providers. The costs of open-weight models are estimated following ~\citet{olmocr}.}
    \label{fig:performance-cost}
\end{figure}

Historical documents---including letters, diaries, government records, and newspapers---are primary sources that offer invaluable insights into the thoughts, experiences, and contexts of individuals and communities in the past. The vast diversity of languages, including historical variants of modern languages and dead languages, alongside evolving writing conventions, handwriting techniques, spelling, and typefaces, requires specialized expertise to study these materials.

Moreover, document degradation caused by environmental factors over time further complicates text extraction~\citep{jimaging5040048}. Historical documents frequently contain features such as marginalia~\citep{cheng-etal-2024-uncovering}, rubrication~\citep{Whetter_2017}, and illumination~\citep{wyatt1861history}, which are rarely encountered in modern texts. In addition, the wide variety of fonts used in printed documents and the lack of consistent orthographic standards (e.g., multiple spellings of the same word within a single document) pose further challenges~\citep{drobac-etal-2017-ocr}.

Handwritten text recognition (HTR)~\citep{jimaging10010018} is even more challenging due to the vast diversity of handwriting styles and calligraphic traditions~\citep{blair2020islamic}, even within the same language. Historical scribes frequently abbreviated words to expedite writing, resulting in thousands of unique abbreviations~\citep{candido-aluisio-2009-building}, especially in pre-modern documents~\citep{Guville2022TranscribingMM, rogos2025making}.

To improve accessibility, many efforts have been undertaken over the past two decades to digitize historical materials by scanning and converting them into machine-readable text~\citep{oberbichler2022integrated}. The current state-of-the-art approach involves training collection-specific models using fine-grained (paragraph-level or line-level) annotated data~\citep{reul2019ocr4all}. Various academic competitions (e.g., RASM~\citep{rasm2018}, REID~\citep{reid2017, reid2019}) have been organized to advance these efforts. Several platforms such as Transkribus~\citep{transcribus2017} and eScriptorium~\citep{eScriptorium}, provide tools for annotation and model training. However, reliance on annotated training data remains a significant bottleneck.

The recent advent of vision-language models (VLMs)~\citep{llava2023, zhang2023internlm, bai2023qwenvl, geminiteam2023, openai2024gpt4technicalreport} with strong zero-shot vision capabilities holds promise for addressing these challenges. OCR-related benchmarks such as DocVQA~\citep{mathew2021docvqa}, CharXiv~\citep{wang2024charxiv}, CC-OCR~\citep{yang2024cc}, OmniDocBench~\citep{ouyang2024omnidocbench}, and OCRBench~\citep{liu2024ocrbench, fu2024ocrbench} have become standard among VLM developers. However, these benchmarks primarily target modern documents. 

\begin{figure*}[th!]
\centering
\subfigure[Printed]{%
\includegraphics[trim=0 0 195 0, clip, width=0.46\textwidth]{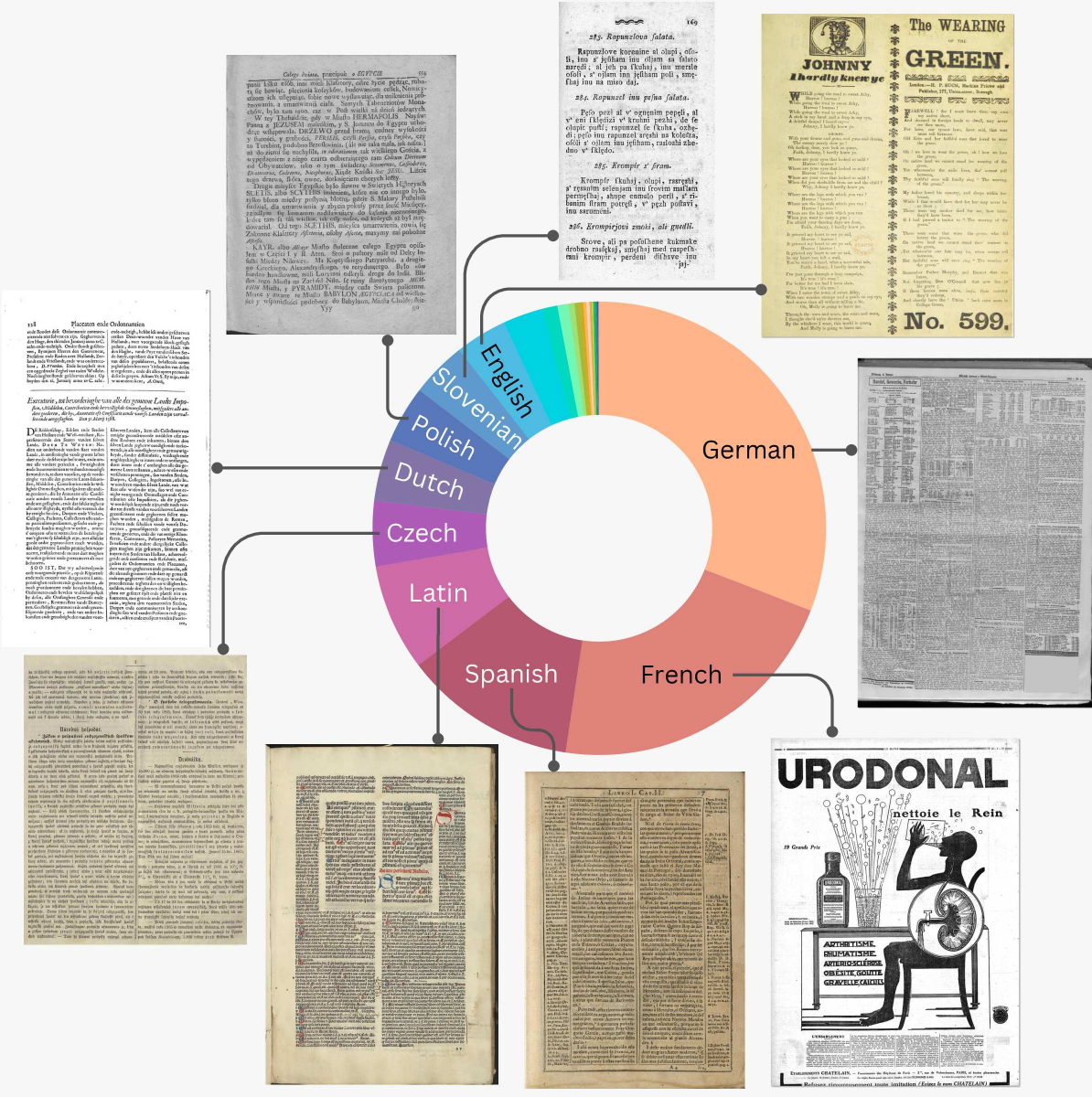}
\label{fig:fig2}
}%
\subfigure[Handwritten]{%
\includegraphics[trim=0 0 100 0, clip, width=0.54\textwidth]{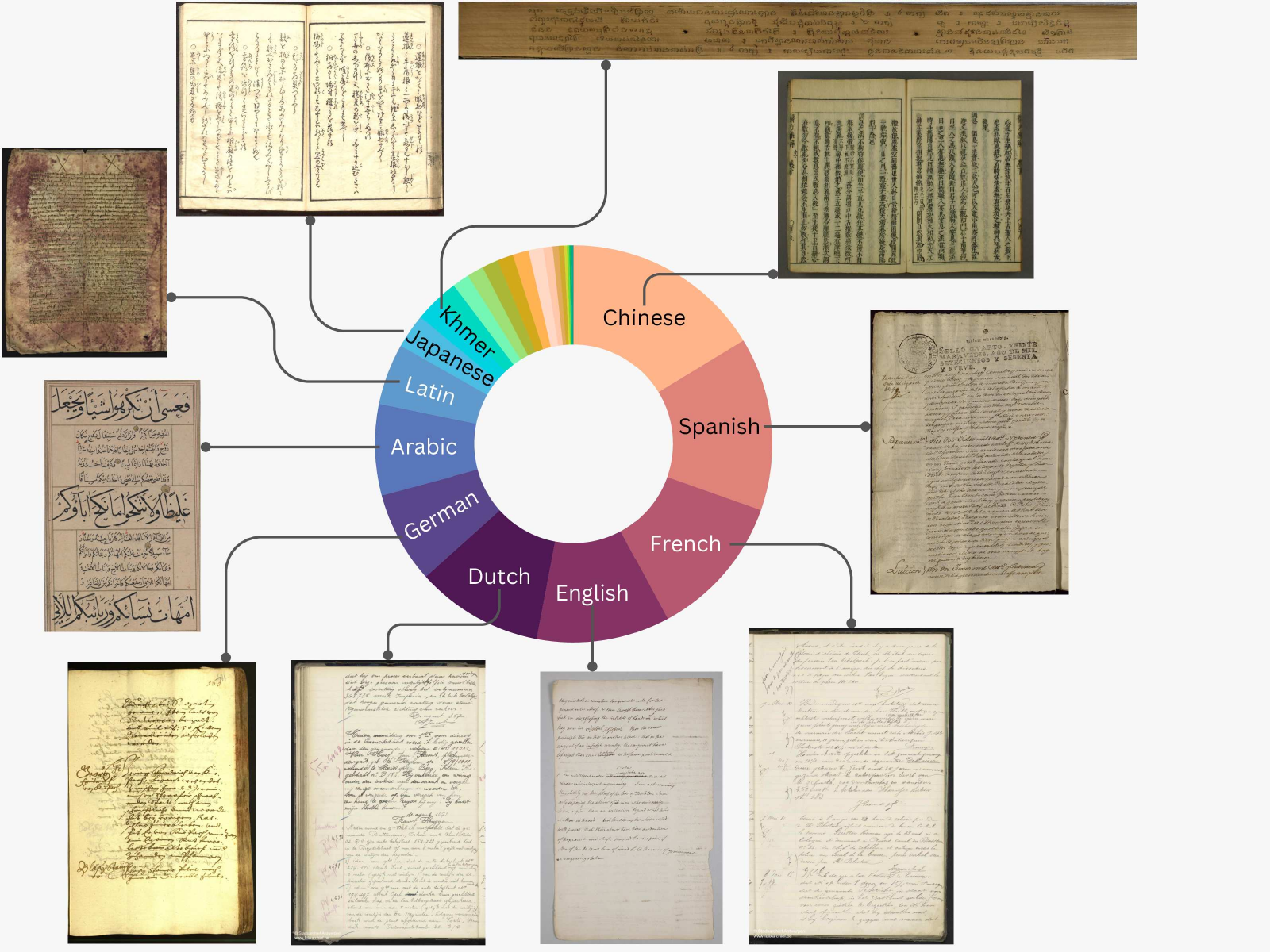}
\label{fig:fig1}
}%
\caption{Overview and examples from the two subsets of \dataset. Donut charts show the distribution of \emph{language clusters}. For example, the German cluster includes Modern German, Old High German, Middle High German, and Middle Low German.}
\label{fig:dataset}
\end{figure*}

{\bf We introduce \dataset, a large-scale page-level dataset for OCR and HTR on historical corpora.} 
\dataset is the most diverse historical dataset to date, created by unifying and cleaning over 150 existing datasets. It contains 99,491 documents from 46 \emph{language clusters}, 29 of which appear in the validation and test sets: Arabic, Bangla, Bulgarian, Catalan, Chinese, Czech, Dutch, English, Finnish, French, German, Greek, Hebrew, Hindi, Italian, Japanese, Khmer, Latin, Norwegian, Persian, Polish, Portuguese, Romanian, Sanskrit, Slovenian, Spanish, Swedish, Turkish, and Vietnamese. The dataset also encompasses 14 \emph{scripts} from European, East Asian, Southeast Asian, Middle Eastern, and Indic script families. Documents include newspapers, books, handwritten diaries, government records, and more, spanning from the 3rd century BC to the 20th century and covering three writing directions. Each page includes an image paired with a full-text annotation, making it suitable for both benchmarking and training VLMs. Figure~\ref{fig:dataset} shows the breakdown of language clusters and several example pages from \dataset.

{\bf We introduce \modelWithEmoji, a low-cost open-weight VLM that outperforms larger models} 
by fine-tuning a relatively small 3-billion-parameter VLM on \dataset. The \model VLM achieves 
82.3\% and 70.1\% on the printed and handwritten subsets of \dataset, respectively, outperforming all open and commercial VLMs by at least 1.4\% and 6.5\%.
This result highlights the effectiveness of specialized training on carefully curated historical documents.

{\bf We benchmark a wide range of VLMs and OCR systems.}
To the best of our knowledge, this is the first study assessing how well VLMs process images of historical documents. Our findings reveal substantial room for improvement among state-of-the-art VLMs in the zero-shot setting. The best-performing commercial model achieves 80.9\% and 63.6\% normalized Levenshtein similarity on printed and handwritten documents, respectively, averaged over all languages in \dataset. The best open-weight VLMs we evaluated lag behind, achieving 72.7\% and 54.5\%, respectively.

\section{Related Work}

\subsection{Document Understanding Benchmarks}
Most existing document benchmarks focus on modern documents and layouts.
PubLayNet~\citep{zhong2019publaynet} and DocBank~\citep{li2020docbank} automatically annotate PDF articles from PubMed Central and arXiv, respectively, for layout segmentation. The RVL-CDIP~\citep{harley2015evaluation} and IIT-CDIP~\citep{lewis2006building} datasets are derived from industry documents for document classification and retrieval. CCpdf~\citep{turski2023ccpdf} and WordScape~\citep{weber2023wordscape} leverage Common Crawl to extract and annotate multilingual, visually rich documents.

Building on this foundation, recent benchmarks such as CC-OCR~\citep{yang2024cc} and OCRBench~\citep{liu2024ocrbench, fu2024ocrbench} compile datasets for text recognition, visual question answering, and key information extraction across many domains into unified benchmarks, primarily to evaluate VLMs. However, these efforts typically remain focused on short outputs, modern documents, and text in English or Chinese~\cite{mmt-bench-2024, liu2023mmbench}.

Concurrently, research trends are shifting toward direct, page-level OCR rather than pipeline methods that rely on bounding-box detection. For instance, recent works introduce models capable of page-level OCR but report only limited evaluations—either on noisy, LLM-annotated data~\cite{olmocr} or only in English and Chinese~\cite{wang2024qwen2, yang2024qwen2}. Our proposed \dataset advances this direction by providing curated, human-annotated page-level transcriptions across many languages, with explicit page-level reading order.

\subsection{Vision-Language Models}
CLIP~\citep{radford2021learning} and its successors, such as SigLIP~\citep{zhai2023sigmoid}, laid the foundation for contrastive pretraining between vision and language.
Instruction-tuned open-weight VLMs like LLaVA~\citep{llava2023}, LLaVA-Next~\citep{liu2024llavanext}, InternVL~\citep{chen2024internvl}, and Qwen-VL~\citep{bai2023qwenvl} have demonstrated strong performance in reasoning and fine-grained text understanding. \citet{lee2023pix2struct, wang2024qwen2} extended these capabilities to variable-resolution input images, and \citet{wang2024qwen2, team2025granite} incorporated localization and bounding-box detection. These efforts bridge the gap between general-purpose VLMs and specialized models for document structure and layout understanding~\citep{blecher2023nougat}.
Proprietary LLMs have also incorporated vision capabilities, such as GPT-4V~\citep{openai2024gpt4technicalreport} and Claude 3~\citep{claude3}. Gemini 1.5~\citep{geminiteam2023} further improved text localization.
Recently, \citet{huang2025visionr1} and \citet{guo2025seed15vltechnicalreport} extended the explicit reasoning capabilities of LLMs to vision.

\section{The \dataset Dataset}

\dataset is a comprehensive benchmark and training resource curated for recognizing printed and handwritten text in historical documents. We standardize diverse annotation formats into a single text string per page in proper reading order. This transformation reduces the problem to pure image-to-text, enabling straightforward evaluation of vision-language models.

Through an extensive survey of NLP, computer vision, and humanities literature from the past two decades, we identified 153 historical corpora containing document page images and their corresponding transcriptions. We also release the first publicly available historical OCR datasets for Persian and Turkish.
\dataset reflects the practical challenges of processing historical scans: images often contain irregularities such as page creases, scanning artifacts, and extra margins (e.g., parts of scanning tables). Although all images are correctly oriented, slight misalignments are common.

\subsection{Dataset Selection}

\dataset comprises two subsets: one for printed documents and one for handwritten documents. Figure~\ref{fig:dataset} provides an overview of the languages included in each subset, along with illustrative examples. Appendix~\ref{appendix:dataset-list} lists all included datasets.

The selection criteria were:
1) Documents must originate prior to the mid-20th century. 2) Datasets must have licenses permitting research use. 3) Datasets must include human-annotated, gold-standard text covering entire pages. 
 
We exclude distantly supervised datasets, such as VieBookRead~\citep{do2024referencebased}.
Many historical datasets provide annotations at the character, word, or line level~\citep{hhd-ethiopic-dataset, romanello_optical_2021}; we include these if their annotations can be combined to reconstruct complete page-level text. 

Many of these datasets were originally transcribed by humanities researchers and have not previously been used for OCR, because they lack bounding-box annotations typically required by OCR models. By adopting a page-level task definition, we can leverage these previously untapped resources. Additionally, we create and include the first publicly available HTR dataset for Ottoman Turkish, transcribed by the authors.

\subsection{Dataset Curation}
\label{sec:dataset-curation}

Each example in \dataset consists of (1) an image of a document page, (2) page-level metadata indicating language and script, and (3) a text string representing the page content. The metadata enables fine-grained analysis of evaluated models.

Because the source datasets come from many different origins, they are transcribed in various formats and follow different conventions. To facilitate effective training, we carefully curate the datasets with the goal of producing an output text string that is (1) in the {\bf proper reading order} of the page, (2) as {\bf accurate} as possible, and (3) as {\bf faithful} as possible to the image, a practice known as \emph{diplomatic transcription}~\citep{diplomatic2019}. We achieve this through a combination of automated (LLM-based) transformations and manual verification, as described below.

\textbf{Page-Level Metadata.}
Each image is annotated with metadata indicating language and script. Such information is not always available from the original datasets. Some datasets lack language annotations entirely, while others (e.g., \citet{europeana, impact, entangled, icdar2017, slovensky}) span multiple languages without indicating which page corresponds to which language.

We leverage ISO 639-3~\citep{iso639-3}, which contains 7,916 languages and historical variants, grouping them into \emph{language clusters} based on macrolanguage classifications. For languages without macrolanguage categories (e.g., English or German), we use ISO reference codes for contemporary forms as umbrella categories for historical variants. We prompt a VLM with descriptions and examples from each cluster to detect the language of each page. When available, we use dataset documentation to constrain possible language clusters. We then manually review subsets for accuracy.

We classify scripts similarly, using ISO 15924~\citep{iso15924} and Glottolog~\cite{glottolog} script families, yielding 14 distinct scripts from 5 families. Tables~\ref{tab:language_stats} and~\ref{tab:dataset-script-statistics} list all language clusters, scripts, and script families in \dataset.

\textbf{Reading Order.}
We standardize various annotation formats (e.g., ALTO XML~\citep{altoxml}, PAGE XML~\citep{pagexml}, JSON, plain text) into a single text string per page in correct reading order. While ALTO and PAGE XML support explicit reading-order annotations, many datasets lack them. For datasets with consistent layouts (e.g., single- or double-column), we develop dataset-specific heuristics. For more complex layouts, we employ a VLM to determine reading order and manually spot-check outputs. Manual checks indicate that more than 98\% of reading orders are accurate.

\textbf{Accuracy.}
We identified annotation issues in sixteen datasets. For instance, \citet{scribble} omits text in top margins, and datasets from \citet{charters} and \citet{scarce} contain frequent transcription errors. To address these problems, we use a VLM (Gemini 2.5 Pro~\citep{google2025gemini-2.5}) to suggest corrections based on image bounding boxes. Corrections are subsequently reviewed and validated by the authors.

\textbf{Faithfulness.}
Our dataset includes scholarly editions from diverse institutions, each following distinct transcription standards~\citep{salamanca-guidelines}. Some editions employ \emph{modernized transcriptions} with standardized spelling, punctuation, and capitalization, while others adopt \emph{diplomatic transcriptions}~\citep{diplomatic2019} to faithfully represent original documents.

For \dataset, we standardize on diplomatic transcriptions, ensuring evaluated systems need only \emph{transcribe what they see}, independent of dataset-specific guidelines. In datasets such as \citet{notarial}, which include both original and expanded forms of abbreviations and hyphens, we retain original transcriptions. We exclude significantly modernized datasets where faithful transcription is unrecoverable.

To restore spacing, line breaks, and diacritics from images, we use a VLM guided by image bounding boxes and transcribed text to perform minimal edits. Character-level edit distances between original and edited transcriptions remain below 5\%. We manually verify samples from each dataset for accuracy.

Finally, we normalize typographic elements such as fractions (e.g., converting \sfrac{1}{4}, Unicode \texttt{U+00BC}, to 1/4) and ligatures (e.g., Unicode \texttt{U+FB06} to ``st'').

\subsection{Dataset Preparation}

\textbf{Data Filtering.}
Some pages contain minimal text, such as chapter dividers and title pages. We remove pages with fewer than 30 tokens (tokenized by Qwen 2.5 VL~\citep{bai2025qwen25vltechnicalreport}). Duplicate pages arising from dataset errors or reuse across datasets are identified and removed using MinHash~\citep{broder1997resemblance}, eliminating approximately 15,000 near-duplicates. Large images are resized to fit within a 2500~$\times$2500 pixel box, preserving aspect ratios. Appendix~\ref{appendix:dataset-stats} provides text and image size statistics.

\textbf{Dataset Splits.}
To facilitate efficient evaluation across diverse language clusters, we construct balanced development and test sets by randomly sampling 60 pages per language cluster per document type (printed and handwritten). These samples are evenly divided into validation and test sets, with the remaining documents forming the training set. Due to limited examples for some languages, only 29 of the 46 language clusters appear in the validation and test sets; the remaining 17 appear only in the training set. The resulting dataset contains 97,151 training samples and 1,170 samples each for validation and test sets. Note that training sizes vary substantially, with as few as 6 and 12 samples for printed Chinese and handwritten Portuguese, respectively. Table~\ref{tab:language_stats} summarizes the dataset distribution across languages and splits.

\section{Experiments and Results}

We first evaluate a broad set of commercial and open-weight VLMs on the \dataset test set in a zero-shot setting. We then investigate the impact of fine-tuning using the \dataset training set.

\subsection{Experimental Setup}

\paragraph{\modelWithEmoji.}
We fine-tune a VLM on \dataset to obtain \model. We selected the 3-billion-parameter Qwen 2.5 VL for its compact size, strong zero-shot OCR performance, and ability to handle high-resolution, variable-sized images--all critical for processing diverse layouts. We fine-tuned this model on the training portion of \dataset.

\paragraph{VLMs.}
We select the top proprietary VLMs from the LM Arena Vision Leaderboard~\citep{chiang2024chatbot}, excluding earlier versions of some models that were superseded by updated releases. The selected models are
GPT-5, GPT-5 Mini, and GPT-5 Nano~\cite{openai2025gpt5};
GPT-4.1, GPT-4.1 Mini, and GPT-4.1 Nano~\citep{openai2025gpt41};
GPT-4o and GPT-4o Mini~\citep{openai2024gpt4technicalreport};
O1~\cite{openai2024-o1}, O3, and O4 Mini~\citep{openai2024-o3-o4mini};
Claude Sonnet 3.7~\citep{claude3} and 4~\cite{claude4};
Claude Opus 4.1~\cite{claude4.1};
and Gemini 2.5 Flash and Pro~\citep{google2025gemini-2.5}.

We also evaluate several strong open-weight, instruction-tuned VLMs of varying sizes: Qwen 2.5 VL (3B and 72B)\citep{bai2025qwen25vltechnicalreport};
Gemma 3 27B\citep{team2024gemma};
MiMo VL (8B)\cite{coreteam2025mimovltechnicalreport};
Nemotron Nano VL (8B)\cite{nvidia_2025_llama_3_1_nemotron_nano_vl_8b_v1};
InternVL 3.5 (30B)\cite{wang2025internvl35advancingopensourcemultimodal};
R (4B)\cite{r4b};
and Phi 4 Multimodal (5B)~\cite{microsoft2025phi4mini};
and VLMs specifically tuned for page-level OCR:
NuMarkdown (8B)\cite{numind_2025_numarkdown_8b},
olmOCR (8B)\cite{olmocr},
RolmOCR (8B)\cite{RolmOCR}, and
Nanonets OCR (3B)\cite{Nanonets-OCR-S}. We exclude models such as DeepSeek-VL2~\cite{wu2024deepseekvl2}, whose supported context length is insufficient for the long texts in \dataset.
All models are evaluated using the same prompt in Table~\ref{tab:system_prompt}.

\textbf{OCR Systems.}
We also benchmark several specialized OCR systems:
Azure OCR~\citep{azureocr}, which supports handwritten text recognition in five languages and printed text recognition in over 100 languages;
and Mistral OCR~\citep{mistral2025ocr}, a multilingual OCR engine optimized for long-form, dense documents.

Unlike VLMs, which directly generate complete text from images, Azure OCR is a \emph{pipeline model}: it first detects text bounding boxes, performs OCR within these boxes, and then determines the correct reading order.

Recognizing the difficulty VLMs face with long inputs, we also evaluate a \emph{hybrid} system that combines Azure OCR with a VLM. In this approach, Azure OCR identifies bounding boxes and their reading order, after which a VLM processes each bounding box individually. This division simplifies the VLM’s task by providing smaller image inputs and requiring shorter textual outputs.

We further report the performance of an oracle model that selects, for each example, the highest score among all models. This establishes an upper bound for ensembling methods that combine outputs from multiple systems.

\subsection{Evaluation Metrics}

To enable direct comparison across fundamentally different approaches, we follow \citet{blecher2023nougat} and evaluate each system’s final textual output against the gold annotations. Specifically, we compute the character-level Levenshtein distance~\citep{levenshtein1966binary}, which measures the number of insertions, deletions, and substitutions required to transform one string into another. We normalize this distance by the length of the longer string and convert it to similarity by subtracting it from 1, yielding a score in [0, 1]. This metric closely aligns with the Character Recognition Rate (1 minus Character Error Rate) commonly used in OCR studies~\citep{neudecker2021ocr, karpinski2018ocr, liebl2020accuracycrnnslinebasedocr}, but is bounded within [0, 1], enabling comparison across varying text lengths. We refer to this metric as \emph{normalized Levenshtein similarity}.

We compute the metric for each example and average the scores separately for printed and handwritten document types. For fairness, we apply the same normalization procedures (described in Section~\ref{sec:dataset-curation}) to all model predictions.
For texts in Arabic script, we perform additional normalization of diacritics and hamza using the PyArabic toolkit~\citep{zerrouki2012pyarabic}.

\subsection{Evaluation Results on Printed Documents}

\begin{table*}[ht!]
\centering
\small
\setlength{\tabcolsep}{3pt} 
\resizebox{1\textwidth}{!}{
\begin{tabular}{l|rrrrrrrrrrrrrrrrrrr}
\toprule

System & ben & bul & ces & deu & eng & fin & fra & hin & jpn & lat & nld & pol & ron & san & slv & spa & swe & zho & Avg \\

\midrule
\multicolumn{20}{c}{VLM Fine-Tuned on \dataset}\\
\midrule

\modelWithEmoji (3B) & \cellcolorvalue{70.9} & \cellcolorvalue{96.1} & \cellcolorvalue{95.6} & \cellcolorvalue{82.3} & \cellcolorvalue{91.0} & \cellcolorvalue{63.2} & \cellcolorvalue{89.6} & \cellcolorvalue{94.6} & \cellcolorvalue{74.1} & \cellcolorvalue{92.0} & \cellcolorvalue{95.7} & \cellcolorvalue{81.6} & \cellcolorvalue{79.6} & \cellcolorvalue{93.1} & \cellcolorvalue{97.6} & \cellcolorvalue{91.4} & \cellcolorvalue{87.2} & \cellcolorvalue{6.2} & \cellcolorvalue{82.3} \\
$\Delta$ with Qwen 2.5 VL (3B) & \textcolor{DarkGreen}{+0.9} & \textcolor{DarkGreen}{+12.8} & \textcolor{DarkGreen}{+2.0} & \textcolor{DarkGreen}{+17.7} & \textcolor{DarkGreen}{+9.1} & \textcolor{DarkGreen}{+32.9} & \textcolor{DarkGreen}{+9.7} & \textcolor{DarkGreen}{+16.7} & \textcolor{DarkGreen}{+37.0} & \textcolor{DarkGreen}{+11.6} & \textcolor{DarkGreen}{+20.0} & \textcolor{DarkGreen}{+14.0} & \textcolor{DarkGreen}{+20.5} & \textcolor{DarkGreen}{+20.6} & \textcolor{DarkGreen}{+2.8} & \textcolor{DarkGreen}{+4.2} & \textcolor{DarkGreen}{+22.9} & \textcolor{DarkGreen}{+5.7} & \textcolor{DarkGreen}{+14.5} \\

\midrule
\multicolumn{20}{c}{Closed VLMs (Zero-Shot)}\\
\midrule
Gemini 2.5 Pro & \cellcolorvalue{94.5} & \cellcolorvalue{93.4} & \cellcolorvalue{95.6} & \cellcolorvalue{76.9} & \cellcolorvalue{93.3} & \cellcolorvalue{55.3} & \cellcolorvalue{93.5} & \cellcolorvalue{93.2} & \cellcolorvalue{61.7} & \cellcolorvalue{87.5} & \cellcolorvalue{97.9} & \cellcolorvalue{80.3} & \cellcolorvalue{66.3} & \cellcolorvalue{89.7} & \cellcolorvalue{97.4} & \cellcolorvalue{91.3} & \cellcolorvalue{81.1} & \cellcolorvalue{6.6} & \cellcolorvalue{80.9} \\
Gemini 2.5 Flash & \cellcolorvalue{91.2} & \cellcolorvalue{91.0} & \cellcolorvalue{94.8} & \cellcolorvalue{69.3} & \cellcolorvalue{84.6} & \cellcolorvalue{32.8} & \cellcolorvalue{80.9} & \cellcolorvalue{78.6} & \cellcolorvalue{47.1} & \cellcolorvalue{85.0} & \cellcolorvalue{87.1} & \cellcolorvalue{73.7} & \cellcolorvalue{60.2} & \cellcolorvalue{82.1} & \cellcolorvalue{97.0} & \cellcolorvalue{94.4} & \cellcolorvalue{68.9} & \cellcolorvalue{8.3} & \cellcolorvalue{73.7} \\
GPT-4.1 Mini & \cellcolorvalue{77.3} & \cellcolorvalue{90.3} & \cellcolorvalue{93.8} & \cellcolorvalue{75.3} & \cellcolorvalue{88.3} & \cellcolorvalue{40.1} & \cellcolorvalue{76.6} & \cellcolorvalue{82.2} & \cellcolorvalue{55.3} & \cellcolorvalue{82.3} & \cellcolorvalue{82.7} & \cellcolorvalue{73.1} & \cellcolorvalue{56.1} & \cellcolorvalue{71.9} & \cellcolorvalue{96.3} & \cellcolorvalue{89.9} & \cellcolorvalue{79.0} & \cellcolorvalue{5.3} & \cellcolorvalue{73.1} \\
Claude Sonnet 3.7 & \cellcolorvalue{85.2} & \cellcolorvalue{86.4} & \cellcolorvalue{94.8} & \cellcolorvalue{64.7} & \cellcolorvalue{80.1} & \cellcolorvalue{25.8} & \cellcolorvalue{76.1} & \cellcolorvalue{89.8} & \cellcolorvalue{29.8} & \cellcolorvalue{85.0} & \cellcolorvalue{76.1} & \cellcolorvalue{71.3} & \cellcolorvalue{64.1} & \cellcolorvalue{82.1} & \cellcolorvalue{95.0} & \cellcolorvalue{84.8} & \cellcolorvalue{70.2} & \cellcolorvalue{1.8} & \cellcolorvalue{70.2} \\
GPT-5 Mini & \cellcolorvalue{78.6} & \cellcolorvalue{83.8} & \cellcolorvalue{92.6} & \cellcolorvalue{67.8} & \cellcolorvalue{86.4} & \cellcolorvalue{39.3} & \cellcolorvalue{83.4} & \cellcolorvalue{78.2} & \cellcolorvalue{44.3} & \cellcolorvalue{81.5} & \cellcolorvalue{80.5} & \cellcolorvalue{70.4} & \cellcolorvalue{43.5} & \cellcolorvalue{65.5} & \cellcolorvalue{92.9} & \cellcolorvalue{90.6} & \cellcolorvalue{72.7} & \cellcolorvalue{4.6} & \cellcolorvalue{69.8} \\
Claude Opus 4.1 & \cellcolorvalue{82.3} & \cellcolorvalue{82.9} & \cellcolorvalue{91.5} & \cellcolorvalue{58.7} & \cellcolorvalue{78.3} & \cellcolorvalue{16.8} & \cellcolorvalue{77.7} & \cellcolorvalue{87.5} & \cellcolorvalue{17.3} & \cellcolorvalue{84.1} & \cellcolorvalue{70.1} & \cellcolorvalue{65.3} & \cellcolorvalue{57.7} & \cellcolorvalue{79.5} & \cellcolorvalue{96.3} & \cellcolorvalue{85.7} & \cellcolorvalue{65.2} & \cellcolorvalue{1.8} & \cellcolorvalue{66.6} \\
O4 Mini & \cellcolorvalue{75.8} & \cellcolorvalue{81.9} & \cellcolorvalue{89.2} & \cellcolorvalue{63.2} & \cellcolorvalue{82.3} & \cellcolorvalue{21.2} & \cellcolorvalue{80.5} & \cellcolorvalue{73.3} & \cellcolorvalue{38.8} & \cellcolorvalue{79.5} & \cellcolorvalue{74.2} & \cellcolorvalue{64.0} & \cellcolorvalue{39.4} & \cellcolorvalue{59.6} & \cellcolorvalue{95.1} & \cellcolorvalue{89.0} & \cellcolorvalue{69.6} & \cellcolorvalue{2.4} & \cellcolorvalue{65.5} \\
GPT-4.1 & \cellcolorvalue{71.4} & \cellcolorvalue{78.3} & \cellcolorvalue{89.5} & \cellcolorvalue{63.7} & \cellcolorvalue{81.8} & \cellcolorvalue{34.5} & \cellcolorvalue{75.0} & \cellcolorvalue{76.4} & \cellcolorvalue{19.7} & \cellcolorvalue{76.9} & \cellcolorvalue{76.6} & \cellcolorvalue{64.8} & \cellcolorvalue{40.9} & \cellcolorvalue{60.6} & \cellcolorvalue{90.6} & \cellcolorvalue{88.0} & \cellcolorvalue{66.5} & \cellcolorvalue{3.9} & \cellcolorvalue{64.4} \\
O3 & \cellcolorvalue{76.5} & \cellcolorvalue{77.8} & \cellcolorvalue{88.0} & \cellcolorvalue{58.4} & \cellcolorvalue{81.6} & \cellcolorvalue{19.7} & \cellcolorvalue{80.1} & \cellcolorvalue{76.3} & \cellcolorvalue{12.0} & \cellcolorvalue{79.6} & \cellcolorvalue{73.1} & \cellcolorvalue{65.5} & \cellcolorvalue{33.6} & \cellcolorvalue{59.8} & \cellcolorvalue{94.8} & \cellcolorvalue{89.4} & \cellcolorvalue{60.0} & \cellcolorvalue{1.5} & \cellcolorvalue{62.7} \\
O1 & \cellcolorvalue{74.7} & \cellcolorvalue{76.0} & \cellcolorvalue{87.5} & \cellcolorvalue{58.9} & \cellcolorvalue{82.1} & \cellcolorvalue{24.3} & \cellcolorvalue{72.2} & \cellcolorvalue{78.1} & \cellcolorvalue{15.6} & \cellcolorvalue{77.5} & \cellcolorvalue{74.0} & \cellcolorvalue{64.2} & \cellcolorvalue{35.3} & \cellcolorvalue{60.2} & \cellcolorvalue{90.4} & \cellcolorvalue{89.0} & \cellcolorvalue{62.6} & \cellcolorvalue{2.0} & \cellcolorvalue{62.5} \\
Claude Sonnet 4 & \cellcolorvalue{73.3} & \cellcolorvalue{77.6} & \cellcolorvalue{86.6} & \cellcolorvalue{56.1} & \cellcolorvalue{76.2} & \cellcolorvalue{15.8} & \cellcolorvalue{69.9} & \cellcolorvalue{76.5} & \cellcolorvalue{17.5} & \cellcolorvalue{83.1} & \cellcolorvalue{70.7} & \cellcolorvalue{63.4} & \cellcolorvalue{51.0} & \cellcolorvalue{69.1} & \cellcolorvalue{93.8} & \cellcolorvalue{78.9} & \cellcolorvalue{58.8} & \cellcolorvalue{2.0} & \cellcolorvalue{62.2} \\
GPT-5 & \cellcolorvalue{73.6} & \cellcolorvalue{77.1} & \cellcolorvalue{86.9} & \cellcolorvalue{59.2} & \cellcolorvalue{81.1} & \cellcolorvalue{19.9} & \cellcolorvalue{79.8} & \cellcolorvalue{61.5} & \cellcolorvalue{14.2} & \cellcolorvalue{78.2} & \cellcolorvalue{71.8} & \cellcolorvalue{64.2} & \cellcolorvalue{33.1} & \cellcolorvalue{53.0} & \cellcolorvalue{90.3} & \cellcolorvalue{89.6} & \cellcolorvalue{58.6} & \cellcolorvalue{1.3} & \cellcolorvalue{60.7} \\
GPT-4o & \cellcolorvalue{66.7} & \cellcolorvalue{63.0} & \cellcolorvalue{82.9} & \cellcolorvalue{48.8} & \cellcolorvalue{78.3} & \cellcolorvalue{15.6} & \cellcolorvalue{70.2} & \cellcolorvalue{70.6} & \cellcolorvalue{11.9} & \cellcolorvalue{74.2} & \cellcolorvalue{62.8} & \cellcolorvalue{54.8} & \cellcolorvalue{40.1} & \cellcolorvalue{48.1} & \cellcolorvalue{89.5} & \cellcolorvalue{86.4} & \cellcolorvalue{47.1} & \cellcolorvalue{2.1} & \cellcolorvalue{56.3} \\
GPT-4o Mini & \cellcolorvalue{50.9} & \cellcolorvalue{58.7} & \cellcolorvalue{76.9} & \cellcolorvalue{48.9} & \cellcolorvalue{77.1} & \cellcolorvalue{17.0} & \cellcolorvalue{66.6} & \cellcolorvalue{57.8} & \cellcolorvalue{11.9} & \cellcolorvalue{69.4} & \cellcolorvalue{56.9} & \cellcolorvalue{52.1} & \cellcolorvalue{43.8} & \cellcolorvalue{42.1} & \cellcolorvalue{85.2} & \cellcolorvalue{85.4} & \cellcolorvalue{47.9} & \cellcolorvalue{2.4} & \cellcolorvalue{52.8} \\
GPT-4.1 Nano & \cellcolorvalue{36.2} & \cellcolorvalue{61.0} & \cellcolorvalue{82.5} & \cellcolorvalue{54.4} & \cellcolorvalue{81.3} & \cellcolorvalue{20.4} & \cellcolorvalue{67.8} & \cellcolorvalue{34.1} & \cellcolorvalue{14.0} & \cellcolorvalue{71.9} & \cellcolorvalue{58.2} & \cellcolorvalue{58.4} & \cellcolorvalue{41.4} & \cellcolorvalue{26.1} & \cellcolorvalue{86.3} & \cellcolorvalue{82.3} & \cellcolorvalue{53.6} & \cellcolorvalue{3.6} & \cellcolorvalue{51.9} \\
GPT-5 Nano & \cellcolorvalue{28.2} & \cellcolorvalue{27.9} & \cellcolorvalue{68.2} & \cellcolorvalue{41.1} & \cellcolorvalue{73.0} & \cellcolorvalue{5.3} & \cellcolorvalue{65.3} & \cellcolorvalue{28.5} & \cellcolorvalue{6.1} & \cellcolorvalue{56.9} & \cellcolorvalue{32.2} & \cellcolorvalue{38.7} & \cellcolorvalue{27.3} & \cellcolorvalue{13.0} & \cellcolorvalue{78.0} & \cellcolorvalue{69.5} & \cellcolorvalue{37.2} & \cellcolorvalue{0.5} & \cellcolorvalue{38.7} \\

\midrule
\multicolumn{20}{c}{Open-Weight VLMs (Zero-Shot)}\\
\midrule

NuMarkdown (8B) & \cellcolorvalue{84.7} & \cellcolorvalue{87.1} & \cellcolorvalue{95.7} & \cellcolorvalue{71.9} & \cellcolorvalue{83.0} & \cellcolorvalue{39.0} & \cellcolorvalue{87.1} & \cellcolorvalue{75.5} & \cellcolorvalue{55.2} & \cellcolorvalue{80.6} & \cellcolorvalue{84.7} & \cellcolorvalue{77.4} & \cellcolorvalue{52.9} & \cellcolorvalue{73.3} & \cellcolorvalue{95.6} & \cellcolorvalue{86.9} & \cellcolorvalue{77.3} & \cellcolorvalue{1.1} & \cellcolorvalue{72.7} \\
olmOCR (8B) & \cellcolorvalue{79.2} & \cellcolorvalue{84.8} & \cellcolorvalue{90.0} & \cellcolorvalue{66.4} & \cellcolorvalue{79.7} & \cellcolorvalue{28.5} & \cellcolorvalue{83.7} & \cellcolorvalue{85.0} & \cellcolorvalue{42.5} & \cellcolorvalue{82.5} & \cellcolorvalue{82.9} & \cellcolorvalue{72.4} & \cellcolorvalue{51.4} & \cellcolorvalue{66.4} & \cellcolorvalue{96.2} & \cellcolorvalue{90.7} & \cellcolorvalue{73.2} & \cellcolorvalue{0.2} & \cellcolorvalue{69.8} \\
Nanonets OCR (3B) & \cellcolorvalue{75.6} & \cellcolorvalue{85.5} & \cellcolorvalue{94.7} & \cellcolorvalue{69.4} & \cellcolorvalue{85.7} & \cellcolorvalue{24.8} & \cellcolorvalue{82.9} & \cellcolorvalue{83.1} & \cellcolorvalue{32.9} & \cellcolorvalue{79.7} & \cellcolorvalue{78.1} & \cellcolorvalue{74.2} & \cellcolorvalue{57.3} & \cellcolorvalue{71.4} & \cellcolorvalue{95.7} & \cellcolorvalue{91.2} & \cellcolorvalue{72.2} & \cellcolorvalue{0.4} & \cellcolorvalue{69.7} \\
Qwen 2.5 VL (3B) & \cellcolorvalue{70.0} & \cellcolorvalue{83.3} & \cellcolorvalue{93.6} & \cellcolorvalue{64.6} & \cellcolorvalue{82.0} & \cellcolorvalue{30.3} & \cellcolorvalue{79.9} & \cellcolorvalue{77.9} & \cellcolorvalue{37.1} & \cellcolorvalue{80.4} & \cellcolorvalue{75.7} & \cellcolorvalue{67.7} & \cellcolorvalue{59.1} & \cellcolorvalue{72.6} & \cellcolorvalue{94.8} & \cellcolorvalue{87.2} & \cellcolorvalue{64.3} & \cellcolorvalue{0.5} & \cellcolorvalue{67.8} \\
RolmOCR (8B) & \cellcolorvalue{82.7} & \cellcolorvalue{79.4} & \cellcolorvalue{87.5} & \cellcolorvalue{63.9} & \cellcolorvalue{88.4} & \cellcolorvalue{17.0} & \cellcolorvalue{78.4} & \cellcolorvalue{85.1} & \cellcolorvalue{35.1} & \cellcolorvalue{79.2} & \cellcolorvalue{73.7} & \cellcolorvalue{65.5} & \cellcolorvalue{57.9} & \cellcolorvalue{67.7} & \cellcolorvalue{86.1} & \cellcolorvalue{90.7} & \cellcolorvalue{70.8} & \cellcolorvalue{1.2} & \cellcolorvalue{67.2} \\
Qwen 2.5 VL (72B) & \cellcolorvalue{87.9} & \cellcolorvalue{89.6} & \cellcolorvalue{83.6} & \cellcolorvalue{53.3} & \cellcolorvalue{75.7} & \cellcolorvalue{10.6} & \cellcolorvalue{80.2} & \cellcolorvalue{89.0} & \cellcolorvalue{47.3} & \cellcolorvalue{85.3} & \cellcolorvalue{65.9} & \cellcolorvalue{57.1} & \cellcolorvalue{57.4} & \cellcolorvalue{70.2} & \cellcolorvalue{95.9} & \cellcolorvalue{89.5} & \cellcolorvalue{50.9} & \cellcolorvalue{3.7} & \cellcolorvalue{66.3} \\
Gemma 3 (27B) & \cellcolorvalue{61.6} & \cellcolorvalue{62.0} & \cellcolorvalue{78.3} & \cellcolorvalue{49.7} & \cellcolorvalue{70.4} & \cellcolorvalue{21.0} & \cellcolorvalue{66.8} & \cellcolorvalue{74.3} & \cellcolorvalue{18.3} & \cellcolorvalue{69.1} & \cellcolorvalue{50.4} & \cellcolorvalue{52.7} & \cellcolorvalue{48.7} & \cellcolorvalue{61.4} & \cellcolorvalue{89.4} & \cellcolorvalue{72.5} & \cellcolorvalue{44.0} & \cellcolorvalue{5.4} & \cellcolorvalue{55.3} \\
MiMo VL (8B) & \cellcolorvalue{63.8} & \cellcolorvalue{72.7} & \cellcolorvalue{82.4} & \cellcolorvalue{51.0} & \cellcolorvalue{71.5} & \cellcolorvalue{7.2} & \cellcolorvalue{72.6} & \cellcolorvalue{51.4} & \cellcolorvalue{26.8} & \cellcolorvalue{82.6} & \cellcolorvalue{50.4} & \cellcolorvalue{49.6} & \cellcolorvalue{40.0} & \cellcolorvalue{44.7} & \cellcolorvalue{92.4} & \cellcolorvalue{75.7} & \cellcolorvalue{50.5} & \cellcolorvalue{2.1} & \cellcolorvalue{54.8} \\
Skywork R1V3 (38B) & \cellcolorvalue{20.7} & \cellcolorvalue{35.8} & \cellcolorvalue{70.6} & \cellcolorvalue{42.3} & \cellcolorvalue{73.4} & \cellcolorvalue{17.6} & \cellcolorvalue{68.7} & \cellcolorvalue{18.8} & \cellcolorvalue{12.2} & \cellcolorvalue{67.3} & \cellcolorvalue{47.2} & \cellcolorvalue{46.8} & \cellcolorvalue{31.5} & \cellcolorvalue{17.1} & \cellcolorvalue{81.6} & \cellcolorvalue{72.3} & \cellcolorvalue{37.0} & \cellcolorvalue{2.7} & \cellcolorvalue{42.4} \\
InternVL 3.5 (30B) & \cellcolorvalue{16.4} & \cellcolorvalue{10.2} & \cellcolorvalue{56.5} & \cellcolorvalue{35.3} & \cellcolorvalue{76.3} & \cellcolorvalue{13.1} & \cellcolorvalue{64.6} & \cellcolorvalue{16.0} & \cellcolorvalue{13.2} & \cellcolorvalue{62.8} & \cellcolorvalue{42.7} & \cellcolorvalue{31.5} & \cellcolorvalue{18.8} & \cellcolorvalue{11.6} & \cellcolorvalue{77.3} & \cellcolorvalue{67.5} & \cellcolorvalue{27.8} & \cellcolorvalue{2.8} & \cellcolorvalue{35.8} \\
R (4B) & \cellcolorvalue{19.0} & \cellcolorvalue{17.2} & \cellcolorvalue{43.9} & \cellcolorvalue{28.4} & \cellcolorvalue{67.7} & \cellcolorvalue{13.9} & \cellcolorvalue{62.0} & \cellcolorvalue{25.2} & \cellcolorvalue{6.3} & \cellcolorvalue{48.8} & \cellcolorvalue{32.0} & \cellcolorvalue{27.3} & \cellcolorvalue{24.6} & \cellcolorvalue{16.9} & \cellcolorvalue{62.5} & \cellcolorvalue{62.8} & \cellcolorvalue{27.0} & \cellcolorvalue{2.9} & \cellcolorvalue{32.7} \\
Nemotron Nano VL (8B) & \cellcolorvalue{7.7} & \cellcolorvalue{14.2} & \cellcolorvalue{35.8} & \cellcolorvalue{34.3} & \cellcolorvalue{57.1} & \cellcolorvalue{6.6} & \cellcolorvalue{60.0} & \cellcolorvalue{11.7} & \cellcolorvalue{0.4} & \cellcolorvalue{55.9} & \cellcolorvalue{32.7} & \cellcolorvalue{33.1} & \cellcolorvalue{14.8} & \cellcolorvalue{9.4} & \cellcolorvalue{59.0} & \cellcolorvalue{63.4} & \cellcolorvalue{30.9} & \cellcolorvalue{1.1} & \cellcolorvalue{29.3} \\

Phi 4 Multimodal (5B) & \cellcolorvalue{2.2} & \cellcolorvalue{1.7} & \cellcolorvalue{4.0} & \cellcolorvalue{5.3} & \cellcolorvalue{42.9} & \cellcolorvalue{5.2} & \cellcolorvalue{11.6} & \cellcolorvalue{3.7} & \cellcolorvalue{1.8} & \cellcolorvalue{11.4} & \cellcolorvalue{7.0} & \cellcolorvalue{6.3} & \cellcolorvalue{5.6} & \cellcolorvalue{3.5} & \cellcolorvalue{25.6} & \cellcolorvalue{16.7} & \cellcolorvalue{7.4} & \cellcolorvalue{0.7} & \cellcolorvalue{9.0} \\

\midrule
\multicolumn{20}{c}{OCR Systems}\\
\midrule

Azure OCR & \cellcolorvalue{15.3} & \cellcolorvalue{91.4} & \cellcolorvalue{91.7} & \cellcolorvalue{77.9} & \cellcolorvalue{86.1} & \cellcolorvalue{65.0} & \cellcolorvalue{84.5} & \cellcolorvalue{72.9} & \cellcolorvalue{53.6} & \cellcolorvalue{82.6} & \cellcolorvalue{91.6} & \cellcolorvalue{81.1} & \cellcolorvalue{50.7} & \cellcolorvalue{80.8} & \cellcolorvalue{93.9} & \cellcolorvalue{85.5} & \cellcolorvalue{79.6} & \cellcolorvalue{9.4} & \cellcolorvalue{71.9} \\
Mistral OCR & \cellcolorvalue{59.4} & \cellcolorvalue{88.1} & \cellcolorvalue{86.5} & \cellcolorvalue{57.1} & \cellcolorvalue{79.3} & \cellcolorvalue{28.6} & \cellcolorvalue{74.7} & \cellcolorvalue{82.2} & \cellcolorvalue{3.3} & \cellcolorvalue{79.8} & \cellcolorvalue{78.9} & \cellcolorvalue{60.9} & \cellcolorvalue{48.6} & \cellcolorvalue{82.7} & \cellcolorvalue{93.8} & \cellcolorvalue{89.3} & \cellcolorvalue{58.1} & \cellcolorvalue{1.3} & \cellcolorvalue{64.0} \\
Azure OCR + Gemini 2.5 Pro & \cellcolorvalue{66.9} & \cellcolorvalue{57.3} & \cellcolorvalue{74.0} & \cellcolorvalue{42.4} & \cellcolorvalue{63.9} & \cellcolorvalue{10.8} & \cellcolorvalue{65.9} & \cellcolorvalue{56.7} & \cellcolorvalue{28.4} & \cellcolorvalue{73.2} & \cellcolorvalue{59.1} & \cellcolorvalue{48.8} & \cellcolorvalue{51.0} & \cellcolorvalue{59.2} & \cellcolorvalue{55.3} & \cellcolorvalue{81.0} & \cellcolorvalue{50.7} & \cellcolorvalue{2.4} & \cellcolorvalue{52.6} \\

\midrule
\multicolumn{20}{c}{Ensemble}\\
\midrule

Oracle & \cellcolorvalue{95.4} & \cellcolorvalue{97.8} & \cellcolorvalue{99.1} & \cellcolorvalue{87.0} & \cellcolorvalue{95.4} & \cellcolorvalue{76.2} & \cellcolorvalue{94.1} & \cellcolorvalue{97.5} & \cellcolorvalue{82.8} & \cellcolorvalue{95.4} & \cellcolorvalue{98.9} & \cellcolorvalue{92.4} & \cellcolorvalue{87.1} & \cellcolorvalue{96.7} & \cellcolorvalue{98.6} & \cellcolorvalue{98.3} & \cellcolorvalue{92.8} & \cellcolorvalue{14.4} & \cellcolorvalue{88.9} \\

\bottomrule
\end{tabular}
}
\caption{Normalized Levenshtein similarity scores of models evaluated on the \textbf{printed} subset of the \dataset test set (higher is better). Columns represent language clusters arranged alphabetically from left to right: Bangla, Bulgarian, Czech, German, English, Finnish, French, Hindi, Japanese, Latin, Dutch, Polish, Romanian, Sanskrit, Slovenian, Spanish, Swedish, Chinese, and the average across all language clusters. Within each system category, models are sorted by average similarity (highest to lowest).}
\label{tab:print_results}
\end{table*}

Table~\ref{tab:print_results} presents results for the printed subset of the \dataset test set. \textbf{\modelWithEmoji attains the highest average performance among all systems (82.3\%)}. Fine-tuning improves Qwen 2.5 VL (3B) across all languages by an average of 14.5\%, with the largest gains for Japanese (37.0\%), Finnish (32.9\%), and Swedish (22.9\%). Compared to Gemini 2.5 Pro, the much smaller \model performs better on 12 of 18 languages (with the largest gaps in Romanian, Japanese, and Finnish) and matches it on Czech. Low performance on Chinese print is attributable to the \dataset training set containing only 6 examples, underscoring the importance of training data size.

\textbf{Among proprietary VLMs, Gemini 2.5 Pro achieves the highest zero-shot performance (80.9\%)}, with other proprietary models performing between 7.2\% and 42.2\% worse.

\textbf{Among open-weight VLMs, OCR-specialized models perform best: NuMarkdown, olmOCR, and Nanonets OCR reach 72.7\%, 69.8\%, and 69.7\%, respectively.}

Among OCR systems, Azure OCR achieves the best average performance (71.9\%), falling 10.4\% short of \model. The hybrid model, combining Azure OCR with Gemini 2.5 Pro, averages 52.6\% and trails at least one of its components in every language. Relative to Azure OCR alone, the hybrid system improves notably for Bengali and Romanian.

We do not observe a clear correlation between model size and performance within model families. For example, the 3B-parameter Qwen 2.5 VL outperforms the 72B-parameter variant from the same family, mirroring trends with Mini models in the GPT-4.1 and GPT-5 families.

Overall, historical Chinese presents the greatest challenge; even the oracle ensemble achieves only 14.4\%. This subset includes many two-page newspaper images with vertical text, non-rectangular article boundaries, and interspersed Latin characters.

\subsection{Evaluation Results on Handwritten Texts}

\begin{table*}[ht!]
\centering
\small
\setlength{\tabcolsep}{3pt} 
\resizebox{\textwidth}{!}{
\begin{tabular}{l|rrrrrrrrrrrrrrrrrrrrrr}
\toprule

System & ara & cat & deu & ell & eng & fas & fra & heb & ita & jpn & khm & lat & nld & nor & por & san & spa & swe & tur & vie & zho & Avg \\

\midrule
\multicolumn{23}{c}{VLM Fine-Tuned on \dataset}\\
\midrule

\modelWithEmoji (3B) & \cellcolorvalue{77.5} & \cellcolorvalue{90.2} & \cellcolorvalue{83.1} & \cellcolorvalue{67.4} & \cellcolorvalue{84.0} & \cellcolorvalue{78.0} & \cellcolorvalue{81.1} & \cellcolorvalue{42.3} & \cellcolorvalue{88.4} & \cellcolorvalue{59.6} & \cellcolorvalue{25.7} & \cellcolorvalue{70.9} & \cellcolorvalue{83.8} & \cellcolorvalue{79.7} & \cellcolorvalue{52.9} & \cellcolorvalue{21.5} & \cellcolorvalue{78.0} & \cellcolorvalue{85.4} & \cellcolorvalue{69.1} & \cellcolorvalue{75.1} & \cellcolorvalue{78.2} & \cellcolorvalue{70.1} \\
$\Delta$ with Qwen 2.5 VL (3B) & \textcolor{DarkGreen}{+22.5} & \textcolor{DarkGreen}{+11.1} & \textcolor{DarkGreen}{+31.9} & \textcolor{DarkGreen}{+62.6} & \textcolor{DarkGreen}{+16.5} & \textcolor{DarkGreen}{+36.8} & \textcolor{DarkGreen}{+21.9} & \textcolor{DarkGreen}{+42.1} & \textcolor{DarkGreen}{+14.8} & \textcolor{DarkGreen}{+54.7} & \textcolor{DarkGreen}{+23.4} & \textcolor{DarkGreen}{+19.2} & \textcolor{DarkGreen}{+19.8} & \textcolor{DarkGreen}{+12.8} & \textcolor{DarkGreen}{+16.8} & \textcolor{DarkGreen}{+21.4} & \textcolor{DarkGreen}{+26.5} & \textcolor{DarkGreen}{+32.9} & \textcolor{DarkGreen}{+42.3} & \textcolor{DarkGreen}{+16.6} & \textcolor{DarkGreen}{+25.4} & \textcolor{DarkGreen}{+27.2} \\

\midrule
\multicolumn{23}{c}{Closed VLMs (Zero-Shot)}\\
\midrule

Gemini 2.5 Pro & \cellcolorvalue{87.0} & \cellcolorvalue{87.9} & \cellcolorvalue{73.9} & \cellcolorvalue{54.6} & \cellcolorvalue{80.0} & \cellcolorvalue{81.6} & \cellcolorvalue{74.5} & \cellcolorvalue{30.4} & \cellcolorvalue{88.0} & \cellcolorvalue{18.0} & \cellcolorvalue{24.6} & \cellcolorvalue{66.5} & \cellcolorvalue{78.6} & \cellcolorvalue{87.7} & \cellcolorvalue{45.9} & \cellcolorvalue{15.8} & \cellcolorvalue{75.5} & \cellcolorvalue{76.1} & \cellcolorvalue{45.7} & \cellcolorvalue{69.3} & \cellcolorvalue{74.6} & \cellcolorvalue{63.6} \\
Gemini 2.5 Flash & \cellcolorvalue{80.7} & \cellcolorvalue{80.4} & \cellcolorvalue{69.4} & \cellcolorvalue{47.1} & \cellcolorvalue{77.2} & \cellcolorvalue{76.6} & \cellcolorvalue{71.6} & \cellcolorvalue{29.0} & \cellcolorvalue{84.3} & \cellcolorvalue{12.9} & \cellcolorvalue{27.2} & \cellcolorvalue{57.4} & \cellcolorvalue{73.9} & \cellcolorvalue{81.2} & \cellcolorvalue{38.1} & \cellcolorvalue{11.4} & \cellcolorvalue{66.3} & \cellcolorvalue{68.7} & \cellcolorvalue{42.8} & \cellcolorvalue{66.3} & \cellcolorvalue{71.0} & \cellcolorvalue{58.7} \\
GPT-4.1 Mini & \cellcolorvalue{60.6} & \cellcolorvalue{78.0} & \cellcolorvalue{55.7} & \cellcolorvalue{38.3} & \cellcolorvalue{73.9} & \cellcolorvalue{50.7} & \cellcolorvalue{64.8} & \cellcolorvalue{28.0} & \cellcolorvalue{79.7} & \cellcolorvalue{14.7} & \cellcolorvalue{12.7} & \cellcolorvalue{57.1} & \cellcolorvalue{66.0} & \cellcolorvalue{77.9} & \cellcolorvalue{36.4} & \cellcolorvalue{10.2} & \cellcolorvalue{61.0} & \cellcolorvalue{57.9} & \cellcolorvalue{31.2} & \cellcolorvalue{38.3} & \cellcolorvalue{62.1} & \cellcolorvalue{50.2} \\
GPT-5 Mini & \cellcolorvalue{54.2} & \cellcolorvalue{78.4} & \cellcolorvalue{56.3} & \cellcolorvalue{34.0} & \cellcolorvalue{76.8} & \cellcolorvalue{45.8} & \cellcolorvalue{66.0} & \cellcolorvalue{25.9} & \cellcolorvalue{75.5} & \cellcolorvalue{12.2} & \cellcolorvalue{14.8} & \cellcolorvalue{51.8} & \cellcolorvalue{60.1} & \cellcolorvalue{78.6} & \cellcolorvalue{37.4} & \cellcolorvalue{9.2} & \cellcolorvalue{59.7} & \cellcolorvalue{49.6} & \cellcolorvalue{30.5} & \cellcolorvalue{32.5} & \cellcolorvalue{48.5} & \cellcolorvalue{47.5} \\
Claude Sonnet 3.7 & \cellcolorvalue{61.2} & \cellcolorvalue{83.2} & \cellcolorvalue{58.6} & \cellcolorvalue{50.4} & \cellcolorvalue{75.8} & \cellcolorvalue{42.8} & \cellcolorvalue{63.9} & \cellcolorvalue{29.6} & \cellcolorvalue{74.8} & \cellcolorvalue{2.4} & \cellcolorvalue{8.2} & \cellcolorvalue{58.0} & \cellcolorvalue{64.4} & \cellcolorvalue{75.2} & \cellcolorvalue{41.8} & \cellcolorvalue{6.5} & \cellcolorvalue{60.7} & \cellcolorvalue{49.7} & \cellcolorvalue{31.6} & \cellcolorvalue{11.1} & \cellcolorvalue{28.3} & \cellcolorvalue{46.6} \\
O4 Mini & \cellcolorvalue{48.6} & \cellcolorvalue{73.1} & \cellcolorvalue{48.9} & \cellcolorvalue{26.8} & \cellcolorvalue{74.3} & \cellcolorvalue{38.0} & \cellcolorvalue{59.7} & \cellcolorvalue{9.8} & \cellcolorvalue{73.3} & \cellcolorvalue{6.3} & \cellcolorvalue{8.1} & \cellcolorvalue{47.6} & \cellcolorvalue{55.2} & \cellcolorvalue{72.3} & \cellcolorvalue{30.8} & \cellcolorvalue{6.7} & \cellcolorvalue{53.3} & \cellcolorvalue{39.2} & \cellcolorvalue{26.3} & \cellcolorvalue{28.7} & \cellcolorvalue{44.0} & \cellcolorvalue{41.5} \\
GPT-4.1 & \cellcolorvalue{50.7} & \cellcolorvalue{66.9} & \cellcolorvalue{48.9} & \cellcolorvalue{28.3} & \cellcolorvalue{70.6} & \cellcolorvalue{42.6} & \cellcolorvalue{57.2} & \cellcolorvalue{21.7} & \cellcolorvalue{68.8} & \cellcolorvalue{11.1} & \cellcolorvalue{10.6} & \cellcolorvalue{48.0} & \cellcolorvalue{51.7} & \cellcolorvalue{71.4} & \cellcolorvalue{34.4} & \cellcolorvalue{9.6} & \cellcolorvalue{49.8} & \cellcolorvalue{38.8} & \cellcolorvalue{31.6} & \cellcolorvalue{25.4} & \cellcolorvalue{31.5} & \cellcolorvalue{41.4} \\
Claude Opus 4.1 & \cellcolorvalue{48.1} & \cellcolorvalue{76.4} & \cellcolorvalue{52.3} & \cellcolorvalue{30.7} & \cellcolorvalue{71.6} & \cellcolorvalue{39.2} & \cellcolorvalue{59.0} & \cellcolorvalue{23.9} & \cellcolorvalue{69.8} & \cellcolorvalue{7.3} & \cellcolorvalue{2.2} & \cellcolorvalue{50.5} & \cellcolorvalue{57.2} & \cellcolorvalue{66.6} & \cellcolorvalue{33.5} & \cellcolorvalue{9.2} & \cellcolorvalue{55.4} & \cellcolorvalue{39.1} & \cellcolorvalue{25.6} & \cellcolorvalue{7.5} & \cellcolorvalue{18.6} & \cellcolorvalue{40.2} \\
Claude Sonnet 4 & \cellcolorvalue{45.3} & \cellcolorvalue{70.0} & \cellcolorvalue{48.6} & \cellcolorvalue{30.4} & \cellcolorvalue{63.0} & \cellcolorvalue{33.0} & \cellcolorvalue{49.6} & \cellcolorvalue{29.3} & \cellcolorvalue{64.3} & \cellcolorvalue{9.6} & \cellcolorvalue{8.0} & \cellcolorvalue{46.5} & \cellcolorvalue{46.4} & \cellcolorvalue{54.5} & \cellcolorvalue{34.7} & \cellcolorvalue{8.6} & \cellcolorvalue{52.1} & \cellcolorvalue{30.8} & \cellcolorvalue{27.1} & \cellcolorvalue{6.0} & \cellcolorvalue{20.5} & \cellcolorvalue{37.1} \\
O1 & \cellcolorvalue{43.1} & \cellcolorvalue{63.9} & \cellcolorvalue{44.2} & \cellcolorvalue{21.3} & \cellcolorvalue{67.7} & \cellcolorvalue{37.7} & \cellcolorvalue{49.0} & \cellcolorvalue{6.0} & \cellcolorvalue{64.7} & \cellcolorvalue{3.4} & \cellcolorvalue{0.5} & \cellcolorvalue{39.1} & \cellcolorvalue{48.0} & \cellcolorvalue{67.3} & \cellcolorvalue{24.0} & \cellcolorvalue{2.9} & \cellcolorvalue{46.8} & \cellcolorvalue{38.3} & \cellcolorvalue{25.8} & \cellcolorvalue{18.5} & \cellcolorvalue{24.7} & \cellcolorvalue{35.1} \\
GPT-4o & \cellcolorvalue{45.1} & \cellcolorvalue{56.8} & \cellcolorvalue{39.1} & \cellcolorvalue{24.1} & \cellcolorvalue{65.5} & \cellcolorvalue{30.3} & \cellcolorvalue{48.9} & \cellcolorvalue{18.4} & \cellcolorvalue{55.2} & \cellcolorvalue{9.2} & \cellcolorvalue{7.1} & \cellcolorvalue{37.9} & \cellcolorvalue{42.8} & \cellcolorvalue{60.9} & \cellcolorvalue{27.5} & \cellcolorvalue{8.8} & \cellcolorvalue{41.2} & \cellcolorvalue{31.7} & \cellcolorvalue{26.2} & \cellcolorvalue{17.9} & \cellcolorvalue{23.6} & \cellcolorvalue{34.2} \\
O3 & \cellcolorvalue{32.3} & \cellcolorvalue{64.5} & \cellcolorvalue{41.5} & \cellcolorvalue{19.8} & \cellcolorvalue{71.7} & \cellcolorvalue{12.2} & \cellcolorvalue{43.2} & \cellcolorvalue{8.6} & \cellcolorvalue{62.3} & \cellcolorvalue{3.3} & \cellcolorvalue{0.1} & \cellcolorvalue{31.7} & \cellcolorvalue{36.2} & \cellcolorvalue{71.3} & \cellcolorvalue{13.8} & \cellcolorvalue{0.2} & \cellcolorvalue{47.0} & \cellcolorvalue{22.5} & \cellcolorvalue{24.6} & \cellcolorvalue{13.5} & \cellcolorvalue{28.7} & \cellcolorvalue{30.9} \\
GPT-5 & \cellcolorvalue{38.4} & \cellcolorvalue{64.3} & \cellcolorvalue{35.4} & \cellcolorvalue{19.3} & \cellcolorvalue{70.6} & \cellcolorvalue{16.7} & \cellcolorvalue{44.7} & \cellcolorvalue{8.1} & \cellcolorvalue{63.6} & \cellcolorvalue{3.8} & \cellcolorvalue{1.1} & \cellcolorvalue{29.6} & \cellcolorvalue{36.8} & \cellcolorvalue{67.3} & \cellcolorvalue{18.4} & \cellcolorvalue{0.8} & \cellcolorvalue{41.2} & \cellcolorvalue{26.3} & \cellcolorvalue{19.1} & \cellcolorvalue{5.4} & \cellcolorvalue{27.8} & \cellcolorvalue{30.4} \\
GPT-4o Mini & \cellcolorvalue{32.0} & \cellcolorvalue{49.1} & \cellcolorvalue{32.4} & \cellcolorvalue{22.3} & \cellcolorvalue{64.0} & \cellcolorvalue{28.0} & \cellcolorvalue{42.4} & \cellcolorvalue{14.6} & \cellcolorvalue{51.3} & \cellcolorvalue{9.7} & \cellcolorvalue{3.1} & \cellcolorvalue{33.2} & \cellcolorvalue{37.2} & \cellcolorvalue{53.8} & \cellcolorvalue{24.7} & \cellcolorvalue{8.6} & \cellcolorvalue{38.4} & \cellcolorvalue{29.7} & \cellcolorvalue{24.7} & \cellcolorvalue{10.6} & \cellcolorvalue{15.4} & \cellcolorvalue{29.8} \\
GPT-4.1 Nano & \cellcolorvalue{26.2} & \cellcolorvalue{50.4} & \cellcolorvalue{33.9} & \cellcolorvalue{22.1} & \cellcolorvalue{59.9} & \cellcolorvalue{23.2} & \cellcolorvalue{42.9} & \cellcolorvalue{12.6} & \cellcolorvalue{53.8} & \cellcolorvalue{8.8} & \cellcolorvalue{9.2} & \cellcolorvalue{33.0} & \cellcolorvalue{35.3} & \cellcolorvalue{54.6} & \cellcolorvalue{20.4} & \cellcolorvalue{4.3} & \cellcolorvalue{34.1} & \cellcolorvalue{25.5} & \cellcolorvalue{18.2} & \cellcolorvalue{12.4} & \cellcolorvalue{14.5} & \cellcolorvalue{28.3} \\
GPT-5 Nano & \cellcolorvalue{8.7} & \cellcolorvalue{24.9} & \cellcolorvalue{18.8} & \cellcolorvalue{4.4} & \cellcolorvalue{50.5} & \cellcolorvalue{7.2} & \cellcolorvalue{28.1} & \cellcolorvalue{1.5} & \cellcolorvalue{30.6} & \cellcolorvalue{2.7} & \cellcolorvalue{1.2} & \cellcolorvalue{16.1} & \cellcolorvalue{14.4} & \cellcolorvalue{41.3} & \cellcolorvalue{6.4} & \cellcolorvalue{0.9} & \cellcolorvalue{16.8} & \cellcolorvalue{5.6} & \cellcolorvalue{4.4} & \cellcolorvalue{1.3} & \cellcolorvalue{11.6} & \cellcolorvalue{14.2} \\

\midrule
\multicolumn{23}{c}{Open-Weight VLMs (Zero-Shot)}\\
\midrule

Qwen 2.5 VL (72B) & \cellcolorvalue{67.8} & \cellcolorvalue{85.7} & \cellcolorvalue{67.0} & \cellcolorvalue{38.7} & \cellcolorvalue{77.0} & \cellcolorvalue{55.9} & \cellcolorvalue{75.1} & \cellcolorvalue{16.3} & \cellcolorvalue{80.4} & \cellcolorvalue{11.9} & \cellcolorvalue{0.5} & \cellcolorvalue{56.1} & \cellcolorvalue{74.1} & \cellcolorvalue{74.8} & \cellcolorvalue{43.8} & \cellcolorvalue{13.3} & \cellcolorvalue{67.8} & \cellcolorvalue{74.8} & \cellcolorvalue{36.8} & \cellcolorvalue{68.1} & \cellcolorvalue{58.0} & \cellcolorvalue{54.5} \\
NuMarkdown (8B) & \cellcolorvalue{55.9} & \cellcolorvalue{81.0} & \cellcolorvalue{64.7} & \cellcolorvalue{24.6} & \cellcolorvalue{70.2} & \cellcolorvalue{48.1} & \cellcolorvalue{67.4} & \cellcolorvalue{17.7} & \cellcolorvalue{85.2} & \cellcolorvalue{11.0} & \cellcolorvalue{11.9} & \cellcolorvalue{61.4} & \cellcolorvalue{72.6} & \cellcolorvalue{68.5} & \cellcolorvalue{41.2} & \cellcolorvalue{10.3} & \cellcolorvalue{60.6} & \cellcolorvalue{69.2} & \cellcolorvalue{16.4} & \cellcolorvalue{71.5} & \cellcolorvalue{66.3} & \cellcolorvalue{51.2} \\
RolmOCR (8B) & \cellcolorvalue{51.8} & \cellcolorvalue{81.9} & \cellcolorvalue{63.9} & \cellcolorvalue{14.9} & \cellcolorvalue{74.1} & \cellcolorvalue{50.9} & \cellcolorvalue{67.0} & \cellcolorvalue{2.8} & \cellcolorvalue{84.3} & \cellcolorvalue{7.9} & \cellcolorvalue{0.6} & \cellcolorvalue{58.6} & \cellcolorvalue{69.1} & \cellcolorvalue{72.3} & \cellcolorvalue{37.9} & \cellcolorvalue{9.2} & \cellcolorvalue{60.7} & \cellcolorvalue{69.2} & \cellcolorvalue{21.1} & \cellcolorvalue{66.6} & \cellcolorvalue{64.5} & \cellcolorvalue{49.0} \\
Nanonets OCR (3B) & \cellcolorvalue{52.0} & \cellcolorvalue{77.8} & \cellcolorvalue{56.0} & \cellcolorvalue{20.4} & \cellcolorvalue{62.3} & \cellcolorvalue{17.3} & \cellcolorvalue{61.6} & \cellcolorvalue{0.8} & \cellcolorvalue{77.5} & \cellcolorvalue{4.3} & \cellcolorvalue{0.3} & \cellcolorvalue{56.4} & \cellcolorvalue{64.9} & \cellcolorvalue{67.9} & \cellcolorvalue{36.2} & \cellcolorvalue{2.0} & \cellcolorvalue{57.3} & \cellcolorvalue{63.2} & \cellcolorvalue{13.9} & \cellcolorvalue{57.6} & \cellcolorvalue{56.8} & \cellcolorvalue{43.2} \\
Qwen 2.5 VL (3B) & \cellcolorvalue{55.0} & \cellcolorvalue{79.1} & \cellcolorvalue{51.2} & \cellcolorvalue{4.8} & \cellcolorvalue{67.5} & \cellcolorvalue{41.2} & \cellcolorvalue{59.2} & \cellcolorvalue{0.2} & \cellcolorvalue{73.6} & \cellcolorvalue{5.0} & \cellcolorvalue{2.2} & \cellcolorvalue{51.7} & \cellcolorvalue{64.1} & \cellcolorvalue{66.9} & \cellcolorvalue{36.0} & \cellcolorvalue{0.2} & \cellcolorvalue{51.6} & \cellcolorvalue{52.5} & \cellcolorvalue{26.7} & \cellcolorvalue{58.6} & \cellcolorvalue{52.8} & \cellcolorvalue{42.9} \\
olmOCR (8B) & \cellcolorvalue{44.0} & \cellcolorvalue{80.6} & \cellcolorvalue{53.4} & \cellcolorvalue{11.7} & \cellcolorvalue{68.0} & \cellcolorvalue{23.0} & \cellcolorvalue{60.5} & \cellcolorvalue{4.3} & \cellcolorvalue{79.4} & \cellcolorvalue{3.9} & \cellcolorvalue{0.4} & \cellcolorvalue{52.2} & \cellcolorvalue{64.6} & \cellcolorvalue{67.2} & \cellcolorvalue{21.5} & \cellcolorvalue{0.5} & \cellcolorvalue{48.0} & \cellcolorvalue{60.6} & \cellcolorvalue{13.1} & \cellcolorvalue{60.0} & \cellcolorvalue{55.0} & \cellcolorvalue{41.5} \\
MiMo VL (8B RL) & \cellcolorvalue{16.6} & \cellcolorvalue{76.1} & \cellcolorvalue{43.9} & \cellcolorvalue{16.3} & \cellcolorvalue{63.0} & \cellcolorvalue{30.7} & \cellcolorvalue{44.8} & \cellcolorvalue{1.4} & \cellcolorvalue{63.1} & \cellcolorvalue{4.1} & \cellcolorvalue{8.2} & \cellcolorvalue{33.8} & \cellcolorvalue{57.6} & \cellcolorvalue{57.2} & \cellcolorvalue{29.2} & \cellcolorvalue{8.1} & \cellcolorvalue{46.6} & \cellcolorvalue{33.3} & \cellcolorvalue{16.9} & \cellcolorvalue{44.5} & \cellcolorvalue{31.6} & \cellcolorvalue{34.6} \\
Gemma 3 (27B) & \cellcolorvalue{35.6} & \cellcolorvalue{53.4} & \cellcolorvalue{37.9} & \cellcolorvalue{29.8} & \cellcolorvalue{62.1} & \cellcolorvalue{33.4} & \cellcolorvalue{46.9} & \cellcolorvalue{22.7} & \cellcolorvalue{53.4} & \cellcolorvalue{9.4} & \cellcolorvalue{9.1} & \cellcolorvalue{40.8} & \cellcolorvalue{42.4} & \cellcolorvalue{56.0} & \cellcolorvalue{33.2} & \cellcolorvalue{8.8} & \cellcolorvalue{41.7} & \cellcolorvalue{28.6} & \cellcolorvalue{24.2} & \cellcolorvalue{20.4} & \cellcolorvalue{26.5} & \cellcolorvalue{34.1} \\
InternVL 3.5 (30B) & \cellcolorvalue{8.3} & \cellcolorvalue{44.4} & \cellcolorvalue{23.1} & \cellcolorvalue{17.2} & \cellcolorvalue{52.4} & \cellcolorvalue{9.6} & \cellcolorvalue{32.7} & \cellcolorvalue{6.3} & \cellcolorvalue{41.2} & \cellcolorvalue{5.6} & \cellcolorvalue{6.9} & \cellcolorvalue{31.2} & \cellcolorvalue{27.5} & \cellcolorvalue{40.5} & \cellcolorvalue{20.2} & \cellcolorvalue{1.8} & \cellcolorvalue{24.9} & \cellcolorvalue{21.4} & \cellcolorvalue{11.6} & \cellcolorvalue{64.4} & \cellcolorvalue{62.5} & \cellcolorvalue{26.4} \\
Skywork R1V3 (38B) & \cellcolorvalue{14.5} & \cellcolorvalue{45.2} & \cellcolorvalue{26.0} & \cellcolorvalue{16.1} & \cellcolorvalue{50.6} & \cellcolorvalue{16.6} & \cellcolorvalue{36.9} & \cellcolorvalue{6.0} & \cellcolorvalue{37.8} & \cellcolorvalue{5.3} & \cellcolorvalue{7.0} & \cellcolorvalue{33.4} & \cellcolorvalue{30.7} & \cellcolorvalue{36.4} & \cellcolorvalue{25.1} & \cellcolorvalue{6.7} & \cellcolorvalue{31.9} & \cellcolorvalue{20.5} & \cellcolorvalue{14.7} & \cellcolorvalue{35.4} & \cellcolorvalue{41.0} & \cellcolorvalue{25.6} \\
R (4B) & \cellcolorvalue{17.7} & \cellcolorvalue{30.0} & \cellcolorvalue{22.7} & \cellcolorvalue{11.9} & \cellcolorvalue{41.5} & \cellcolorvalue{18.8} & \cellcolorvalue{30.0} & \cellcolorvalue{8.9} & \cellcolorvalue{30.5} & \cellcolorvalue{5.2} & \cellcolorvalue{3.6} & \cellcolorvalue{28.6} & \cellcolorvalue{24.8} & \cellcolorvalue{30.5} & \cellcolorvalue{21.2} & \cellcolorvalue{5.1} & \cellcolorvalue{23.6} & \cellcolorvalue{21.1} & \cellcolorvalue{13.0} & \cellcolorvalue{33.0} & \cellcolorvalue{34.9} & \cellcolorvalue{21.7} \\
Nemotron Nano VL (8B) & \cellcolorvalue{6.2} & \cellcolorvalue{28.5} & \cellcolorvalue{21.1} & \cellcolorvalue{4.2} & \cellcolorvalue{38.3} & \cellcolorvalue{7.7} & \cellcolorvalue{26.0} & \cellcolorvalue{3.5} & \cellcolorvalue{34.0} & \cellcolorvalue{0.9} & \cellcolorvalue{2.9} & \cellcolorvalue{23.5} & \cellcolorvalue{22.7} & \cellcolorvalue{37.0} & \cellcolorvalue{17.1} & \cellcolorvalue{2.0} & \cellcolorvalue{23.0} & \cellcolorvalue{15.7} & \cellcolorvalue{4.3} & \cellcolorvalue{1.0} & \cellcolorvalue{0.8} & \cellcolorvalue{15.3} \\

Phi 4 Multimodal (5B) & \cellcolorvalue{1.1} & \cellcolorvalue{9.1} & \cellcolorvalue{8.6} & \cellcolorvalue{2.6} & \cellcolorvalue{20.8} & \cellcolorvalue{1.0} & \cellcolorvalue{6.5} & \cellcolorvalue{0.9} & \cellcolorvalue{12.5} & \cellcolorvalue{0.7} & \cellcolorvalue{1.5} & \cellcolorvalue{8.1} & \cellcolorvalue{5.2} & \cellcolorvalue{13.3} & \cellcolorvalue{5.9} & \cellcolorvalue{0.6} & \cellcolorvalue{4.5} & \cellcolorvalue{5.4} & \cellcolorvalue{0.9} & \cellcolorvalue{1.5} & \cellcolorvalue{1.3} & \cellcolorvalue{5.3} \\

\midrule
\multicolumn{23}{c}{OCR Systems}\\
\midrule

Azure OCR & \cellcolorvalue{74.3} & \cellcolorvalue{72.7} & \cellcolorvalue{49.4} & \cellcolorvalue{20.5} & \cellcolorvalue{73.5} & \cellcolorvalue{45.4} & \cellcolorvalue{60.6} & \cellcolorvalue{38.6} & \cellcolorvalue{70.9} & \cellcolorvalue{11.4} & \cellcolorvalue{3.7} & \cellcolorvalue{53.0} & \cellcolorvalue{58.0} & \cellcolorvalue{64.1} & \cellcolorvalue{37.7} & \cellcolorvalue{14.6} & \cellcolorvalue{62.2} & \cellcolorvalue{46.5} & \cellcolorvalue{39.8} & \cellcolorvalue{55.1} & \cellcolorvalue{50.6} & \cellcolorvalue{47.7} \\
Azure OCR + Gemini 2.5 Pro & \cellcolorvalue{59.2} & \cellcolorvalue{64.9} & \cellcolorvalue{47.1} & \cellcolorvalue{32.4} & \cellcolorvalue{53.9} & \cellcolorvalue{49.9} & \cellcolorvalue{55.3} & \cellcolorvalue{18.0} & \cellcolorvalue{68.4} & \cellcolorvalue{5.8} & \cellcolorvalue{20.4} & \cellcolorvalue{41.6} & \cellcolorvalue{32.0} & \cellcolorvalue{56.5} & \cellcolorvalue{29.4} & \cellcolorvalue{11.9} & \cellcolorvalue{44.5} & \cellcolorvalue{57.2} & \cellcolorvalue{31.6} & \cellcolorvalue{48.8} & \cellcolorvalue{20.1} & \cellcolorvalue{40.4} \\
Mistral OCR & \cellcolorvalue{60.3} & \cellcolorvalue{56.3} & \cellcolorvalue{29.0} & \cellcolorvalue{5.8} & \cellcolorvalue{58.6} & \cellcolorvalue{46.2} & \cellcolorvalue{38.7} & \cellcolorvalue{2.7} & \cellcolorvalue{62.8} & \cellcolorvalue{3.8} & \cellcolorvalue{3.6} & \cellcolorvalue{35.6} & \cellcolorvalue{33.0} & \cellcolorvalue{52.4} & \cellcolorvalue{14.0} & \cellcolorvalue{1.7} & \cellcolorvalue{37.4} & \cellcolorvalue{32.1} & \cellcolorvalue{37.5} & \cellcolorvalue{2.4} & \cellcolorvalue{4.4} & \cellcolorvalue{29.4} \\

\midrule
\multicolumn{23}{c}{Ensemble}\\
\midrule

Oracle & \cellcolorvalue{88.8} & \cellcolorvalue{92.6} & \cellcolorvalue{89.4} & \cellcolorvalue{74.9} & \cellcolorvalue{89.7} & \cellcolorvalue{83.5} & \cellcolorvalue{88.0} & \cellcolorvalue{59.1} & \cellcolorvalue{93.1} & \cellcolorvalue{65.4} & \cellcolorvalue{32.9} & \cellcolorvalue{76.2} & \cellcolorvalue{87.8} & \cellcolorvalue{89.3} & \cellcolorvalue{57.8} & \cellcolorvalue{26.1} & \cellcolorvalue{82.6} & \cellcolorvalue{88.9} & \cellcolorvalue{74.9} & \cellcolorvalue{78.5} & \cellcolorvalue{82.9} & \cellcolorvalue{76.3} \\

\bottomrule
\end{tabular}
}

\caption{Normalized Levenshtein similarity scores of models evaluated on the \textbf{handwritten} subset of the \dataset test set (higher is better). Columns represent language clusters arranged alphabetically from left to right: Arabic, Catalan, German, Greek, English, Persian, French, Hebrew, Italian, Japanese, Khmer, Latin, Dutch, Norwegian, Portuguese, Sanskrit, Spanish, Swedish, Turkish, Vietnamese, Chinese, and the average across all language clusters. Within each system category, models are sorted by average similarity (highest to lowest).}

\label{tab:handwriting_results}
\end{table*}

Table~\ref{tab:handwriting_results} presents evaluation results on the handwritten subset of the \dataset test set. \textbf{\modelWithEmoji achieves the highest average performance among all models (70.1\%)}, improving upon the original Qwen 2.5 VL (3B) by 27.2\%. The largest gains from fine-tuning are observed for Greek (62.6\%), Japanese (54.7\%), Turkish (42.3\%), Hebrew (42.1\%), and Persian (36.8\%). Compared to Gemini 2.5 Pro, the fine-tuned model performs better on 18 of 21 languages, trailing only on Arabic, Persian, and Norwegian.

\textbf{Among proprietary models, Gemini 2.5 Pro achieves the highest average similarity (63.6\%)}. Notably, the performance gap between Gemini models and other proprietary VLMs is considerably larger for handwritten documents than for printed ones.

\textbf{Among open-weight VLMs, Qwen 2.5 VL (72B) leads with an average performance of 54.5\%}. Smaller, specialized OCR VLMs—NuMarkdown, RolmOCR, and Nanonets OCR—follow with 51.2\%, 49.0\%, and 43.2\%, respectively.

As in the printed setting, the hybrid system combining Azure OCR with Gemini 2.5 Pro does not outperform its individual components.

All models struggle particularly with Sanskrit, Khmer, and Hebrew; even the oracle ensemble achieves only 26.1\%, 32.9\%, and 59.1\% on these languages, respectively. Greek and Turkish are also challenging for all zero-shot models we evaluated.

Across both handwritten and printed subsets, multiple VLMs outperform OCR systems, underscoring the promise of VLMs for page-level OCR over pipeline methods.

\subsection{Error Analysis}
\label{sec:error-analysis}

\textbf{Reading Order Errors.} In a sample of 50 predictions, 42\% from zero-shot Qwen 2.5 VL (3B) contain major reading-order errors, compared to 16\% for \model.

The most common issue involves column ordering: many \dataset pages have two or more columns, which some models mishandle. Reading-order errors are also more frequent in East Asian scripts, where the zero-shot model often fails to recognize top-to-bottom writing direction, as illustrated in Figure~\ref{fig:chinese-reading-order}. \model performs substantially better in this regard.

\textbf{Major Hallucinations.}
Thirty-six percent of the 50 predictions from the zero-shot Qwen 2.5 VL (3B) exhibit major hallucinations. These arise when a model cannot reliably recognize the text and instead generates content that seems contextually plausible. Figure~\ref{fig:dutch-letter-bound} shows an 18th-century Dutch letter where the model correctly identifies the document as a letter but, unable to transcribe its content, produces text such as ``Dit is een brief'' (``This is a letter.''). None of the predictions from \model exhibit hallucinations.

\textbf{Errors in Gold Annotations.}
We found that 2 of the 50 examples contain gold-standard text that omits parts of the page. These omissions stem from annotation errors in the original datasets that persisted through our cleaning process. For example, the transcription in Figure~\ref{fig:middle-high-german} begins halfway through the left column.

\textbf{Other Minor Errors.}
We observe recurring minor errors across languages and periods. Models often confuse visually similar characters, such as transcribing `o' for `ø'. These errors frequently occur in proper names. In Figure~\ref{fig:norwegian-letter}, the model renders ``Suso'' as ``Süss'' and ``Birgittiner'' as ``Bisittiner''. Scan quality can also impact accuracy, as in Figure~\ref{fig:dutch-letter-bound}, where a letter bound in a volume leads to many character errors near the spine due to page curvature.

Appendix~\ref{appendix:example_output} provides additional examples and error analyses.

\section{Conclusion}

In this paper, we introduced \model, a compact vision-language model fine-tuned from Qwen 2.5 VL for historical text recognition. To support its training and evaluation, we curated \dataset, the largest and most diverse human-annotated text recognition dataset  historical text recognition assembled to date. It unifies 155 corpora and nearly 100,000 pages spanning 22 centuries across 46 language clusters. Its diversity in languages, scripts, layouts, and formats makes it a powerful resource for benchmarking and improving VLMs on historical texts.

\dataset reveals the difficulty of the task: Gemini 2.5 Pro, the best-performing commercial VLM, achieves only 80.9\% (printed) and 63.6\% (handwritten) normalized Levenshtein similarity. Fine-tuned on \dataset, \model achieves 82.3\% and 70.1\%, surpassing Gemini 2.5 Pro while being 15.5 times more cost-effective. These results represent improvements of 14.5\% and 27.2\% over the base model, highlighting the benefits of training on a diverse historical corpus.

Even imperfect model outputs can aid in digitization of documents as valuable first drafts, substantially reducing the effort required for scholarly correction. Such transcriptions make historical documents accessible to a broader audience, including readers without specialized training in historical scripts.

\section*{Limitations}
Although \dataset is by far the most diverse resource for historical text recognition to date, it still underrepresents some languages. This imbalance reflects the scarcity of annotated historical datasets for many lower-resource languages. Notably, \dataset currently does not include any languages native to the African continent.

Moreover, we only experimented with standard supervised fine-tuning. Future work should explore more advanced training methods in this setting.

\section*{Ethical Considerations}
No crowdsourcing was conducted in this research. All included datasets have licenses permitting research use (see Appendix~\ref{appendix:dataset-list} for detailed license information). The \dataset dataset is released under the \emph{Creative Commons Attribution Share Alike 4.0} license to comply with these licenses. We do not anticipate any potential harm arising from the use of this dataset.

\section*{Acknowledgement}
We thank Professor Trevor Getz for valuable discussions.

This work is supported in part by the Verdant Foundation, the Alfred P. Sloan Foundation, Google, Stanford Human-Centered Artificial Intelligence (HAI) Institute, David and Helen Gurley Brown Institute for Media Innovation, Microsoft Azure AI credits, and the NAIRR Pilot program.

\bibliography{anthology,custom}

\begin{thebibliography}{240}
\providecommand{\natexlab}[1]{#1}

\bibitem[{ald(2023)}]{aldicam}
 2023.
\newblock Aldicam-2023: Diachronic and interactive linguistic atlas of the community of madrid.
\newblock \url{https://aldicam.corpuscodea.es/consultas.php}.
\newblock Accessed: 2025-05-18.

\bibitem[{osm(2025)}]{osmanagha}
 2025.
\newblock Osman agha dataset: Images and transcription from british library ms. or. 3213 and kreutel's 1980 edition.
\newblock Based on London, British Library, MS. Or. 3213 and Richard F. Kreutel's 1980 transcription: *Die Autobiographie des Dolmetschers Osman Aga aus Temeschwar*, The E. J. W. Gibb Memorial Trust, Cambridge.

\bibitem[{Abadie et~al.(2022)Abadie, Bacciochi, Carlinet, Chazalon, Cristofoli, Dum{\'e}nieu, and Perret}]{abadie_dataset_22}
Nathalie Abadie, St{\'e}phane Bacciochi, Edwin Carlinet, Joseph Chazalon, Pascal Cristofoli, Bertrand Dum{\'e}nieu, and Julien Perret. 2022.
\newblock \href {https://doi.org/10.5281/zenodo.6394464} {{A} {D}ataset of {F}rench {T}rade {D}irectories from the 19th {C}entury ({FTD})}.

\bibitem[{Agency(2007)}]{iso639-3}
ISO 639-3: Language~Coding Agency. 2007.
\newblock Iso 639-3.
\newblock \url{https://iso639-3.sil.org/}.
\newblock Accessed: 2025-03-25.

\bibitem[{Agostini(2024)}]{agostini2024lidi}
Giorgia Agostini. 2024.
\newblock \href {https://doi.org/10.5281/zenodo.12639497} {giorgiaagostini/lidi1.0-project: Gt upload}.

\bibitem[{AI(2025)}]{RolmOCR}
Reducto AI. 2025.
\newblock Rolmocr: A faster, lighter open source ocr model.

\bibitem[{Ainonen et~al.(2022)Ainonen, Andresen, Bakker, Boylan, Manna, Dziemski, Henderson et~al.}]{ainonen2022onb3891}
Tuija Ainonen, Suse Andresen, Loïs Bakker, Amy Boylan, Silvia~Della Manna, Wiktor Dziemski, C.~E.~M. Henderson, and 1 others. 2022.
\newblock \href {https://doi.org/10.5281/zenodo.7467249} {Ground truth for Önb, cod. 3891}.

\bibitem[{Alba et~al.(2023)Alba, Rubin, Boschetti, Fischer, Clérice, and Chagué}]{alba2023htromance}
Rachele Alba, Giorgia Rubin, Federico Boschetti, Franz Fischer, Thibault Clérice, and Alix Chagué. 2023.
\newblock \href {https://doi.org/10.5281/zenodo.8256728} {Htromance: Medieval italian corpus of ground-truth for handwritten text recognition and layout segmentation}.
\newblock If you use this dataset, please cite it using the metadata from this file.

\bibitem[{Alexandre et~al.(2025)Alexandre, Rui, Gonçalo, Catarina, and Pedro}]{htr_united_https-githubcom-arch-w-iforal-dataset}
Matos Alexandre, Neves Rui, Monteiro Gonçalo, Coelho Catarina, and Bastos Pedro. 2025.
\newblock \href {https://github.com/Arch-W/iForal-Dataset} {iforal-dataset}.

\bibitem[{AlKendi et~al.(2024)AlKendi, Gechter, Heyberger, and Guyeux}]{jimaging10010018}
Wissam AlKendi, Franck Gechter, Laurent Heyberger, and Christophe Guyeux. 2024.
\newblock \href {https://doi.org/10.3390/jimaging10010018} {Advancements and challenges in handwritten text recognition: A comprehensive survey}.
\newblock \emph{Journal of Imaging}, 10(1).

\bibitem[{Anthropic(2024)}]{claude3}
Anthropic. 2024.
\newblock \href {https://www.anthropic.com/news/claude-3-family} {Introducing the next generation of claude}.

\bibitem[{Anthropic(2025{\natexlab{a}})}]{claude4.1}
Anthropic. 2025{\natexlab{a}}.
\newblock \href {https://www.anthropic.com/news/claude-opus-4-1} {Claude opus 4.1}.

\bibitem[{Anthropic(2025{\natexlab{b}})}]{claude4}
Anthropic. 2025{\natexlab{b}}.
\newblock \href {https://www.anthropic.com/news/claude-4} {Introducing claude 4}.

\bibitem[{Arnold(2022)}]{arnold2022ecpo}
Matthias Arnold. 2022.
\newblock \href {https://doi.org/10.11588/DATA/Z3J0DV} {Early chinese periodicals online (ecpo) [metadata]}.

\bibitem[{Aruta et~al.(2023)Aruta, Lenzi, Huërou, Possamaï, and Pinche}]{aruta2023liber}
Davide Aruta, Martina Lenzi, Armelle~Le Huërou, Marylène Possamaï, and Ariane Pinche. 2023.
\newblock \href {https://github.com/CIHAM-HTR/Liber/data} {Liber}.
\newblock HTR datasets from 14th–15th century manuscripts, including Pierre Bersuire's Old French translation of Titus Livius and Nicolas Trevet's commentaries. Licensed under CC BY 4.0. Released on 2023-04-13.

\bibitem[{Attwood et~al.(2022)Attwood, Sweeney, Stitts, Audebrand, D'Amico, Geelhaar, Hofmann, and Gnasso}]{wien}
Attwood, Sweeney, Stitts, Audebrand, D'Amico, Geelhaar, Hofmann, and Gnasso. 2022.
\newblock \href {https://doi.org/10.5281/zenodo.7537204} {Wien Önb cod. 2160 f. 164-184 ground truth from htr winter school 2022}.

\bibitem[{Bai et~al.(2023)Bai, Bai, Yang, Wang, Tan, Wang, Lin, Zhou, and Zhou}]{bai2023qwenvl}
Jinze Bai, Shuai Bai, Shusheng Yang, Shijie Wang, Sinan Tan, Peng Wang, Junyang Lin, Chang Zhou, and Jingren Zhou. 2023.
\newblock \href {https://arxiv.org/abs/2308.12966} {Qwen-vl: A versatile vision-language model for understanding, localization, text reading, and beyond}.
\newblock \emph{Preprint}, arXiv:2308.12966.

\bibitem[{Bai et~al.(2025)Bai, Chen, Liu, Wang, Ge, Song, Dang, Wang, Wang, Tang, Zhong, Zhu, Yang, Li, Wan, Wang, Ding, Fu, Xu, Ye, Zhang, Xie, Cheng, Zhang, Yang, Xu, and Lin}]{bai2025qwen25vltechnicalreport}
Shuai Bai, Keqin Chen, Xuejing Liu, Jialin Wang, Wenbin Ge, Sibo Song, Kai Dang, Peng Wang, Shijie Wang, Jun Tang, Humen Zhong, Yuanzhi Zhu, Mingkun Yang, Zhaohai Li, Jianqiang Wan, Pengfei Wang, Wei Ding, Zheren Fu, Yiheng Xu, and 8 others. 2025.
\newblock \href {https://arxiv.org/abs/2502.13923} {Qwen2.5-vl technical report}.

\bibitem[{Barcha(2025)}]{books}
Pedro Barcha. 2025.
\newblock Old books dataset.
\newblock \url{https://github.com/PedroBarcha/old-books-dataset}.
\newblock Accessed: 2025-05-18.

\bibitem[{Beckert(2024)}]{beckert2024hofdiarium}
Stefan Beckert. 2024.
\newblock \href {https://doi.org/10.5281/zenodo.14356190} {Ground truth set for handwritten text recognition (htr/ocr): Dresdner hofdiarium 1665 (mscr.dresd.k.80) – 17th century kurrent manuscript}.

\bibitem[{Belay et~al.(2023)Belay, Guyon, Mengiste, Tilahun, Liwicki, Tegegne, and Egele}]{hhd-ethiopic-dataset}
Birhanu~Hailu Belay, Isabelle Guyon, Tadele Mengiste, Bezawork Tilahun, Marcus Liwicki, Tesfa Tegegne, and Romain Egele. 2023.
\newblock Hhd-ethiopic a historical handwritten dataset for ethiopic ocr with baseline models and human-level performance.

\bibitem[{(Belgium)(2024{\natexlab{a}})}]{clarysse}
CrossLang (Belgium). 2024{\natexlab{a}}.
\newblock \href {https://doi.org/10.5281/zenodo.13769222} {Manually validated pagexml files for images in "dagboek ernest clarysse"}.

\bibitem[{(Belgium)(2024{\natexlab{b}})}]{celestino}
CrossLang (Belgium). 2024{\natexlab{b}}.
\newblock \href {https://doi.org/10.5281/zenodo.13760586} {Manually validated pagexml files for images in "diario del soldato bruno celestino"}.

\bibitem[{(Belgium)(2024{\natexlab{c}})}]{domingue}
CrossLang (Belgium). 2024{\natexlab{c}}.
\newblock \href {https://doi.org/10.5281/zenodo.13784411} {Manually validated pagexml files for images in monography "mémoire sur st domingue par h ? m. michel"}.

\bibitem[{Berger et~al.(2022)Berger, Bolte, Führer, Hausleitner, Hutterer, Lüthi, Nancu, Passoni, Pataki, Schröcksnadel, Verri, Wegener, and Hofert}]{klostern}
Michael Berger, Henrike Bolte, Veronika Führer, Felix Hausleitner, Sarah Hutterer, Tim Lüthi, Mihaela Nancu, Erica Passoni, Katalin Pataki, Sophie Schröcksnadel, Giovanni Verri, Dennis Wegener, and Sandra Hofert. 2022.
\newblock \href {https://doi.org/10.5281/zenodo.7466928} {Klosterneuburg, stiftsbibl., cod. 48 - ground truth: Initial release}.

\bibitem[{Beyer and Solberg(2023)}]{norhand}
Yngvil Beyer and Per~Erik Solberg. 2023.
\newblock \href {https://doi.org/10.5281/zenodo.10255840} {Norhand v3 / dataset for handwritten text recognition in norwegian}.

\bibitem[{Biay et~al.(2022)Biay, Boby, Konstantinova, and Cappe}]{biay2022decameronfr}
Sébastien Biay, Victor Boby, Kristina Konstantinova, and Zoé Cappe. 2022.
\newblock \href {https://doi.org/10.5281/zenodo.6126376} {tnah-2021-decameronfr}.

\bibitem[{{Bibliothèque Interuniversitaire de la Sorbonne}(2024)}]{nubis2024ocr}
{Bibliothèque Interuniversitaire de la Sorbonne}. 2024.
\newblock \href {https://nubis.univ-paris1.fr/} {Nubis-ocr: Ground truth dataset for printed books from the nubis digital library}.
\newblock Ground truth dataset of 57 pages from 19 French and Latin books (1602–1989), transcribed using eScriptorium with the CATMuS Print [Large] model and manually corrected. Includes ALTO files, images, OCR model, and transcribed texts.

\bibitem[{Blair(2020)}]{blair2020islamic}
S.S. Blair. 2020.
\newblock \href {https://books.google.com/books?id=m6QxEAAAQBAJ} {\emph{Islamic Calligraphy}}.
\newblock Edinburgh University Press.

\bibitem[{Blecher et~al.(2024)Blecher, Cucurull, Scialom, and Stojnic}]{blecher2023nougat}
Lukas Blecher, Guillem Cucurull, Thomas Scialom, and Robert Stojnic. 2024.
\newblock \href {https://openreview.net/forum?id=fUtxNAKpdV} {Nougat: Neural optical understanding for academic documents}.
\newblock In \emph{The Twelfth International Conference on Learning Representations, {ICLR} 2024, Vienna, Austria, May 7-11, 2024}. OpenReview.net.

\bibitem[{Boenig(2024)}]{ocrd}
Matthias Boenig. 2024.
\newblock \href {https://github.com/OCR-D/gt_structure_text} {{gt\_structure\_text}}.

\bibitem[{Booth et~al.(2024)Booth, Thomas, and Gaizauskas}]{bln}
Callum~William Booth, Alan Thomas, and Robert Gaizauskas. 2024.
\newblock \href {https://aclanthology.org/2024.lrec-main.219} {{BLN}600: A parallel corpus of machine/human transcribed nineteenth century newspaper texts}.
\newblock In \emph{Proceedings of the 2024 Joint International Conference on Computational Linguistics, Language Resources and Evaluation (LREC-COLING 2024)}, pages 2440--2446, Torino, Italia. ELRA and ICCL.

\bibitem[{Bordier et~al.(2023)Bordier, Levenson, Brisville-Fertin, Clérice, and Chagué}]{bordier2023htromance}
Julie Bordier, Matthias~Gille Levenson, Olivier Brisville-Fertin, Thibault Clérice, and Alix Chagué. 2023.
\newblock \href {https://github.com/HTRomance-Project/middle-ages-in-spain} {Htromance, medieval spain corpus of ground-truth for handwritten text recognition and layout segmentation}.
\newblock If you use this dataset, please cite it using the metadata from this file.

\bibitem[{Broder(1997)}]{broder1997resemblance}
Andrei~Z Broder. 1997.
\newblock On the resemblance and containment of documents.
\newblock In \emph{Proceedings. Compression and Complexity of SEQUENCES 1997 (Cat. No. 97TB100171)}, pages 21--29. IEEE.

\bibitem[{Candido and Alu{\'i}sio(2009)}]{candido-aluisio-2009-building}
Arnaldo~Junior Candido and Sandra~Maria Alu{\'i}sio. 2009.
\newblock \href {https://aclanthology.org/2009.tal-2.4/} {Building a corpus-based historical {P}ortuguese dictionary : Challenges and opportunities}.
\newblock In \emph{Traitement Automatique des Langues, Volume 50, Num{\'e}ro 2 : Langues anciennes [Ancient Languages]}, pages 73--102, France. ATALA (Association pour le Traitement Automatique des Langues).

\bibitem[{Carta et~al.(2022)Carta, Jacsont, and Élina Leblanc}]{carta2022fondue}
Constance Carta, Pauline Jacsont, and Élina Leblanc. 2022.
\newblock Fondue spanish chapbooks 19th c. dataset.

\bibitem[{Cascianelli et~al.(2021)Cascianelli, Cornia, Baraldi, Piazzi, Schiuma, and Cucchiara}]{cascianelli2021learning}
Silvia Cascianelli, Marcella Cornia, Lorenzo Baraldi, Maria~Ludovica Piazzi, Rosiana Schiuma, and Rita Cucchiara. 2021.
\newblock Learning to read l’infinito: handwritten text recognition with synthetic training data.
\newblock In \emph{Computer Analysis of Images and Patterns: 19th International Conference, CAIP 2021, Virtual Event, September 28--30, 2021, Proceedings, Part II 19}, pages 340--350. Springer.

\bibitem[{Ceard et~al.(2022)Ceard, Lebreton, and Sajdak}]{ceard2022berlioz}
Lien Ceard, Fanny Lebreton, and Cécile Sajdak. 2022.
\newblock \href {https://doi.org/10.5281/zenodo.6126475} {Projet correspondance berlioz}.
\newblock Created as part of a transcription project of the active correspondence from Hector Berlioz to his sister Nanci Berlioz, by students of the TNAH master's program at the École nationale des chartes.

\bibitem[{{Center for Open Data in the Humanities}(2019)}]{codhjapaneseclassics}
{Center for Open Data in the Humanities}. 2019.
\newblock \href {http://codh.rois.ac.jp/} {Japanese classics dataset (collection of the national museum of japanese literature, etc.)}.

\bibitem[{{Center for Open Data in the Humanities (CODH)} and {National Institute for Japanese Language and Linguistics (NINJAL)}(2025)}]{codhmodernmagazineocr}
{Center for Open Data in the Humanities (CODH)} and {National Institute for Japanese Language and Linguistics (NINJAL)}. 2025.
\newblock \href {https://doi.org/10.20676/00000415} {Modern magazine ocr training dataset}.

\bibitem[{Chagu{\'e} et~al.(2023{\natexlab{a}})Chagu{\'e}, Champougny, Meissel, Genero, Skilbeck-Gaborit, Vanneau, Bey, Le~Fourner, Albert, Riondet, and Martini}]{Chague_Time_Us_Corpus}
Alix Chagu{\'e}, K{\'e}vin Champougny, Nina Meissel, Jean-Damien Genero, Eden Skilbeck-Gaborit, Laurie Vanneau, Laura Bey, Victoria Le~Fourner, Ana{\"\i}s Albert, Charles Riondet, and Manuela Martini. 2023{\natexlab{a}}.
\newblock {Time Us Corpus}.
\newblock Available at \url{https://github.com/HTR-United/timeuscorpus}.

\bibitem[{Chagu{\'e} et~al.()Chagu{\'e}, Ciss{\'e}, and Kichou}]{celestine}
Alix Chagu{\'e}, Julie Ciss{\'e}, and Radia Kichou.
\newblock \href {https://archives.somme.fr/ark:/58483/tjrd8pq42716} {Gt celestine doniau-danest}.

\bibitem[{Chagu{\'e} et~al.(2023{\natexlab{b}})Chagu{\'e}, Cl{\'e}rice, Mazoue, and Van~Kote}]{cremma-testament}
Alix Chagu{\'e}, Thibault Cl{\'e}rice, Ana{\"\i}s Mazoue, and Elsa Van~Kote. 2023{\natexlab{b}}.
\newblock Cremma-an-testamentdepoilus.
\newblock Available at \url{https://github.com/HTR-United/CREMMA-AN-TestamentDePoilus}.

\bibitem[{Chagué(2021)}]{chague2021tapuscorpus}
Alix Chagué. 2021.
\newblock \href {https://doi.org/10.5072/zenodo.977649} {Tapuscorpus}.
\newblock If you use this software, please cite it as above.

\bibitem[{Chagué and Pérez(2023)}]{chague2023peraire}
Alix Chagué and Gilles Pérez. 2023.
\newblock \href {https://doi.org/10.5281/zenodo.7185907} {Peraire ground truth}.

\bibitem[{Chen et~al.(2024)Chen, Wu, Wang, Su, Chen, Xing, Zhong, Zhang, Zhu, Lu, Li, Luo, Lu, Qiao, and Dai}]{chen2024internvl}
Zhe Chen, Jiannan Wu, Wenhai Wang, Weijie Su, Guo Chen, Sen Xing, Muyan Zhong, Qinglong Zhang, Xizhou Zhu, Lewei Lu, Bin Li, Ping Luo, Tong Lu, Yu~Qiao, and Jifeng Dai. 2024.
\newblock \href {https://doi.org/10.1109/CVPR52733.2024.02283} {Intern {VL:} scaling up vision foundation models and aligning for generic visual-linguistic tasks}.
\newblock In \emph{{IEEE/CVF} Conference on Computer Vision and Pattern Recognition, {CVPR} 2024, Seattle, WA, USA, June 16-22, 2024}, pages 24185--24198. {IEEE}.

\bibitem[{Cheng et~al.(2024)Cheng, Frankem{\"o}lle, Axelsson, and Vats}]{cheng-etal-2024-uncovering}
Liang Cheng, Jonas Frankem{\"o}lle, Adam Axelsson, and Ekta Vats. 2024.
\newblock \href {https://aclanthology.org/2024.latechclfl-1.12/} {Uncovering the handwritten text in the margins: End-to-end handwritten text detection and recognition}.
\newblock In \emph{Proceedings of the 8th Joint SIGHUM Workshop on Computational Linguistics for Cultural Heritage, Social Sciences, Humanities and Literature (LaTeCH-CLfL 2024)}, pages 111--120, St. Julians, Malta. Association for Computational Linguistics.

\bibitem[{Chiang et~al.(2024)Chiang, Zheng, Sheng, Angelopoulos, Li, Li, Zhu, Zhang, Jordan, Gonzalez, and Stoica}]{chiang2024chatbot}
Wei{-}Lin Chiang, Lianmin Zheng, Ying Sheng, Anastasios~Nikolas Angelopoulos, Tianle Li, Dacheng Li, Banghua Zhu, Hao Zhang, Michael~I. Jordan, Joseph~E. Gonzalez, and Ion Stoica. 2024.
\newblock \href {https://openreview.net/forum?id=3MW8GKNyzI} {Chatbot arena: An open platform for evaluating llms by human preference}.
\newblock In \emph{Forty-first International Conference on Machine Learning, {ICML} 2024, Vienna, Austria, July 21-27, 2024}. OpenReview.net.

\bibitem[{Chiffoleau(2021)}]{dahn}
Floriane Chiffoleau. 2021.
\newblock \href {https://doi.org/10.5281/zenodo.5911868} {dahncorpus}.

\bibitem[{Christensen et~al.(2022)Christensen, Davoury, Haedo, Kervegan, and Sanchez-Oeconomo}]{christensen2022exposition1878}
Kelly Christensen, Baudoin Davoury, Anahi Haedo, Paul Kervegan, and Esteban Sanchez-Oeconomo. 2022.
\newblock \href {https://doi.org/10.5281/zenodo.6126447} {Projet exposition universelle de 1878}.
\newblock Created as part of a transcription project on the 1878 International Congress of Ethnographic Sciences held during the Exposition Universelle in Paris, by students of the TNAH master's program at the École nationale des chartes.

\bibitem[{Claass et~al.(2021)Claass, Gain, and Martin-Vigier}]{claass2021papiersbarye}
Victor Claass, Justine Gain, and Suzanne Martin-Vigier. 2021.
\newblock \href {https://gitlab.inha.fr/snr/LesPapiersBarye/-/tree/main/LesPapiersBarye-Transcriptions} {Les papiers barye — transcriptions brutes en format alto}.
\newblock Transcriptions created with Transkribus in 2020–2021 from the Antoine-Louis Barye archival collection (1795–1875), held at the Bibliothèque de l'INHA, Jacques Doucet collections. The collection is catalogued in Calames.

\bibitem[{Clausner et~al.(2017)Clausner, Antonacopoulos, Derrick, and Pletschacher}]{reid2017}
Christian Clausner, Apostolos Antonacopoulos, Tom Derrick, and Stefan Pletschacher. 2017.
\newblock \href {https://doi.org/10.1109/ICDAR.2017.230} {Icdar2017 competition on recognition of early indian printed documents - reid2017}.
\newblock In \emph{2017 14th IAPR International Conference on Document Analysis and Recognition (ICDAR)}, volume~01, pages 1411--1416.

\bibitem[{Clausner et~al.(2019{\natexlab{a}})Clausner, Antonacopoulos, Derrick, and Pletschacher}]{reid2019}
Christian Clausner, Apostolos Antonacopoulos, Tom Derrick, and Stefan Pletschacher. 2019{\natexlab{a}}.
\newblock Icdar2019 competition on recognition of early indian printed documents--reid2019.
\newblock In \emph{2019 International Conference on Document Analysis and Recognition (ICDAR)}, pages 1527--1532. IEEE.

\bibitem[{Clausner et~al.(2018)Clausner, Antonacopoulos, McGregor, and Wilson-Nunn}]{rasm2018}
Christian Clausner, Apostolos Antonacopoulos, Nora McGregor, and Daniel Wilson-Nunn. 2018.
\newblock \href {https://api.semanticscholar.org/CorpusID:56596588} {Icfhr 2018 competition on recognition of historical arabic scientific manuscripts – rasm2018}.
\newblock \emph{2018 16th International Conference on Frontiers in Handwriting Recognition (ICFHR)}, pages 471--476.

\bibitem[{Clausner et~al.(2015)Clausner, Pletschacher, and Antonacopoulos}]{europeana}
Christian Clausner, Stefan Pletschacher, and Apostolos Antonacopoulos. 2015.
\newblock Prima europeana newspaper dataset.
\newblock \url{https://www.primaresearch.org/datasets}.
\newblock Accessed: 2025-03-25.

\bibitem[{Clausner et~al.(2019{\natexlab{b}})Clausner, Pletschacher, and Antonacopoulos}]{rasm2019}
Christos Clausner, Stefan Pletschacher, and Apostolos Antonacopoulos. 2019{\natexlab{b}}.
\newblock {RASM2019} dataset: Historical arabic scientific manuscripts.
\newblock \url{https://www.primaresearch.org/RASM2019/resources}.
\newblock Accessed: 2025-03-24.

\bibitem[{Claustre et~al.(2023)Claustre, Smith, Torres~Aguilar, Bretthauer, Brochard, Canteaut, Cottereau, Delivré, Denglos, Jolivet, Julerot, Kouamé, Lusset, Massoni, Nadiras, Perreaux, Regazzi, and Treglia}]{endp}
Julie Claustre, Darwin Smith, Sergio Torres~Aguilar, Isabelle Bretthauer, Pierre Brochard, Olivier Canteaut, Emilie Cottereau, Fabrice Delivré, Mathilde Denglos, Vincent Jolivet, Véronique Julerot, Thierry Kouamé, Elisabeth Lusset, Anne Massoni, Sebastien Nadiras, Nicolas Perreaux, Hugo Regazzi, and Mathilde Treglia. 2023.
\newblock \href {https://doi.org/10.5281/zenodo.7575693} {The e-ndp project : collaborative digital edition of the chapter registers of notre-dame of paris (1326-1504). ground-truth for handwriting text recognition (htr) on late medieval manuscripts.}

\bibitem[{Cl{\'e}rice et~al.(2022{\natexlab{a}})Cl{\'e}rice, Chagu{\'e}, Faure, Norindr, Mazoue, and Davoury}]{cremma-mss17}
Thibault Cl{\'e}rice, Alix Chagu{\'e}, Margaux Faure, Jade Norindr, Anais Mazoue, and Baudoin Davoury. 2022{\natexlab{a}}.
\newblock Cremma manuscrits du 17e.
\newblock Dataset published by HTR United. Edited by Alix Chagu{\'e} and Thibault Cl{\'e}rice. \url{https://github.com/HTR-United/CREMMA-MSS-17}.

\bibitem[{Cl{\'e}rice et~al.(2022{\natexlab{b}})Cl{\'e}rice, Chagu{\'e}, and Vlachou-Efstathiou}]{cremma-medieval-lat}
Thibault Cl{\'e}rice, Alix Chagu{\'e}, and Malamatenia Vlachou-Efstathiou. 2022{\natexlab{b}}.
\newblock Cremma medii aevi.
\newblock \url{https://github.com/HTR-United/CREMMA-Medieval-LAT}. DOI: 10.5281/zenodo.7013436.

\bibitem[{Clérice(2021)}]{cremma-16-17-print}
Thibault Clérice. 2021.
\newblock \href {https://github.com/HTR-United/cremma-16-17-print} {Cremma early modern books}.

\bibitem[{Clérice and Chagué(2023)}]{cremma-mss20}
Thibault Clérice and Alix Chagué. 2023.
\newblock \href {https://github.com/HTR-United/CREMMA-MSS-20} {Cremma manuscrits du 20e}.

\bibitem[{Clérice et~al.(2023)Clérice, Chagué, Davoury, Doat, Faure, and Humeau}]{cremma-mss19}
Thibault Clérice, Alix Chagué, Baudouin Davoury, Soline Doat, Margaux Faure, and Maxime Humeau. 2023.
\newblock \href {https://github.com/HTR-United/CREMMA-MSS-19} {Cremma manuscrits du 19e}.

\bibitem[{Constum et~al.(2022)Constum, Kempf, Paquet, Tranouez, Chatelain, Bree, and Merveille}]{constum2022popp}
Thomas Constum, Nicolas Kempf, Thierry Paquet, Pierrick Tranouez, Clément Chatelain, Sandra Bree, and François Merveille. 2022.
\newblock \href {https://doi.org/10.5281/zenodo.6581158} {Popp datasets: datasets for handwriting recognition from french population census}.

\bibitem[{Craene et~al.(2022)Craene, rayondemiel, Clérice, and Reignier}]{decraene2022argus}
Valentin~De Craene, rayondemiel, Thibault Clérice, and Virgile Reignier. 2022.
\newblock \href {https://doi.org/10.5281/zenodo.6126366} {psl-chartes-htr-students/tnah-2021-argusdesbrevets: 1.0}.

\bibitem[{Cugy et~al.(2022)Cugy, Fieschi, Peyrard, Prohin, Sarda, de~l'histoire de~l'art (INHA), and nationale~de France}]{doucet}
Pascale Cugy, Caroline Fieschi, Alix Peyrard, Lucie Prohin, Marie-Anne Sarda, Institut~National de~l'histoire de~l'art (INHA), and Bibliothèque nationale~de France. 2022.
\newblock La correspondance jacques doucet - rené jean.
\newblock \url{https://gitlab.inha.fr/snr/LaCorrespondanceDoucetReneJean}.
\newblock Project website: \url{https://skylab.inha.fr/PENSE/LettresDeJacquesDoucetAReneJean1908-1929/}. Letters and documents (1908--1929) conserved at the Bibliothèque nationale de France, Département des manuscrits, NAF 13124.

\bibitem[{Dang et~al.(2022)Dang, Nguyen, Pham, Nguyen, Chau, Ngo, Nguyen, Phan, Trinh, Nguyen et~al.}]{nomnaocr}
Hoang-Quan Dang, Duy-Anh Nguyen, Phu-Phuoc Pham, Ngoc-Thinh Nguyen, Tan Chau, Duc-Vu Ngo, Trung-Hieu Nguyen, Chau-Thang Phan, The-Hien Trinh, Minh-Tri Nguyen, and 1 others. 2022.
\newblock Nomnaocr: The first dataset for optical character recognition on han-nom script.
\newblock In \emph{2022 RIVF International Conference on Computing and Communication Technologies (RIVF)}, pages 476--481. IEEE.

\bibitem[{David~Glück(2018)}]{salamanca-guidelines}
Andreas~Wagner David~Glück, Cindy Rico~Carmona. 2018.
\newblock Edition guidelines.
\newblock \url{https://www.salamanca.school/guidelines.html}.

\bibitem[{Dell et~al.(2023)Dell, Carlson, Bryan, Silcock, Arora, Shen, D'Amico{-}Wong, Le, Querubin, and Heldring}]{stories}
Melissa Dell, Jacob Carlson, Tom Bryan, Emily Silcock, Abhishek Arora, Zejiang Shen, Luca D'Amico{-}Wong, Quan Le, Pablo Querubin, and Leander Heldring. 2023.
\newblock \href {http://papers.nips.cc/paper\_files/paper/2023/hash/ffeb860479ccae44d84c0de32acd693d-Abstract-Datasets\_and\_Benchmarks.html} {American stories: {A} large-scale structured text dataset of historical {U.S.} newspapers}.
\newblock In \emph{Advances in Neural Information Processing Systems 36: Annual Conference on Neural Information Processing Systems 2023, NeurIPS 2023, New Orleans, LA, USA, December 10 - 16, 2023}.

\bibitem[{{Deutsches Textarchiv}()}]{dta2025}
{Deutsches Textarchiv}.
\newblock \href {https://www.deutschestextarchiv.de/} {Grundlage für ein referenzkorpus der neuhochdeutschen sprache}.

\bibitem[{Do et~al.(2024)Do, Tran, Vo, and Kim}]{do2024referencebased}
Thao Do, Dinh~Phu Tran, An~Vo, and Daeyoung Kim. 2024.
\newblock \href {https://arxiv.org/abs/2410.13305} {Reference-based post-ocr processing with llm for precise diacritic text in historical document recognition}.
\newblock \emph{Preprint}, arXiv:2410.13305.

\bibitem[{Doat et~al.(2022)Doat, Falcoz, Faure, Mazoué, and Menu}]{doat2022notredame}
Soline Doat, Elsa Falcoz, Margaux Faure, Anaïs Mazoué, and Ariane Menu. 2022.
\newblock \href {https://doi.org/10.5281/zenodo.6126491} {Projet notre-dame}.

\bibitem[{Dolfing et~al.(2020)Dolfing, Bellegarda, Chorowski, Marxer, and Laurent}]{scribble}
Hans~J.G.A. Dolfing, Jerome Bellegarda, Jan Chorowski, Ricard Marxer, and Antoine Laurent. 2020.
\newblock {The ``ScribbleLens'' Dutch historical handwriting corpus}.
\newblock In \emph{International Conference on Frontiers of Handwriting Recognition (ICFHR)}, page To Appear.
\newblock \url{http://www.openslr.org/84/}.

\bibitem[{Drobac et~al.(2017)Drobac, Kauppinen, and Lind{\'e}n}]{drobac-etal-2017-ocr}
Senka Drobac, Pekka Kauppinen, and Krister Lind{\'e}n. 2017.
\newblock \href {https://aclanthology.org/W17-0209/} {{OCR} and post-correction of historical {F}innish texts}.
\newblock In \emph{Proceedings of the 21st Nordic Conference on Computational Linguistics}, pages 70--76, Gothenburg, Sweden. Association for Computational Linguistics.

\bibitem[{Dubois et~al.(2024)Dubois, Clérice, Mamie, Schlaeppi, Rudaz, and Schmied}]{dubois2024valais}
Alain Dubois, Thibault Clérice, Delphine Mamie, Darius Schlaeppi, Clémence Rudaz, and Marie-Caroline Schmied. 2024.
\newblock \href {https://github.com/PonteIneptique/valais-recensement; https://recensements.vallesiana.ch} {Tables du recensement du valais}.

\bibitem[{Efstathiou et~al.(2022)Efstathiou, Leroy, and Maulu}]{vlachouefstathiou2022boccace}
Malamatenia~Vlachou Efstathiou, Noé Leroy, and Marco Maulu. 2022.
\newblock \href {https://doi.org/10.5281/zenodo.6126613} {git-project-boccace}.

\bibitem[{Eichenberger and Suwelack(2021)}]{faithful}
Nicole Eichenberger and Hedwig Suwelack. 2021.
\newblock \href {https://doi.org/10.5281/zenodo.5582483} {Faithful transcriptions data set: Tei/xml-encoded transcriptions of medieval theological manuscripts}.

\bibitem[{Fernández-Mota et~al.(2014)Fernández-Mota, Almazán, Cirera, Fornés, and Lladós}]{fernandezmota2014bh2m}
David Fernández-Mota, Joan Almazán, Núria Cirera, Alicia Fornés, and Josep Lladós. 2014.
\newblock Bh2m: The barcelona historical handwritten marriages database.
\newblock In \emph{Proceedings of the 22nd International Conference on Pattern Recognition (ICPR)}, pages 256--261.

\bibitem[{Fischer et~al.(2009)Fischer, Frinken, and Bunke}]{parzival}
Andreas Fischer, Valentin Frinken, and Horst Bunke. 2009.
\newblock Parzival database.
\newblock \url{https://fki.tic.heia-fr.ch/databases/parzival-database}.
\newblock Accessed: 2025-03-25.

\bibitem[{Fischer et~al.(2012)Fischer, Keller, Frinken, and Bunke}]{saintgall}
Andreas Fischer, Alex Keller, Valentin Frinken, and Horst Bunke. 2012.
\newblock Saint gall database.
\newblock \url{https://fki.tic.heia-fr.ch/databases/saint-gall-database}.
\newblock Accessed: 2025-03-25.

\bibitem[{Frincu et~al.(2023)Frincu, Frincu, and Penteliuc}]{frincu-etal-2023-challenges}
Marc Frincu, Simina Frincu, and Marius~E. Penteliuc. 2023.
\newblock \href {https://aclanthology.org/2023.ldk-1.20/} {Challenges and solutions in transliterating 19th century {R}omanian texts from the transitional to the {L}atin script}.
\newblock In \emph{Proceedings of the 4th Conference on Language, Data and Knowledge}, pages 226--231, Vienna, Austria. NOVA CLUNL, Portugal.

\bibitem[{Fu et~al.(2025)Fu, Yang, Kuang, Song, Li, Zhu, Luo, Wang, Lu, Huang et~al.}]{fu2024ocrbench}
Ling Fu, Biao Yang, Zhebin Kuang, Jiajun Song, Yuzhe Li, Linghao Zhu, Qidi Luo, Xinyu Wang, Hao Lu, Mingxin Huang, and 1 others. 2025.
\newblock \href {https://arxiv.org/abs/2501.00321} {Ocrbench v2: An improved benchmark for evaluating large multimodal models on visual text localization and reasoning}.
\newblock \emph{ArXiv preprint}, abs/2501.00321.

\bibitem[{Gabay(2024)}]{fondue-fr-print-16}
Simon Gabay. 2024.
\newblock \href {https://github.com/FoNDUE-HTR/FONDUE-FR-PRINT-16} {Fondue-fr-print-16}.

\bibitem[{Gabay and Carrasco~Luján(2024)}]{fondue-es-print-19}
Simon Gabay and Carmen Carrasco~Luján. 2024.
\newblock \href {https://github.com/FoNDUE-HTR/FONDUE-ES-PRINT-19} {Fondue-es-print-19}.

\bibitem[{Gabay and Dolto(2024)}]{fondue-fr-print-20}
Simon Gabay and Sophie Dolto. 2024.
\newblock \href {https://github.com/FoNDUE-HTR/FONDUE-FR-PRINT-20} {Fondue-fr-print-20}.

\bibitem[{Gabay et~al.(2023{\natexlab{a}})Gabay, Nahon, Cicchini, Jaureguy, and Chappuis}]{fondue-fr-mss-18}
Simon Gabay, Peter Nahon, Marco Cicchini, Yvan Jaureguy, and Loraine Chappuis. 2023{\natexlab{a}}.
\newblock \href {https://github.com/FoNDUE-HTR/FONDUE-FR-MSS-18} {Fondue-fr-mss-18}.

\bibitem[{Gabay et~al.(2023{\natexlab{b}})Gabay, Perregaux, and Da~Silva~Fernandes}]{fondue-it-print-20}
Simon Gabay, Myriam Perregaux, and Jessica Da~Silva~Fernandes. 2023{\natexlab{b}}.
\newblock \href {https://github.com/FoNDUE-HTR/FONDUE-EN-PRINT-20} {Fondue-en-print-20}.

\bibitem[{Gabay et~al.(2023{\natexlab{c}})Gabay, Perregaux, and Fernandes}]{gabay2023fondue}
Simon Gabay, Myriam Perregaux, and Jessica Da~Silva Fernandes. 2023{\natexlab{c}}.
\newblock \href {https://github.com/FoNDUE-HTR/FONDUE-EN-PRINT-20} {Fondue-en-print-20}.

\bibitem[{Gabay et~al.(2023{\natexlab{d}})Gabay, Pinche, Fabert, and Christensen}]{imprime-18e-siecle}
Simon Gabay, Ariane Pinche, Eliott Fabert, and Kelly Christensen. 2023{\natexlab{d}}.
\newblock Donn{\'e}es imprim{\'e}s du 18e si{\`e}cle.
\newblock Dataset published by HTR United. Edited by Alix Chagu{\'e} and Thibault Cl{\'e}rice. Available at \url{https://github.com/Gallicorpora/HTR-imprime-18e-siecle}.

\bibitem[{Gabay et~al.(2023{\natexlab{e}})Gabay, Pinche, Fabert, Vlachou-Efstathiou, and Christensen}]{imprime-17e-siecle}
Simon Gabay, Ariane Pinche, Eliott Fabert, malamatenia Vlachou-Efstathiou, and Kelly Christensen. 2023{\natexlab{e}}.
\newblock Imprim{\'e}s 17e si{\`e}cle.
\newblock HTR United (publisher). Edited by Alix Chagu{\'e} and Thibault Cl{\'e}rice. Dataset. Available at \url{https://github.com/Gallicorpora/HTR-imprime-17e-siecle}.

\bibitem[{Gabay et~al.(2023{\natexlab{f}})Gabay, Pinche, Leroy, and Christensen}]{htr-incunable-15e}
Simon Gabay, Ariane Pinche, No{\'e} Leroy, and Kelly Christensen. 2023{\natexlab{f}}.
\newblock Donn{\'e}es htr incunables du 15e si{\`e}cle.
\newblock \url{https://github.com/Gallicorpora/HTR-incunable-15e-siecle}.
\newblock Edited by Alix Chagu{\'e} and Thibault Cl{\'e}rice. Dataset. Publisher: HTR United.

\bibitem[{Gabay et~al.(2022)Gabay, Pinche, Vlachou-Efstathiou, and Christensen}]{gabay2022htr16e}
Simon Gabay, Ariane Pinche, Malamatenia Vlachou-Efstathiou, and Kelly Christensen. 2022.
\newblock \href {https://github.com/Gallicorpora/HTR-imprime-16e-siecle} {Données htr imprimés du 16e siècle}.

\bibitem[{{Gemini Team Google}(2023)}]{geminiteam2023}
{Gemini Team Google}. 2023.
\newblock \href {https://arxiv.org/abs/2312.11805} {Gemini: A family of highly capable multimodal models}.
\newblock \emph{ArXiv preprint}, abs/2312.11805.

\bibitem[{{Gemma Team} et~al.(2024){Gemma Team}, Mesnard, Hardin, Dadashi, Bhupatiraju, Pathak, Sifre, Rivi{\`e}re, Kale, Love et~al.}]{team2024gemma}
{Gemma Team}, Thomas Mesnard, Cassidy Hardin, Robert Dadashi, Surya Bhupatiraju, Shreya Pathak, Laurent Sifre, Morgane Rivi{\`e}re, Mihir~Sanjay Kale, Juliette Love, and 1 others. 2024.
\newblock \href {https://arxiv.org/abs/2403.08295} {Gemma: Open models based on gemini research and technology}.
\newblock \emph{ArXiv preprint}, abs/2403.08295.

\bibitem[{Ghereghlou(2018)}]{safavi}
Kioumars Ghereghlou, editor. 2018.
\newblock \href {https://doi.org/10.2307/j.ctv9b2vrs.5} {\emph{Persian Text}}, volume~98, pages 1--372.
\newblock American Oriental Society.
\newblock Accessed 20 May 2025.

\bibitem[{Glaise et~al.(2023)Glaise, Clérice, Boschetti, Fischer, and Chagué}]{glaise2023htromance}
Anthony Glaise, Thibault Clérice, Federico Boschetti, Franz Fischer, and Alix Chagué. 2023.
\newblock \href {https://doi.org/10.5281/zenodo.8362890} {Htromance, medieval latin corpus of ground-truth for handwritten text recognition and layout segmentation}.

\bibitem[{Grüning et~al.(2016)Grüning, Leifert, Michael, Strauß, Weidemann, and Labahn}]{konzils}
Tobias Grüning, Gundram Leifert, Johannes Michael, Tobias Strauß, Max Weidemann, and Roger Labahn. 2016.
\newblock \href {https://doi.org/10.5281/zenodo.215383} {read\_dataset\_german\_konzilsprotokolle}.

\bibitem[{Gu{\'e}ville and Wrisley(2022)}]{Guville2022TranscribingMM}
Estelle Gu{\'e}ville and David~Joseph Wrisley. 2022.
\newblock \href {https://arxiv.org/abs/2207.07726} {Transcribing medieval manuscripts for machine learning}.
\newblock \emph{ArXiv preprint}, abs/2207.07726.

\bibitem[{Guimarães et~al.(2022)Guimarães, Maurel, Ozturk, and Chagué}]{guimaraes2022memorials}
Ingrid Guimarães, Perrine Maurel, Yagmur Ozturk, and Alix Chagué. 2022.
\newblock \href {https://doi.org/10.5281/zenodo.6126625} {Memorials for jane lathrop stanford}.

\bibitem[{Guo et~al.(2025)Guo, Wu, Zhu, Leng, Shi, Chen, Fan, Wang, Jiang, Wang, Chen, Huang, Lei, Yuan, Luo, Liu, Ye, Qian, Yan, Zhao, Peng, Li, Yuan, Wu, Cheng, Liu, Wang, Zeng, Liu, Qin, Ding, Xiao, Zhang, Zhang, Xiong, Peng, Chen, Li, Hu, Lin, Hu, Zhang, Wu, Li, Liu, Ling, Qin, Wang, He, Zhang, Yi, Liao, Huang, Zhang, Deng, Deng, Lin, Yuan, Li, Gou, Lou, Wei, Liu, Li, Zhu, Zhong, Li, Zhang, Wu, Li, Xiao, Lin, Yang, Wang, Ji, Hao, Shen, Li, Li, Wu, Zhu, Jiao, Feng, Chen, Duan, Liu, Zeng, Tang, Sun, Chen, Long, Feng, Zhan, Fang, Lu, Hua, Liu, Shen, Zhang, Shen, Wang, Pan, Zhang, Li, Li, Li, Shi, Han, Xiang, Chen, Chen, Li, Yan, Chi, Liu, Du, Wang, Pan, Chen, Chen, Wu, Yuan, Shuai, Tao, Zheng, Zhang, Zhang, Wang, Yang, Zhao, Xu, Liang, Yan, Zhong, Cao, Wu, Liu, Chang, Cai, Ao, Yang, Zhang, Zhong, Jia, Weng, Yu, Huang, Zhu, Yang, Wang, Long, Yin, Li, Zhu, Jia, Zhang, Liu, Zhang, Yang, Luo, Chen, Zhong, Xiao, Li, Wu, Wen, Du, Zhang, Ye, Wu, Liu, Yue, Zhou, Yuan, Xu, Yang, Zhang, Fang, Li, Ren, Xiong, Hong,
  Wang, Sun, Wang, Cai, Zha, An, Zhao, Xu, Chen, Wu, Zheng, Wang, Huang, Zhu, and Song}]{guo2025seed15vltechnicalreport}
Dong Guo, Faming Wu, Feida Zhu, Fuxing Leng, Guang Shi, Haobin Chen, Haoqi Fan, Jian Wang, Jianyu Jiang, Jiawei Wang, Jingji Chen, Jingjia Huang, Kang Lei, Liping Yuan, Lishu Luo, Pengfei Liu, Qinghao Ye, Rui Qian, Shen Yan, and 178 others. 2025.
\newblock \href {https://arxiv.org/abs/2505.07062} {Seed1.5-vl technical report}.

\bibitem[{Guéville and Wrisley(2021)}]{gueville2021groundtruth}
Estelle Guéville and David~Joseph Wrisley. 2021.
\newblock \href {https://doi.org/10.5281/zenodo.7653691} {Ground truth used in htr for the paris bible project}.
\newblock Released on 2021-10-10.

\bibitem[{Halter-Pernet et~al.(2021)Halter-Pernet, Teuscher, Hodel, Barwitzki, Egloff, Henggeler, Nadig, Steinmann, Stettler, and Prada~Ziegler}]{charters}
Colette Halter-Pernet, Simon Teuscher, Tobias Hodel, Lukas Barwitzki, Salome Egloff, Fabian Henggeler, Michael Nadig, Anina Steinmann, Sabine Stettler, and Ismail Prada~Ziegler. 2021.
\newblock \href {https://doi.org/10.5281/zenodo.5179361} {Charters and records of königsfelden abbey and bailiwick (1308-1662)}.

\bibitem[{Hammarström et~al.(2024)Hammarström, Forkel, Haspelmath, and Bank}]{glottolog}
Harald Hammarström, Robert Forkel, Martin Haspelmath, and Sebastian Bank. 2024.
\newblock \href {https://doi.org/10.5281/zenodo.14006617} {Glottolog 5.1}.
\newblock \url{https://glottolog.org/}.

\bibitem[{Harley et~al.(2015)Harley, Ufkes, and Derpanis}]{harley2015evaluation}
Adam~W Harley, Alex Ufkes, and Konstantinos~G Derpanis. 2015.
\newblock Evaluation of deep convolutional nets for document image classification and retrieval.
\newblock In \emph{2015 13th international conference on document analysis and recognition (ICDAR)}, pages 991--995. IEEE.

\bibitem[{Hodel et~al.(2021)Hodel, Schoch, and Dängeli}]{hodel2021stabs}
Tobias Hodel, David Schoch, and Peter Dängeli. 2021.
\newblock \href {https://doi.org/10.5281/zenodo.5153263} {Handwritten text recognition ground truth set: Stabs ratsbücher o10, urfehdenbuch x}.

\bibitem[{Hoeben(2024)}]{hoeben2024verard}
Laurie Hoeben. 2024.
\newblock \href {https://github.com/LaurieHoeben/Verard-corpus} {Antoine verard extracts}.
\newblock Transcriptions of parts of Antoine Vérard’s editions princeps of "Tristan", "Merlin" and "Gyron le Courtoys".

\bibitem[{Huang et~al.(2025)Huang, Jia, Zhai, Cao, Ye, Zhao, Xu, Hu, and Lin}]{huang2025visionr1}
Wenxuan Huang, Bohan Jia, Zijie Zhai, Shaosheng Cao, Zheyu Ye, Fei Zhao, Zhe Xu, Yao Hu, and Shaohui Lin. 2025.
\newblock \href {https://arxiv.org/abs/2503.06749} {Vision-r1: Incentivizing reasoning capability in multimodal large language models}.
\newblock \emph{Preprint}, arXiv:2503.06749.

\bibitem[{Humbel et~al.(2024)Humbel, Vlachidis, Nyhan, and of~the British~Museum}]{humbel2024sloanelab}
Marco Humbel, Andreas Vlachidis, Julianne Nyhan, and The~Trustees of~the British~Museum. 2024.
\newblock Sloane lab htr model.
\newblock Handwritten Text Recognition training data (layout segmentation and transcriptions) for the Sloane Lab HTR model.

\bibitem[{{International Organization for Standardization}(2022)}]{iso15924}
{International Organization for Standardization}. 2022.
\newblock Information and documentation — codes for the representation of names of scripts.
\newblock \url{https://iso639-3.sil.org/}.
\newblock Accessed: 2025-03-25.

\bibitem[{Jaech et~al.(2024)Jaech, Kalai, Lerer, Richardson, El-Kishky, Low, Helyar, Madry, Beutel, Carney et~al.}]{openai2024-o1}
Aaron Jaech, Adam Kalai, Adam Lerer, Adam Richardson, Ahmed El-Kishky, Aiden Low, Alec Helyar, Aleksander Madry, Alex Beutel, Alex Carney, and 1 others. 2024.
\newblock \href {https://arxiv.org/abs/2412.16720} {Openai o1 system card}.
\newblock \emph{ArXiv preprint}, abs/2412.16720.

\bibitem[{Jahan and Gabay(2021)}]{ocr17plus}
Claire Jahan and Simon Gabay. 2021.
\newblock \href {https://doi.org/none} {{OCR17 +}}.

\bibitem[{Joyeux-Prunel et~al.(2023)Joyeux-Prunel, Gabay, Rizzello, Berlincourt, Rizzi, Tesser, Bukvic, Diaz, Aebi, and Bickel}]{joyeuxprunel2023fondue}
Béatrice Joyeux-Prunel, Simon Gabay, Martina Rizzello, Valéry Berlincourt, Elena~Maria Rizzi, Stefania Tesser, Victoria Bukvic, Jaime Diaz, Guillaume Aebi, and Raoul Bickel. 2023.
\newblock \href {https://github.com/FoNDUE-HTR/FONDUE-MLT-ART} {Fondue-mlt-art}.

\bibitem[{Kahle et~al.(2017)Kahle, Colutto, Hackl, and Mühlberger}]{transcribus2017}
Philip Kahle, Sebastian Colutto, Günter Hackl, and Günter Mühlberger. 2017.
\newblock \href {https://doi.org/10.1109/ICDAR.2017.307} {Transkribus - a service platform for transcription, recognition and retrieval of historical documents}.
\newblock In \emph{2017 14th IAPR International Conference on Document Analysis and Recognition (ICDAR)}, volume~04, pages 19--24.

\bibitem[{Kamlah et~al.(2024)Kamlah, Schmidt, Shigapov, and Weil}]{kamlah2024reichsanzeiger}
Jan Kamlah, Thomas Schmidt, Renat Shigapov, and Stefan Weil. 2024.
\newblock \href {https://doi.org/10.1016/j.dib.2024.110274} {reichsanzeiger-gt}.

\bibitem[{Karpinski et~al.(2018)Karpinski, Lohani, and Belaid}]{karpinski2018ocr}
Romain Karpinski, Devashish Lohani, and Abdel Belaid. 2018.
\newblock \href {https://hal.science/hal-01981731} {Metrics for complete evaluation of ocr performance}.
\newblock In \emph{Proceedings of the 22nd International Conference on Image Processing, Computer Vision, and Pattern Recognition (IPCV)}, Las Vegas, United States.
\newblock HAL Id: hal-01981731.

\bibitem[{Kassis et~al.(2017)Kassis, Abdalhaleem, Droby, Alaasam, and El-Sana}]{kassis2017vmlhd}
Majeed Kassis, Alaa Abdalhaleem, Ahmad Droby, Reem Alaasam, and Jihad El-Sana. 2017.
\newblock Vml-hd: The historical arabic documents dataset for recognition systems.
\newblock In \emph{1st International Workshop on Arabic Script Analysis and Recognition}. IEEE.

\bibitem[{Katuščák et~al.(2024)Katuščák, Nižníková, Mikušková, Halfarová, Gajdošová, Málková, Taufrová, Nagy, Kováčová-Pohlová, Šmida, and Kociánová}]{slovensky}
Dušan Katuščák, Lucia Nižníková, Michaela Mikušková, Nikola Halfarová, Terezie Gajdošová, Lenka Málková, Nikol Taufrová, Imrich Nagy, Klára Kováčová-Pohlová, Matej Šmida, and Nela Kociánová. 2024.
\newblock \href {https://doi.org/10.5281/zenodo.11218527} {Slovenský supermodel p\&t1 (sspt1) : Matej bel university skriptor project datasets}.

\bibitem[{Kavukcuoglu(2025)}]{google2025gemini-2.5}
Koray Kavukcuoglu. 2025.
\newblock Gemini 2.5: Our most intelligent ai model.
\newblock \url{https://blog.google/technology/google-deepmind/gemini-model-thinking-updates-march-2025/#gemini-2-5-thinking}.
\newblock Accessed: 2025-05-18.

\bibitem[{Keijser(2024)}]{voc}
Liesbeth Keijser. 2024.
\newblock \href {https://doi.org/10.5281/zenodo.11209325} {6000 ground truth of voc and notarial deeds 3.000.000 htr of voc, wic and notarial deeds}.

\bibitem[{Kemper(2021)}]{gado}
Simon~Carlos Kemper. 2021.
\newblock \href {https://doi.org/10.5281/zenodo.4980170} {Gado2: multilingual newspapers from the netherlands indies}.

\bibitem[{Kiessling et~al.(2019)Kiessling, Tissot, Stokes, and Stökl Ben~Ezra}]{eScriptorium}
Benjamin Kiessling, Robin Tissot, Peter Stokes, and Daniel Stökl Ben~Ezra. 2019.
\newblock \href {https://doi.org/10.1109/ICDARW.2019.10032} {escriptorium: An open source platform for historical document analysis}.
\newblock In \emph{2019 International Conference on Document Analysis and Recognition Workshops (ICDARW)}, volume~2, pages 19--19.

\bibitem[{Kölsch(2025)}]{annotationdb}
Andreas Kölsch. 2025.
\newblock Handwritten annotation detection dataset (annotationdb).
\newblock \url{https://tc11.cvc.uab.es/datasets/AnnotationDB_1}.
\newblock ID: AnnotationDB\_1, Accessed: 2025-05-18.

\bibitem[{Landesbibliothek et~al.(2024)Landesbibliothek, Ost, Stello, and Heim}]{reichenau}
Badische Landesbibliothek, Katharina Ost, Annika Stello, and Gerrit Heim. 2024.
\newblock \href {https://doi.org/10.5281/zenodo.11046062} {Training data incunabula reichenau}.

\bibitem[{Lassner et~al.(2021)Lassner, Coburger, Neudecker, and Baillot}]{ocr-data}
David Lassner, Julius Coburger, Clemens Neudecker, and Anne Baillot. 2021.
\newblock \href {https://doi.org/10.5281/zenodo.4742068} {Data set of the paper "publishing an ocr ground truth data set for reuse in an unclear copyright setting"}.

\bibitem[{{LECTAUREP} et~al.(2021){LECTAUREP}, Rostaing, Durand, and Chagué}]{lectaurep2021repertoires}
{LECTAUREP}, Aurélia Rostaing, Marc Durand, and Alix Chagué. 2021.
\newblock \href {https://doi.org/10.5072/zenodo.977691} {Notaires de paris – répertoires: Ground truth for various parisian registries of notary deeds (french 19th and 20th centuries)}.
\newblock Released on 2021-12-07.

\bibitem[{Lee et~al.(2023)Lee, Joshi, Turc, Hu, Liu, Eisenschlos, Khandelwal, Shaw, Chang, and Toutanova}]{lee2023pix2struct}
Kenton Lee, Mandar Joshi, Iulia~Raluca Turc, Hexiang Hu, Fangyu Liu, Julian~Martin Eisenschlos, Urvashi Khandelwal, Peter Shaw, Ming{-}Wei Chang, and Kristina Toutanova. 2023.
\newblock \href {https://proceedings.mlr.press/v202/lee23g.html} {Pix2struct: Screenshot parsing as pretraining for visual language understanding}.
\newblock In \emph{International Conference on Machine Learning, {ICML} 2023, 23-29 July 2023, Honolulu, Hawaii, {USA}}, volume 202 of \emph{Proceedings of Machine Learning Research}, pages 18893--18912. {PMLR}.

\bibitem[{Lefranc(2024)}]{arletta}
Lith Lefranc. 2024.
\newblock \href {https://doi.org/10.5281/zenodo.11191457} {Arletta: open-source handwritten text recognition models for historic dutch}.

\bibitem[{Levenshtein(1965)}]{levenshtein1966binary}
Vladimir~I. Levenshtein. 1965.
\newblock \href {https://api.semanticscholar.org/CorpusID:60827152} {Binary codes capable of correcting deletions, insertions, and reversals}.
\newblock \emph{Soviet physics. Doklady}, 10:707--710.

\bibitem[{Levenson(2023)}]{castilian}
Matthias~Gille Levenson. 2023.
\newblock \href {https://doi.org/10.5281/zenodo.8340483} {Towards a general open dataset and model for late medieval castilian text recognition (htr/ocr)}.

\bibitem[{Lewis et~al.(2006)Lewis, Agam, Argamon, Frieder, Grossman, and Heard}]{lewis2006building}
David Lewis, Gady Agam, Shlomo Argamon, Ophir Frieder, David Grossman, and Jefferson Heard. 2006.
\newblock Building a test collection for complex document information processing.
\newblock In \emph{Proceedings of the 29th annual international ACM SIGIR conference on Research and development in information retrieval}, pages 665--666.

\bibitem[{Li et~al.(2020)Li, Xu, Cui, Huang, Wei, Li, and Zhou}]{li2020docbank}
Minghao Li, Yiheng Xu, Lei Cui, Shaohan Huang, Furu Wei, Zhoujun Li, and Ming Zhou. 2020.
\newblock \href {https://doi.org/10.18653/v1/2020.coling-main.82} {{D}oc{B}ank: A benchmark dataset for document layout analysis}.
\newblock In \emph{Proceedings of the 28th International Conference on Computational Linguistics}, pages 949--960, Barcelona, Spain (Online). International Committee on Computational Linguistics.

\bibitem[{{Library of Congress}(2025)}]{altoxml}
{Library of Congress}. 2025.
\newblock {ALTO: Analyzed Layout and Text Object}.
\newblock \url{https://www.loc.gov/standards/alto/}.
\newblock Accessed: 2025-3-22.

\bibitem[{Liebl and Burghardt(2020)}]{liebl2020accuracycrnnslinebasedocr}
Bernhard Liebl and Manuel Burghardt. 2020.
\newblock \href {https://arxiv.org/abs/2008.02777} {On the accuracy of crnns for line-based ocr: A multi-parameter evaluation}.

\bibitem[{Limon-Bonnet et~al.(2024)Limon-Bonnet, Rostaing, and Chagué}]{limonbonnet2024bronod}
Marie-Françoise Limon-Bonnet, Aurélia Rostaing, and Alix Chagué. 2024.
\newblock \href {https://doi.org/10.5281/zenodo.10631356} {lectaurep-bronod, ground truth for maitre bronod's documents (french xviiith century)}.

\bibitem[{Liu et~al.(2024{\natexlab{a}})Liu, Li, Li, Li, Zhang, Shen, and Lee}]{liu2024llavanext}
Haotian Liu, Chunyuan Li, Yuheng Li, Bo~Li, Yuanhan Zhang, Sheng Shen, and Yong~Jae Lee. 2024{\natexlab{a}}.
\newblock \href {https://llava-vl.github.io/blog/2024-01-30-llava-next/} {Llava-next: Improved reasoning, ocr, and world knowledge}.

\bibitem[{Liu et~al.(2023{\natexlab{a}})Liu, Li, Wu, and Lee}]{llava2023}
Haotian Liu, Chunyuan Li, Qingyang Wu, and Yong~Jae Lee. 2023{\natexlab{a}}.
\newblock \href {http://papers.nips.cc/paper\_files/paper/2023/hash/6dcf277ea32ce3288914faf369fe6de0-Abstract-Conference.html} {Visual instruction tuning}.
\newblock In \emph{Advances in Neural Information Processing Systems 36: Annual Conference on Neural Information Processing Systems 2023, NeurIPS 2023, New Orleans, LA, USA, December 10 - 16, 2023}.

\bibitem[{Liu et~al.(2023{\natexlab{b}})Liu, Duan, Zhang, Li, Zhang, Zhao, Yuan, Wang, He, Liu, Chen, and Lin}]{liu2023mmbench}
Yuan Liu, Haodong Duan, Yuanhan Zhang, Bo~Li, Songyang Zhang, Wangbo Zhao, Yike Yuan, Jiaqi Wang, Conghui He, Ziwei Liu, Kai Chen, and Dahua Lin. 2023{\natexlab{b}}.
\newblock \href {https://arxiv.org/abs/2307.06281} {Mmbench: Is your multi-modal model an all-around player?}
\newblock \emph{Preprint}, arXiv:2307.06281.

\bibitem[{Liu et~al.(2024{\natexlab{b}})Liu, Li, Huang, Yang, Yu, Li, Yin, Liu, Jin, and Bai}]{liu2024ocrbench}
Yuliang Liu, Zhang Li, Mingxin Huang, Biao Yang, Wenwen Yu, Chunyuan Li, Xu-Cheng Yin, Cheng-Lin Liu, Lianwen Jin, and Xiang Bai. 2024{\natexlab{b}}.
\newblock Ocrbench: on the hidden mystery of ocr in large multimodal models.
\newblock \emph{Science China Information Sciences}, 67(12):220102.

\bibitem[{L’Eveque et~al.(2022)L’Eveque, Ekaterina, and Kasparian}]{leveque2022kovaleswky}
Zoé L’Eveque, Kate Ekaterina, and Anahide Kasparian. 2022.
\newblock \href {https://doi.org/10.5281/zenodo.6126633} {Projet kovaleswky – 1893}.

\bibitem[{Maheshwari et~al.(2022{\natexlab{a}})Maheshwari, Singh, Krishna, and Ramakrishnan}]{icdar2017}
Ayush Maheshwari, Nikhil Singh, Amrith Krishna, and Ganesh Ramakrishnan. 2022{\natexlab{a}}.
\newblock \href {https://doi.org/10.18653/v1/2022.findings-emnlp.466} {A benchmark and dataset for post-{OCR} text correction in {S}anskrit}.
\newblock In \emph{Findings of the Association for Computational Linguistics: EMNLP 2022}, pages 6258--6265, Abu Dhabi, United Arab Emirates. Association for Computational Linguistics.

\bibitem[{Maheshwari et~al.(2022{\natexlab{b}})Maheshwari, Singh, Krishna, and Ramakrishnan}]{postocr_sanskrit}
Ayush Maheshwari, Nikhil Singh, Amrith Krishna, and Ganesh Ramakrishnan. 2022{\natexlab{b}}.
\newblock \href {https://doi.org/10.18653/v1/2022.findings-emnlp.466} {A benchmark and dataset for post-{OCR} text correction in {S}anskrit}.
\newblock In \emph{Findings of the Association for Computational Linguistics: EMNLP 2022}, pages 6258--6265, Abu Dhabi, United Arab Emirates. Association for Computational Linguistics.

\bibitem[{Mandal et~al.(2025)Mandal, Talewar, Ahuja, and Juvatkar}]{Nanonets-OCR-S}
Souvik Mandal, Ashish Talewar, Paras Ahuja, and Prathamesh Juvatkar. 2025.
\newblock Nanonets-ocr-s: A model for transforming documents into structured markdown with intelligent content recognition and semantic tagging.

\bibitem[{Marthot-Santaniello and Hodel(2022)}]{marthotsantaniello2022zenon}
Isabelle Marthot-Santaniello and Tobias Hodel. 2022.
\newblock \href {https://doi.org/10.5281/zenodo.6565706} {Ground-truthed data set of zenon papyri for handwritten text recognition}.

\bibitem[{Mathew et~al.(2021)Mathew, Karatzas, and Jawahar}]{mathew2021docvqa}
Minesh Mathew, Dimosthenis Karatzas, and CV~Jawahar. 2021.
\newblock Docvqa: A dataset for vqa on document images.
\newblock In \emph{Proceedings of the IEEE/CVF winter conference on applications of computer vision}, pages 2200--2209.

\bibitem[{Maxime et~al.(2024)Maxime, Mathilde, Alix, and Marcello}]{htr_united_https-gitlabhuma-numfr-ecrinum-anthologia-htr-cpgr23}
Guénette Maxime, Verstraete Mathilde, Chagué Alix, and Vitali-Rosati Marcello. 2024.
\newblock \href {https://gitlab.huma-num.fr/ecrinum/anthologia/htr_cpgr23} {Ground truth for the palatine anthology}.

\bibitem[{Mazoue et~al.(2024)Mazoue, Clérice, and Chagué}]{cremma-mss16}
Anaïs Mazoue, Thibault Clérice, and Alix Chagué. 2024.
\newblock \href {https://github.com/HTR-United/CREMMA-MSS-16} {Cremma-mss-16}.

\bibitem[{Merkel-Hilf(2022)}]{merkelhilf2022devanagari}
Nicole Merkel-Hilf. 2022.
\newblock \href {https://doi.org/10.11588/DATA/EGOKEI} {Ground truth data for printed devanagari}.

\bibitem[{Michalcová et~al.(2022)Michalcová, Bazelides, Hajič, Pěnkavová, Maniaková, Kreisingerová, Filipová, hung Lu, and Dvořáková}]{michalcova2022paderov}
Anna Michalcová, Kamil Bazelides, Jan Hajič, Eliška Pěnkavová, Laura Maniaková, Hana Kreisingerová, Jitka Filipová, Chi hung Lu, and Martina Dvořáková. 2022.
\newblock \href {https://doi.org/10.5281/zenodo.7467034} {Padeřov-bible-handwriting-ground-truth: Initial release}.

\bibitem[{Microsoft et~al.(2025)Microsoft, :, Abouelenin, Ashfaq, Atkinson, Awadalla, Bach, Bao, Benhaim, Cai, Chaudhary, Chen, Chen, Chen, Chen, Chen, Chen, ling Chen, Dai, Dai, Fan, Gao, Gao, Garg, Goswami, Hao, Hendy, Hu, Jin, Khademi, Kim, Kim, Lee, Li, Li, Liang, Lin, Lin, Liu, Liu, Lopez, Luo, Madan, Mazalov, Mitra, Mousavi, Nguyen, Pan, Perez-Becker, Platin, Portet, Qiu, Ren, Ren, Roy, Shang, Shen, Singhal, Som, Song, Sych, Vaddamanu, Wang, Wang, Wang, Wu, Xu, Xu, Yang, Yang, Yu, Zabir, Zhang, Zhang, Zhang, and Zhou}]{microsoft2025phi4mini}
Microsoft, :, Abdelrahman Abouelenin, Atabak Ashfaq, Adam Atkinson, Hany Awadalla, Nguyen Bach, Jianmin Bao, Alon Benhaim, Martin Cai, Vishrav Chaudhary, Congcong Chen, Dong Chen, Dongdong Chen, Junkun Chen, Weizhu Chen, Yen-Chun Chen, Yi~ling Chen, Qi~Dai, and 57 others. 2025.
\newblock \href {https://arxiv.org/abs/2503.01743} {Phi-4-mini technical report: Compact yet powerful multimodal language models via mixture-of-loras}.
\newblock \emph{Preprint}, arXiv:2503.01743.

\bibitem[{{Microsoft}(2024)}]{azureocr}
{Microsoft}. 2024.
\newblock {Azure AI Document Intelligence}.
\newblock \url{https://learn.microsoft.com/en-us/azure/ai-services/document-intelligence/}.
\newblock Accessed: 2025-03-20.

\bibitem[{{Mistral AI}(2025)}]{mistral2025ocr}
{Mistral AI}. 2025.
\newblock {Mistral OCR}.
\newblock \url{https://mistral.ai/news/mistral-ocr}.
\newblock Accessed: 2025-03-24.

\bibitem[{Muehlberger and Hackl(2019)}]{newseye_austrian}
Guenter Muehlberger and Guenter Hackl. 2019.
\newblock \href {https://doi.org/10.5281/zenodo.3387369} {Newseye / read ocr training dataset from austrian newspapers (19th c.)}.

\bibitem[{Muehlberger and Hackl(2020)}]{newseye_french}
Guenter Muehlberger and Guenter Hackl. 2020.
\newblock \href {https://doi.org/10.5281/zenodo.4293602} {Newseye / read ocr training dataset from french newspapers (18th, 19th, early 20th c.)}.

\bibitem[{Muehlberger and Hackl(2021{\natexlab{a}})}]{newseye_finnish}
Guenter Muehlberger and Guenter Hackl. 2021{\natexlab{a}}.
\newblock \href {https://doi.org/10.5281/zenodo.4599472} {Newseye / read ocr training dataset from finnish newspapers (18th, 19th, early 20th c.)}.

\bibitem[{Muehlberger and Hackl(2021{\natexlab{b}})}]{newseye_swedish}
Guenter Muehlberger and Guenter Hackl. 2021{\natexlab{b}}.
\newblock \href {https://doi.org/10.5281/zenodo.4599624} {Newseye / read ocr training dataset from swedish newspapers (18th, 19th, early 20th c.)}.

\bibitem[{Najam and Faizullah(2024)}]{scarce}
Rayyan Najam and Safiullah Faizullah. 2024.
\newblock \href {https://doi.org/10.17632/xz6f8bw3w8.1} {A scarce dataset for ancient arabic handwritten text recognition}.
\newblock Mendeley Data, V1.
\newblock Accessed: 2025-03-25.

\bibitem[{Neudecker et~al.(2021)Neudecker, Baierer, Gerber, Clausner, Antonacopoulos, and Pletschacher}]{neudecker2021ocr}
Clemens Neudecker, Konstantin Baierer, Mike Gerber, Christian Clausner, Apostolos Antonacopoulos, and Stefan Pletschacher. 2021.
\newblock \href {https://doi.org/10.1145/3476887.3476888} {A survey of ocr evaluation tools and metrics}.
\newblock In \emph{Proceedings of the 6th International Workshop on Historical Document Imaging and Processing}, HIP '21, page 13–18, New York, NY, USA. Association for Computing Machinery.

\bibitem[{No{\"e}mie et~al.(2022)No{\"e}mie, Salah, and Vidal-Gor{\`e}ne}]{tarima}
Lucas No{\"e}mie, Cl{\'e}ment Salah, and Chahan Vidal-Gor{\`e}ne. 2022.
\newblock \href {https://enc.hal.science/hal-03874725} {{New Results for the Text Recognition of Arabic Maghrib{\=i} Manuscripts - Managing an Under-resourced Script}}.
\newblock Working paper or preprint.

\bibitem[{Norindr et~al.(2023)Norindr, Mikhalchuk, Clérice, and Chagué}]{norindr2023htromance}
Jade Norindr, Anna Mikhalchuk, Thibault Clérice, and Alix Chagué. 2023.
\newblock \href {https://github.com/HTRomance-Project/modern-roman-languages} {Htromance, modern language corpus of ground-truth for handwritten text recognition and layout segmentation}.

\bibitem[{Novotný et~al.(2021)Novotný, Seidlová, Vrabcová, and Horák}]{ahisto}
Vít Novotný, Kristýna Seidlová, Tereza Vrabcová, and Aleš Horák. 2021.
\newblock \href {https://nlp.fi.muni.cz/raslan/2021/paper10.pdf} {When tesseract brings friends: Layout analysis, language identification, and super-resolution in the optical character recognition of medieval texts}.
\newblock In \emph{Proceedings of Recent Advances in Slavonic Natural Language Processing, {RASLAN} 2021}, pages 91--100. Tribun {EU}.

\bibitem[{{Numind}(2025)}]{numind_2025_numarkdown_8b}
{Numind}. 2025.
\newblock \href {https://huggingface.co/numind/NuMarkdown-8B-Thinking} {{NuMarkdown-8B-Thinking}}.
\newblock Model page.

\bibitem[{{NVIDIA}(2025)}]{nvidia_2025_llama_3_1_nemotron_nano_vl_8b_v1}
{NVIDIA}. 2025.
\newblock \href {https://huggingface.co/nvidia/Llama-3.1-Nemotron-Nano-VL-8B-V1} {{Llama-3.1-Nemotron-Nano-VL-8B-V1}}.
\newblock Model page.

\bibitem[{Oberbichler et~al.(2022)Oberbichler, Boro{\c{s}}, Doucet, Marjanen, Pfanzelter, Rautiainen, Toivonen, and Tolonen}]{oberbichler2022integrated}
Sarah Oberbichler, Emanuela Boro{\c{s}}, Antoine Doucet, Jani Marjanen, Eva Pfanzelter, Juha Rautiainen, Hannu Toivonen, and Mikko Tolonen. 2022.
\newblock Integrated interdisciplinary workflows for research on historical newspapers: Perspectives from humanities scholars, computer scientists, and librarians.
\newblock \emph{Journal of the Association for Information Science and Technology}, 73(2):225--239.

\bibitem[{OpenAI(2025{\natexlab{a}})}]{openai2025gpt5}
OpenAI. 2025{\natexlab{a}}.
\newblock \href {https://openai.com/index/introducing-gpt-5/} {Introducing gpt-5}.
\newblock Accessed: 2025-09-18.

\bibitem[{OpenAI(2025{\natexlab{b}})}]{openai2024-o3-o4mini}
OpenAI. 2025{\natexlab{b}}.
\newblock \href {https://openai.com/index/introducing-o3-and-o4-mini/} {Introducing openai o3 and o4-mini}.
\newblock Accessed: 2025-09-18.

\bibitem[{OpenAI et~al.(2023)OpenAI, Achiam, Adler, Agarwal, Ahmad, Akkaya, Aleman, Almeida, Altenschmidt, Altman, Anadkat, Avila, Babuschkin, Balaji, Balcom, Baltescu, Bao, Bavarian, Belgum, Bello, Berdine, Bernadett-Shapiro, Berner, Bogdonoff, Boiko, Boyd, Brakman, Brockman, Brooks, Brundage, Button, Cai, Campbell, Cann, Carey, Carlson, Carmichael, Chan, Chang, Chantzis, Chen, Chen, Chen, Chen, Chen, Chess, Cho, Chu, Chung, Cummings, Currier, Dai, Decareaux, Degry, Deutsch, Deville, Dhar, Dohan, Dowling, Dunning, Ecoffet, Eleti, Eloundou, Farhi, Fedus, Felix, Fishman, Forte, Fulford, Gao, Georges, Gibson, Goel, Gogineni, Goh, Gontijo-Lopes, Gordon, Grafstein, Gray, Greene, Gross, Gu, Guo, Hallacy, Han, Harris, He, Heaton, Heidecke, Hesse, Hickey, Hickey, Hoeschele, Houghton, Hsu, Hu, Hu, Huizinga, Jain, Jain, Jang, Jiang, Jiang, Jin, Jin, Jomoto, Jonn, Jun, Kaftan, Łukasz Kaiser, Kamali, Kanitscheider, Keskar, Khan, Kilpatrick, Kim, Kim, Kim, Kirchner, Kiros, Knight, Kokotajlo, Łukasz Kondraciuk,
  Kondrich, Konstantinidis, Kosic, Krueger, Kuo, Lampe, Lan, Lee, Leike, Leung, Levy, Li, Lim, Lin, Lin, Litwin, Lopez, Lowe, Lue, Makanju, Malfacini, Manning, Markov, Markovski, Martin, Mayer, Mayne, McGrew, McKinney, McLeavey, McMillan, McNeil, Medina, Mehta, Menick, Metz, Mishchenko, Mishkin, Monaco, Morikawa, Mossing, Mu, Murati, Murk, Mély, Nair, Nakano, Nayak, Neelakantan, Ngo, Noh, Ouyang, O'Keefe, Pachocki, Paino, Palermo, Pantuliano, Parascandolo, Parish, Parparita, Passos, Pavlov, Peng, Perelman, de~Avila Belbute~Peres, Petrov, de~Oliveira~Pinto, Michael, Pokorny, Pokrass, Pong, Powell, Power, Power, Proehl, Puri, Radford, Rae, Ramesh, Raymond, Real, Rimbach, Ross, Rotsted, Roussez, Ryder, Saltarelli, Sanders, Santurkar, Sastry, Schmidt, Schnurr, Schulman, Selsam, Sheppard, Sherbakov, Shieh, Shoker, Shyam, Sidor, Sigler, Simens, Sitkin, Slama, Sohl, Sokolowsky, Song, Staudacher, Such, Summers, Sutskever, Tang, Tezak, Thompson, Tillet, Tootoonchian, Tseng, Tuggle, Turley, Tworek, Uribe, Vallone,
  Vijayvergiya, Voss, Wainwright, Wang, Wang, Wang, Ward, Wei, Weinmann, Welihinda, Welinder, Weng, Weng, Wiethoff, Willner, Winter, Wolrich, Wong, Workman, Wu, Wu, Wu, Xiao, Xu, Yoo, Yu, Yuan, Zaremba, Zellers, Zhang, Zhang, Zhao, Zheng, Zhuang, Zhuk, and Zoph}]{openai2024gpt4technicalreport}
OpenAI, Josh Achiam, Steven Adler, Sandhini Agarwal, Lama Ahmad, Ilge Akkaya, Florencia~Leoni Aleman, Diogo Almeida, Janko Altenschmidt, Sam Altman, Shyamal Anadkat, Red Avila, Igor Babuschkin, Suchir Balaji, Valerie Balcom, Paul Baltescu, Haiming Bao, Mohammad Bavarian, Jeff Belgum, and 262 others. 2023.
\newblock \href {https://arxiv.org/abs/2303.08774} {Gpt-4 technical report}.

\bibitem[{OpenAI et~al.(2025)OpenAI, Kumar, Yu, Hallman, Pokrass, Goucher, Ganesh, Cheng, McKinzie, Zhang, Koch, Wei, Medina, Wong, Kavanaugh, Bekerman, Hu, Ren, Singal, Kiros, Ai, Lin, Chien, McGrath, Lee, Wang, Lu, Georgiev, Luther, Jing, Schwarzer, Castro, Keskar, Lopes, Zhao, Chen, Sanjeev, Gordon, Sanders, Zhou, Song, Xie, Jin, and Zhang}]{openai2025gpt41}
OpenAI, Ananya Kumar, Jiahui Yu, John Hallman, Michelle Pokrass, Adam Goucher, Adi Ganesh, Bowen Cheng, Brandon McKinzie, Brian Zhang, Chris Koch, Colin Wei, David Medina, Edmund Wong, Erin Kavanaugh, Florent Bekerman, Haitang Hu, Hongyu Ren, Ishaan Singal, and 25 others. 2025.
\newblock \href {https://openai.com/blog/introducing-gpt-4-1-in-the-api} {Introducing gpt-4.1 in the api}.
\newblock Accessed: 2025-05-18.

\bibitem[{Ouyang et~al.(2024)Ouyang, Qu, Zhou, Zhu, Zhang, Lin, Wang, Zhao, Jiang, Zhao et~al.}]{ouyang2024omnidocbench}
Linke Ouyang, Yuan Qu, Hongbin Zhou, Jiawei Zhu, Rui Zhang, Qunshu Lin, Bin Wang, Zhiyuan Zhao, Man Jiang, Xiaomeng Zhao, and 1 others. 2024.
\newblock \href {https://arxiv.org/abs/2412.07626} {Omnidocbench: Benchmarking diverse pdf document parsing with comprehensive annotations}.
\newblock \emph{ArXiv preprint}, abs/2412.07626.

\bibitem[{O’Neill and Hill(2022)}]{pracalit}
Alexander~James O’Neill and Nathan Hill. 2022.
\newblock \href {https://doi.org/10.5334/johd.90} {Text recognition for nepalese manuscripts in pracalit script}.
\newblock \emph{Journal of Open Humanities Data}.

\bibitem[{Pantke et~al.(2014)Pantke, Dennhardt, Fecker, Märgner, and Fingscheidt}]{hadara}
Werner Pantke, Martin Dennhardt, Daniel Fecker, Volker Märgner, and Tim Fingscheidt. 2014.
\newblock \href {https://doi.org/10.1109/ICFHR.2014.11} {An historical handwritten arabic dataset for segmentation-free word spotting - hadara80p}.
\newblock In \emph{2014 14th International Conference on Frontiers in Handwriting Recognition}, pages 15--20.

\bibitem[{Papadopoulos et~al.(2013)Papadopoulos, Pletschacher, Clausner, and Antonacopoulos}]{impact}
Christos Papadopoulos, Stefan Pletschacher, Christian Clausner, and Apostolos Antonacopoulos. 2013.
\newblock \href {https://doi.org/10.1145/2501115.2501130} {The impact dataset of historical document images}.
\newblock In \emph{Proceedings of the 2nd International Workshop on Historical Document Imaging and Processing}, HIP '13, page 123–130, New York, NY, USA. Association for Computing Machinery.

\bibitem[{Papazoglou et~al.(2020)Papazoglou, Pratikakis, Markou, and Tsochatzidis}]{papazoglou2020eparchos}
Aleksandros Papazoglou, Ioannis Pratikakis, Kleopatra Markou, and Lazaros Tsochatzidis. 2020.
\newblock \href {https://doi.org/10.5281/zenodo.4095301} {Eparchos – historical greek handwritten document dataset}.

\bibitem[{Pascual et~al.(2022)Pascual, d'Espèrey, and Gabay}]{pascual2022chavigny}
Margot Pascual, Louis-Fiacre~Franchet d'Espèrey, and Simon Gabay. 2022.
\newblock \href {https://doi.org/10.5281/zenodo.6126655} {Château de chavigny}.

\bibitem[{Penteliuc(2022)}]{romanian_transitional}
Marius Penteliuc. 2022.
\newblock 19th-century romanian transitional script.
\newblock \url{https://www.kaggle.com/datasets/mariuspenteliuc/rts-ocr}.
\newblock Accessed: 2025-03-24.

\bibitem[{Pham(2020)}]{jcsr}
Kim Pham. 2020.
\newblock \href {https://doi.org/10.5281/zenodo.4243023} {University of denver collections as data - htr train and validation set jcrs\_2020\_5\_27}.

\bibitem[{Pinche(2022)}]{cremma-medieval}
Ariane Pinche. 2022.
\newblock \href {https://doi.org/10.5281/zenodo.5235185} {Cremma medieval}.

\bibitem[{Pinche et~al.(2022)Pinche, Gabay, Leroy, and Christensen}]{pinche2022htr15c}
Ariane Pinche, Simon Gabay, Noé Leroy, and Kelly Christensen. 2022.
\newblock \href {https://github.com/Gallicorpora/HTR-MSS-15e-Siecle} {Données htr manuscrits du 15e siècle}.

\bibitem[{Pinche and Pierreville(2023)}]{pinche2023fabliaux}
Ariane Pinche and Corinne Pierreville. 2023.
\newblock \href {https://github.com/CIHAM-HTR/Fabliaux/data} {Fabliaux}.

\bibitem[{Platanou et~al.(2022)Platanou, Pavlopoulos, and Papaioannou}]{platanou2022paleogreek}
Paraskevi Platanou, John Pavlopoulos, and Georgios Papaioannou. 2022.
\newblock Handwritten paleographic greek text recognition dataset (century-based).
\newblock This dataset is part of the work: Platanou, P., Pavlopoulos, J., and Papaioannou, G. (2022). Handwritten Paleographic Greek Text Recognition: A Century-Based Approach. In *Proceedings of the Thirteenth Language Resources and Evaluation Conference*, pp. 6585–6589, Marseille, France: ELRA.

\bibitem[{Pletschacher and Antonacopoulos(2010)}]{pagexml}
Stefan Pletschacher and Apostolos Antonacopoulos. 2010.
\newblock The {PAGE} ({P}age {A}nalysis and {G}round-{T}ruth {E}lements) {F}ormat {F}ramework.
\newblock In \emph{2010 20th International Conference on Pattern Recognition}, pages 257--260. IEEE.

\bibitem[{Poznanski et~al.(2025)Poznanski, Rangapur, Borchardt, Dunkelberger, Huff, Lin, Rangapur, Wilhelm, Lo, and Soldaini}]{olmocr}
Jake Poznanski, Aman Rangapur, Jon Borchardt, Jason Dunkelberger, Regan Huff, Daniel Lin, Aman Rangapur, Christopher Wilhelm, Kyle Lo, and Luca Soldaini. 2025.
\newblock \href {https://arxiv.org/abs/2502.18443} {olmocr: Unlocking trillions of tokens in pdfs with vision language models}.

\bibitem[{Pradier et~al.(2022)Pradier, Gabay, Janès, Oeconomo, and Kervegan}]{pradier2022fonduecat}
Frederine Pradier, Simon Gabay, Juliette Janès, Esteban~Sánchez Oeconomo, and Paul Kervegan. 2022.
\newblock \href {https://github.com/FoNDUE-HTR/FONDUE-MLT-CAT} {Fondue – datasets for historical catalogues}.

\bibitem[{Pratikakis et~al.(2021)Pratikakis, Papazoglou, Symeonidis, and Tsochatzidis}]{pratikakis2021stavronikita}
Ioannis Pratikakis, Aleksandros Papazoglou, Symeon Symeonidis, and Lazaros Tsochatzidis. 2021.
\newblock \href {https://doi.org/10.5281/zenodo.5578251} {Stavronikita monastery greek handwritten document collection no.114}.

\bibitem[{{PRHLT Research Center}(2022)}]{prhlt2022hisclima}
{PRHLT Research Center}. 2022.
\newblock \href {https://doi.org/10.5281/zenodo.7442971} {Hisclima dataset}.

\bibitem[{Quirós et~al.(2018)Quirós, Serrano, Bosch, Toselli, Congost, Saguer, and Vidal}]{notarial}
Lorenzo Quirós, Lluís Serrano, Vicente Bosch, Alejandro~H. Toselli, Rosa Congost, Enric Saguer, and Enrique Vidal. 2018.
\newblock \href {https://doi.org/10.5281/zenodo.1322666} {Oficio de hipotecas de girona. a dataset of spanish notarial deeds (18th century) for handwritten text recognition and layout analysis of historical documents.}

\bibitem[{Rabaev et~al.(2019)Rabaev, Bakarat, and El-Sana}]{pinkas}
Irina Rabaev, Berat~Kurar Bakarat, and Jihad El-Sana. 2019.
\newblock \href {https://doi.org/10.1109/ICDAR.2019.00122\_dataset} {The pinkas dataset}.

\bibitem[{Radford et~al.(2021)Radford, Kim, Hallacy, Ramesh, Goh, Agarwal, Sastry, Askell, Mishkin, Clark, Krueger, and Sutskever}]{radford2021learning}
Alec Radford, Jong~Wook Kim, Chris Hallacy, Aditya Ramesh, Gabriel Goh, Sandhini Agarwal, Girish Sastry, Amanda Askell, Pamela Mishkin, Jack Clark, Gretchen Krueger, and Ilya Sutskever. 2021.
\newblock \href {http://proceedings.mlr.press/v139/radford21a.html} {Learning transferable visual models from natural language supervision}.
\newblock In \emph{Proceedings of the 38th International Conference on Machine Learning, {ICML} 2021, 18-24 July 2021, Virtual Event}, volume 139 of \emph{Proceedings of Machine Learning Research}, pages 8748--8763. {PMLR}.

\bibitem[{{Research Group Texts for the History of Spanish [GITHE]}(2022)}]{codea}
{Research Group Texts for the History of Spanish [GITHE]}. 2022.
\newblock {CODEA+ 2022 (Corpus of Spanish documents prior to 1900)}.
\newblock \url{https://doi.org/10.37536/CODEA.2015}.
\newblock [online; accessed 2025-03-09].

\bibitem[{Reul et~al.(2019)Reul, Christ, Hartelt, Balbach, Wehner, Springmann, Wick, Grundig, B{\"u}ttner, and Puppe}]{reul2019ocr4all}
Christian Reul, Dennis Christ, Alexander Hartelt, Nico Balbach, Maximilian Wehner, Uwe Springmann, Christoph Wick, Christine Grundig, Andreas B{\"u}ttner, and Frank Puppe. 2019.
\newblock Ocr4all—an open-source tool providing a (semi-) automatic ocr workflow for historical printings.
\newblock \emph{Applied Sciences}, 9(22):4853.

\bibitem[{Rogos-Hebda(2025)}]{rogos2025making}
Justyna Rogos-Hebda. 2025.
\newblock Making sense of abbreviation in late middle english literary manuscripts.
\newblock In \emph{Abbreviating Middle English: Scribal Practices, Visual Texts and Medieval Multimodalities}, pages 157--169. Springer.

\bibitem[{Romanello et~al.(2021)Romanello, Sven, and Robertson}]{romanello_optical_2021}
Matteo Romanello, Najem-Meyer Sven, and Bruce Robertson. 2021.
\newblock \href {https://doi.org/10.1145/3476887.3476911} {Optical {{Character Recognition}} of 19th {{Century Classical Commentaries}}: The {{Current State}} of {{Affairs}}}.
\newblock In \emph{The 6th {{International Workshop}} on {{Historical Document Imaging}} and {{Processing}} ({{HIP}} '21)}, {Lausanne}. {Association for Computing Machinery}.

\bibitem[{Romein et~al.(2020)Romein, {de Gruijter}, and Veldhoen}]{entangled}
C.A. Romein, Michel {de Gruijter}, and {Sara Floor} Veldhoen. 2020.
\newblock The datafication of early modern ordinances.
\newblock \emph{DH Benelux Journal}, 2.

\bibitem[{Romero et~al.(2019)Romero, Toselli, Vidal, S\'{a}nchez, Alonso, and Marqu\'{e}s}]{diplomatic2019}
Ver\'{o}nica Romero, Alejandro~H. Toselli, Enrique Vidal, Joan~Andreu S\'{a}nchez, Carlos Alonso, and Lourdes Marqu\'{e}s. 2019.
\newblock \href {https://doi.org/10.1007/978-3-030-30754-7_11} {Modern vs diplomatic transcripts for historical handwritten text recognition}.
\newblock In \emph{New Trends in Image Analysis and Processing – ICIAP 2019: ICIAP International Workshops, BioFor, PatReCH, e-BADLE, DeepRetail, and Industrial Session, Trento, Italy, September 9–10, 2019, Revised Selected Papers}, page 103–114, Berlin, Heidelberg. Springer-Verlag.

\bibitem[{Rostaing et~al.(2024)Rostaing, Denis, and Chagué}]{rostaing2024mariages}
Aurélia Rostaing, Nathalie Denis, and Alix Chagué. 2024.
\newblock \href {https://doi.org/10.5281/zenodo.10632593} {Lectaurep-mariages-et-divorces: Ground truth for the enregistrements des contrats de mariages et des séparations et divorces in paris (french 19th century)}.
\newblock Released on 2024-02-07.

\bibitem[{{Sacro Convento di San Francesco in Assisi}(2024)}]{assisi_fondo_antico_sacro_convento}
{Sacro Convento di San Francesco in Assisi}. 2024.
\newblock \href {https://www.internetculturale.it/it/1175/assisi-fondo-antico-del-sacro-convento-mediatheca-franciscana} {Assisi, fondo antico del sacro convento – mediatheca franciscana}.
\newblock Internet Culturale. Central Institute for the Union Catalogue of Italian Libraries (ICCU), Ministry of Culture.

\bibitem[{Saeed et~al.(2024)Saeed, Chan, Mijar, Moukarzel, Habchi, Younes, Elias, Wong, and Khater}]{muharaf}
Mehreen Saeed, Adrian Chan, Anupam Mijar, Joseph Moukarzel, Georges Habchi, Carlos Younes, Amin Elias, Chau-Wai Wong, and Akram Khater. 2024.
\newblock \href {https://doi.org/10.5281/zenodo.11492215} {Muharaf-public}.

\bibitem[{Saini et~al.(2019)Saini, Dobson, Morrey, Liwicki, and Simistira~Liwicki}]{icdar}
Rajkumar Saini, Derek Dobson, Jon Morrey, Marcus Liwicki, and Foteini Simistira~Liwicki. 2019.
\newblock \href {https://doi.org/10.1109/ICDAR.2019.00241} {Icdar 2019 historical document reading challenge on large structured chinese family records}.
\newblock In \emph{2019 International Conference on Document Analysis and Recognition (ICDAR)}, pages 1499--1504.

\bibitem[{{Salamanca School Project}(2025)}]{salamanca}
{Salamanca School Project}. 2025.
\newblock The school of salamancal.
\newblock \url{https://www.salamanca.school/works.html}.
\newblock Accessed: 2025-05-18.

\bibitem[{Sarbach-Pulicani et~al.(2022)Sarbach-Pulicani, Miaille, Escoda, Saïag, and Gabay}]{sarbachpulicani2022corse}
Vincent Sarbach-Pulicani, Théophile Miaille, Adrien Escoda, Violette Saïag, and Simon Gabay. 2022.
\newblock \href {https://doi.org/10.5281/zenodo.6126641} {Ocr d'une poésie corse}.

\bibitem[{Schaefer and Litvine(2023)}]{schaefer2023hooker}
John Schaefer and Alexis Litvine. 2023.
\newblock \href {https://doi.org/10.5281/zenodo.8038689} {Joseph hooker htr model}.

\bibitem[{Schultze et~al.(2024)Schultze, Kerkfeld, Kuebart, Weber, Wolter, and Selgert}]{chronicling_germany}
Christian Schultze, Niklas Kerkfeld, Kara Kuebart, Princilia Weber, Moritz Wolter, and Felix Selgert. 2024.
\newblock \href {https://arxiv.org/abs/2401.16845} {Chronicling germany: An annotated historical newspaper dataset}.

\bibitem[{Shi et~al.(2023)Shi, Liu, Peng, Jian, Huang, and Jin}]{m5hisdoc}
Yongxin Shi, Chongyu Liu, Dezhi Peng, Cheng Jian, Jiarong Huang, and Lianwen Jin. 2023.
\newblock \href {http://papers.nips.cc/paper\_files/paper/2023/hash/f7b424d242cc6bb7708cff241367334d-Abstract-Datasets\_and\_Benchmarks.html} {M5hisdoc: {A} large-scale multi-style chinese historical document analysis benchmark}.
\newblock In \emph{Advances in Neural Information Processing Systems 36: Annual Conference on Neural Information Processing Systems 2023, NeurIPS 2023, New Orleans, LA, USA, December 10 - 16, 2023}.

\bibitem[{Solfrini et~al.(2023)Solfrini, Gabay, Gross, Beaulnes, Oliveira, Dejouy, and Camillocci}]{solfrini2023setaf}
Sonia Solfrini, Simon Gabay, Geneviève Gross, Pierre-Olivier Beaulnes, Aurélia~Marques Oliveira, Mylène Dejouy, and Daniela~Solfaroli Camillocci. 2023.
\newblock \href {https://github.com/SETAFDH/HTR-SETAF-Jean-Michel} {Données ocr et segmentation des imprimés de jean michel (projet setaf)}.
\newblock OCR data for the SETAF project: 16th-century French prints in Gothic characters.

\bibitem[{Springmann and Fink(2018)}]{fraktur}
Uwe Springmann and Fabian Fink. 2018.
\newblock Ground truth (gt) data for fraktur/gothic prints.
\newblock \url{https://github.com/ubtue/gt-fraktur}.
\newblock Accessed: 2025-03-25.

\bibitem[{Stotz and Ströbel(2021)}]{gwalther}
Peter Stotz and Phillip Ströbel. 2021.
\newblock \href {https://doi.org/10.5281/zenodo.4780947} {bullinger-digital/gwalther-handwriting-ground- truth: Initial release}.

\bibitem[{Ströbel and Clematide(2019)}]{nzz}
Phillip Ströbel and Simon Clematide. 2019.
\newblock Improving ocr of black letter in historical newspapers: The unreasonable effectiveness of htr models on low-resolution images.
\newblock In \emph{Proceedings of the Digital Humanities 2019, (DH2019)}.
\newblock Accepted.

\bibitem[{Stökl Ben~Ezra et~al.(2021)Stökl Ben~Ezra, Brown-DeVost, Jablonski, Kiessling, Lolli, and Lapin}]{stokl_ben_ezra_2021_5167263}
Daniel Stökl Ben~Ezra, Bronson Brown-DeVost, Pawel Jablonski, Benjamin Kiessling, Elena Lolli, and Hayim Lapin. 2021.
\newblock \href {https://doi.org/10.5281/zenodo.5167263} {Biblia - an open annotated dataset}.

\bibitem[{Sulaiman et~al.(2019)Sulaiman, Omar, and Nasrudin}]{jimaging5040048}
Alaa Sulaiman, Khairuddin Omar, and Mohammad~F. Nasrudin. 2019.
\newblock \href {https://doi.org/10.3390/jimaging5040048} {Degraded historical document binarization: A review on issues, challenges, techniques, and future directions}.
\newblock \emph{Journal of Imaging}, 5(4).

\bibitem[{Sánchez et~al.(2017)Sánchez, Toselli, Romero, and Vidal}]{icdar2015}
J.A. Sánchez, A.H. Toselli, V.~Romero, and E.~Vidal. 2017.
\newblock \href {https://doi.org/10.5281/zenodo.248733} {Icdar 2015 competition htrts: Handwritten text recognition on the transcriptorium dataset}.

\bibitem[{Sánchez et~al.(2016)Sánchez, Romero, Toselli, and Vidal}]{icfhr2016}
Joan~Andreu Sánchez, Verónica Romero, Alejandro~H. Toselli, and Enrique Vidal. 2016.
\newblock \href {https://doi.org/10.5281/zenodo.218236} {Read dataset bozen}.

\bibitem[{Team et~al.(2025)Team, Karlinsky, Arbelle, Daniels, Nassar, Alfassi, Wu, Schwartz, Joshi, Kondic, Shabtay, Li, Herzig, Abedin, Perek, Harary, Barzelay, Goldfarb, Oliva, Wieles, Bhattacharjee, Huang, Auer, Gutfreund, Beymer, Wood, Kuehne, Hansen, Shtok, Wong, Bathen, Mishra, Lysak, Dolfi, Yurochkin, Livathinos, Harel, Azulai, Naparstek, de~Lima, Panda, Doveh, Gupta, Das, Zawad, Kim, He, Brooks, Goodhart, Govindjee, Leist, Ibrahim, Soffer, Cox, Soule, Lastras, Desai, Ofek-koifman, Raghavan, Syeda-Mahmood, Staar, Drory, and Feris}]{team2025granite}
Granite~Vision Team, Leonid Karlinsky, Assaf Arbelle, Abraham Daniels, Ahmed Nassar, Amit Alfassi, Bo~Wu, Eli Schwartz, Dhiraj Joshi, Jovana Kondic, Nimrod Shabtay, Pengyuan Li, Roei Herzig, Shafiq Abedin, Shaked Perek, Sivan Harary, Udi Barzelay, Adi~Raz Goldfarb, Aude Oliva, and 44 others. 2025.
\newblock \href {https://arxiv.org/abs/2502.09927} {Granite vision: a lightweight, open-source multimodal model for enterprise intelligence}.
\newblock \emph{Preprint}, arXiv:2502.09927.

\bibitem[{Teklia(2025{\natexlab{a}})}]{laliberte}
Teklia. 2025{\natexlab{a}}.
\newblock Finlam la liberté newspaper dataset.
\newblock \url{https://huggingface.co/datasets/Teklia/Newspapers-finlam-La-Liberte}.
\newblock Accessed: 2025-05-18.

\bibitem[{Teklia(2025{\natexlab{b}})}]{finlam}
Teklia. 2025{\natexlab{b}}.
\newblock Finlam newspaper segmentation dataset.
\newblock \url{https://huggingface.co/datasets/Teklia/Newspapers-finlam}.
\newblock Accessed: 2025-05-18.

\bibitem[{Tommasi(2024)}]{tommasi2024episearch}
Tatiana Tommasi. 2024.
\newblock \href {https://github.com/vedph/episearch-htr} {episearch-htr: Transcriptions, metadata, and htr models from giovanni antonio astori’s correspondence and epigraphic manuscript}.

\bibitem[{Toselli and Vidal(2021)}]{fcr}
Alejandro~H. Toselli and Enrique Vidal. 2021.
\newblock \href {https://doi.org/10.5281/zenodo.4767732} {The finnish court records dataset}.

\bibitem[{Turski et~al.(2023)Turski, Stanis{\l}awek, Kaczmarek, Dyda, and Grali{\'n}ski}]{turski2023ccpdf}
Micha{\l} Turski, Tomasz Stanis{\l}awek, Karol Kaczmarek, Pawe{\l} Dyda, and Filip Grali{\'n}ski. 2023.
\newblock Ccpdf: Building a high quality corpus for visually rich documents from web crawl data.
\newblock In \emph{International Conference on Document Analysis and Recognition}, pages 348--365. Springer.

\bibitem[{{Tübingen University Library}(2023)}]{tuebingen2023malayalam}
{Tübingen University Library}. 2023.
\newblock \href {https://doi.org/10.11588/DATA/L2KRZO} {Ground truth data for printed malayalam}.

\bibitem[{Valy et~al.(2017)Valy, Verleysen, Chhun, and Burie}]{sleuk}
Dona Valy, Michel Verleysen, Sophea Chhun, and Jean-Christophe Burie. 2017.
\newblock \href {https://doi.org/10.1145/3151509.3151510} {A new khmer palm leaf manuscript dataset for document analysis and recognition: Sleukrith set}.
\newblock In \emph{Proceedings of the 4th International Workshop on Historical Document Imaging and Processing}, HIP '17, page 1–6, New York, NY, USA. Association for Computing Machinery.

\bibitem[{Van~Kote et~al.(2024)Van~Kote, Faure, Norindr, Clérice, and Chagué}]{cremma-mss18}
Elsa Van~Kote, Margaux Faure, Jade Norindr, Thibault Clérice, and Alix Chagué. 2024.
\newblock \href {https://github.com/HTR-United/CREMMA-MSS-18} {Cremma-mss-18}.

\bibitem[{Vandyck et~al.(2024)Vandyck, Dummer, Kuhry, Koczarski, Besson, Chevalier-Royet, and Levenson}]{vandyck2024transcriboquest}
Caroline Vandyck, Jessie Dummer, Emmanuelle Kuhry, Zdzislaw Koczarski, Sylvain Besson, Caroline Chevalier-Royet, and Matthias~Gille Levenson. 2024.
\newblock \href {https://doi.org/10.5281/zenodo.13757440} {Transcriboquest 2024 medieval literary}.

\bibitem[{Vidal-Gorène et~al.(2021)Vidal-Gorène, Lucas, Salah, Decours-Perez, and Dupin}]{rasam1}
Chahan Vidal-Gorène, Noëmie Lucas, Clément Salah, Aliénor Decours-Perez, and Boris Dupin. 2021.
\newblock Rasam -- a dataset for the recognition and analysis of scripts in arabic maghrebi.
\newblock In \emph{Document Analysis and Recognition -- ICDAR 2021 Workshops}, pages 265--281, Cham. Springer International Publishing.

\bibitem[{Vlachou-Efstathiou(2024)}]{vlachouefstathiou2024eutyches}
Malamatenia Vlachou-Efstathiou. 2024.
\newblock \href {https://bit.ly/cffinit} {Eutyches "de uerbo" glossed}.
\newblock If you use this software, please cite it using the metadata from this file.

\bibitem[{voor~de Nederlandse Letteren~(DBNL)(2019)}]{dbnl}
De~Digitale~Bibliotheek voor~de Nederlandse Letteren~(DBNL). 2019.
\newblock \href {https://doi.org/https://zenodo.org/records/3239290#.XPfIqhYza70} {Dbnl ocr data set}.

\bibitem[{Wang et~al.(2024{\natexlab{a}})Wang, Bai, Tan, Wang, Fan, Bai, Chen, Liu, Wang, Ge et~al.}]{wang2024qwen2}
Peng Wang, Shuai Bai, Sinan Tan, Shijie Wang, Zhihao Fan, Jinze Bai, Keqin Chen, Xuejing Liu, Jialin Wang, Wenbin Ge, and 1 others. 2024{\natexlab{a}}.
\newblock \href {https://arxiv.org/abs/2409.12191} {Qwen2-vl: Enhancing vision-language model's perception of the world at any resolution}.
\newblock \emph{ArXiv preprint}, abs/2409.12191.

\bibitem[{Wang et~al.(2025)Wang, Gao, Gu, Pu, Cui, Wei, Liu, Jing, Ye, Shao, Wang, Chen, Zhang, Yang, Wang, Wei, Yin, Li, Cui, Chen, Ding, Tian, Wu, Xie, Li, Yang, Duan, Wang, Hou, Hao, Zhang, Li, Zhao, Duan, Deng, Fu, He, Wang, He, Shi, He, Xiong, Lv, Wu, Shao, Zhang, Deng, Qi, Ge, Guo, Zhang, Zhang, Cao, Lin, Tang, Gao, Huang, Gu, Lyu, Tang, Wang, Lv, Ouyang, Wang, Dou, Zhu, Lu, Lin, Dai, Su, Zhou, Chen, Qiao, Wang, and Luo}]{wang2025internvl35advancingopensourcemultimodal}
Weiyun Wang, Zhangwei Gao, Lixin Gu, Hengjun Pu, Long Cui, Xingguang Wei, Zhaoyang Liu, Linglin Jing, Shenglong Ye, Jie Shao, Zhaokai Wang, Zhe Chen, Hongjie Zhang, Ganlin Yang, Haomin Wang, Qi~Wei, Jinhui Yin, Wenhao Li, Erfei Cui, and 56 others. 2025.
\newblock \href {https://arxiv.org/abs/2508.18265} {Internvl3.5: Advancing open-source multimodal models in versatility, reasoning, and efficiency}.

\bibitem[{Wang et~al.(2024{\natexlab{b}})Wang, Xia, He, Chen, Liu, Zhu, Liang, Wu, Liu, Malladi, Chevalier, Arora, and Chen}]{wang2024charxiv}
Zirui Wang, Mengzhou Xia, Luxi He, Howard Chen, Yitao Liu, Richard Zhu, Kaiqu Liang, Xindi Wu, Haotian Liu, Sadhika Malladi, Alexis Chevalier, Sanjeev Arora, and Danqi Chen. 2024{\natexlab{b}}.
\newblock \href {http://papers.nips.cc/paper\_files/paper/2024/hash/cdf6f8e9fd9aeaf79b6024caec24f15b-Abstract-Datasets\_and\_Benchmarks\_Track.html} {Charxiv: Charting gaps in realistic chart understanding in multimodal llms}.
\newblock In \emph{Advances in Neural Information Processing Systems 38: Annual Conference on Neural Information Processing Systems 2024, NeurIPS 2024, Vancouver, BC, Canada, December 10 - 15, 2024}.

\bibitem[{Weber et~al.(2023)Weber, Siebenschuh, Butler, Alexandrov, Thanner, Tsolakis, Jabbar, Foster, Li, Stevens, and Zhang}]{weber2023wordscape}
Maurice Weber, Carlo Siebenschuh, Rory Butler, Anton Alexandrov, Valdemar Thanner, Georgios Tsolakis, Haris Jabbar, Ian~T. Foster, Bo~Li, Rick Stevens, and Ce~Zhang. 2023.
\newblock \href {http://papers.nips.cc/paper\_files/paper/2023/hash/52c1ce1a0eaf61e8b6e3a899c1b9c61f-Abstract-Datasets\_and\_Benchmarks.html} {Wordscape: a pipeline to extract multilingual, visually rich documents with layout annotations from web crawl data}.
\newblock In \emph{Advances in Neural Information Processing Systems 36: Annual Conference on Neural Information Processing Systems 2023, NeurIPS 2023, New Orleans, LA, USA, December 10 - 16, 2023}.

\bibitem[{Whetter(2017)}]{Whetter_2017}
K.~S. Whetter. 2017.
\newblock \emph{The Manuscript and Meaning of Malory’s Morte Darthur: Rubrication, Commemoration, Memorialization}.
\newblock Arthurian Studies. Boydell and Brewer.

\bibitem[{White et~al.(2022)White, Clérice, and Karaisl}]{whitecarolineminuscule}
Nick White, Thibault Clérice, and Antonia Karaisl. 2022.
\newblock \href {https://github.com/rescribe/carolineminuscule-groundtruth} {Caroline minuscule by rescribe}.
\newblock Project website: https://rescribe.xyz/.

\bibitem[{Wu et~al.(2024)Wu, Chen, Pan, Liu, Liu, Dai, Gao, Ma, Wu, Wang, Xie, Wu, Hu, Wang, Sun, Li, Piao, Guan, Liu, Xie, You, Dong, Yu, Zhang, Zhao, Wang, and Ruan}]{wu2024deepseekvl2}
Zhiyu Wu, Xiaokang Chen, Zizheng Pan, Xingchao Liu, Wen Liu, Damai Dai, Huazuo Gao, Yiyang Ma, Chengyue Wu, Bingxuan Wang, Zhenda Xie, Yu~Wu, Kai Hu, Jiawei Wang, Yaofeng Sun, Yukun Li, Yishi Piao, Kang Guan, Aixin Liu, and 8 others. 2024.
\newblock \href {https://arxiv.org/abs/2412.10302} {Deepseek-vl2: Mixture-of-experts vision-language models for advanced multimodal understanding}.

\bibitem[{Wyatt and Tymms(1861)}]{wyatt1861history}
M.D. Wyatt and W.R. Tymms. 1861.
\newblock \href {https://books.google.com/books?id=npgR0QEACAAJ} {\emph{The History, Theory, and Practice of Illuminating}}.
\newblock Day and son.

\bibitem[{Xiaomi(2025)}]{coreteam2025mimovltechnicalreport}
LLM-Core-Team Xiaomi. 2025.
\newblock \href {https://arxiv.org/abs/2506.03569} {Mimo-vl technical report}.

\bibitem[{Yang et~al.(2024{\natexlab{a}})Yang, Yang, Zhang, Hui, Zheng, Yu, Li, Liu, Huang, Wei et~al.}]{yang2024qwen2}
An~Yang, Baosong Yang, Beichen Zhang, Binyuan Hui, Bo~Zheng, Bowen Yu, Chengyuan Li, Dayiheng Liu, Fei Huang, Haoran Wei, and 1 others. 2024{\natexlab{a}}.
\newblock \href {https://arxiv.org/abs/2412.15115} {Qwen2.5 technical report}.
\newblock \emph{ArXiv preprint}, abs/2412.15115.

\bibitem[{Yang et~al.(2018)Yang, Jin, Huang, Yang, Lai, and Sun}]{tkh-and-mth}
Hailin Yang, Lianwen Jin, Weiguo Huang, Zhaoyang Yang, Songxuan Lai, and Jifeng Sun. 2018.
\newblock \href {https://doi.org/10.1109/ACCESS.2018.2840218} {Dense and tight detection of chinese characters in historical documents: Datasets and a recognition guided detector}.
\newblock \emph{IEEE Access}, 6:30174--30183.

\bibitem[{Yang et~al.(2025)Yang, Ni, Xiang, Hu, Peng, and Jiang}]{r4b}
Qi~Yang, Bolin Ni, Shiming Xiang, Han Hu, Houwen Peng, and Jie Jiang. 2025.
\newblock \href {https://arxiv.org/abs/2508.21113} {R-4b: Incentivizing general-purpose auto-thinking capability in mllms via bi-mode annealing and reinforce learning}.

\bibitem[{Yang et~al.(2024{\natexlab{b}})Yang, Tang, Li, Wang, Wan, Zhong, Liu, Yang, Wang, Liu et~al.}]{yang2024cc}
Zhibo Yang, Jun Tang, Zhaohai Li, Pengfei Wang, Jianqiang Wan, Humen Zhong, Xuejing Liu, Mingkun Yang, Peng Wang, Yuliang Liu, and 1 others. 2024{\natexlab{b}}.
\newblock \href {https://arxiv.org/abs/2412.02210} {Cc-ocr: A comprehensive and challenging ocr benchmark for evaluating large multimodal models in literacy}.
\newblock \emph{ArXiv preprint}, abs/2412.02210.

\bibitem[{Ying et~al.(2024)Ying, Meng, Wang, Li, Lin, Yang, Zhang, Zhang, Lin, Liu, Lei, Lu, Chen, Xu, Zhang, Zhang, Gao, Wang, Qiao, Luo, Zhang, and Shao}]{mmt-bench-2024}
Kaining Ying, Fanqing Meng, Jin Wang, Zhiqian Li, Han Lin, Yue Yang, Hao Zhang, Wenbo Zhang, Yuqi Lin, Shuo Liu, Jiayi Lei, Quanfeng Lu, Runjian Chen, Peng Xu, Renrui Zhang, Haozhe Zhang, Peng Gao, Yali Wang, Yu~Qiao, and 3 others. 2024.
\newblock \href {https://openreview.net/forum?id=R4Ng8zYaiz} {Mmt-bench: {A} comprehensive multimodal benchmark for evaluating large vision-language models towards multitask {AGI}}.
\newblock In \emph{Forty-first International Conference on Machine Learning, {ICML} 2024, Vienna, Austria, July 21-27, 2024}. OpenReview.net.

\bibitem[{Zerrouki(2010)}]{zerrouki2012pyarabic}
Taha Zerrouki. 2010.
\newblock \href {https://pypi.python.org/pypi/pyarabic} {pyarabic, an arabic language library for python}.

\bibitem[{Zhai et~al.(2023)Zhai, Mustafa, Kolesnikov, and Beyer}]{zhai2023sigmoid}
Xiaohua Zhai, Basil Mustafa, Alexander Kolesnikov, and Lucas Beyer. 2023.
\newblock \href {https://doi.org/10.1109/ICCV51070.2023.01100} {Sigmoid loss for language image pre-training}.
\newblock In \emph{{IEEE/CVF} International Conference on Computer Vision, {ICCV} 2023, Paris, France, October 1-6, 2023}, pages 11941--11952. {IEEE}.

\bibitem[{Zhang et~al.(2023)Zhang, Dong, Wang, Cao, Xu, Ouyang, Zhao, Duan, Zhang, Ding et~al.}]{zhang2023internlm}
Pan Zhang, Xiaoyi Dong, Bin Wang, Yuhang Cao, Chao Xu, Linke Ouyang, Zhiyuan Zhao, Haodong Duan, Songyang Zhang, Shuangrui Ding, and 1 others. 2023.
\newblock \href {https://arxiv.org/abs/2309.15112} {Internlm-xcomposer: A vision-language large model for advanced text-image comprehension and composition}.
\newblock \emph{ArXiv preprint}, abs/2309.15112.

\bibitem[{Zhong et~al.(2019)Zhong, Tang, and Yepes}]{zhong2019publaynet}
Xu~Zhong, Jianbin Tang, and Antonio~Jimeno Yepes. 2019.
\newblock Publaynet: largest dataset ever for document layout analysis.
\newblock In \emph{2019 International conference on document analysis and recognition (ICDAR)}, pages 1015--1022. IEEE.

\end{thebibliography}

\appendix
\section{\dataset Details}
\subsection{Dataset Statistics}
\label{appendix:dataset-stats}

\begin{figure}[ht!]
\centering
\includegraphics[width=\linewidth]{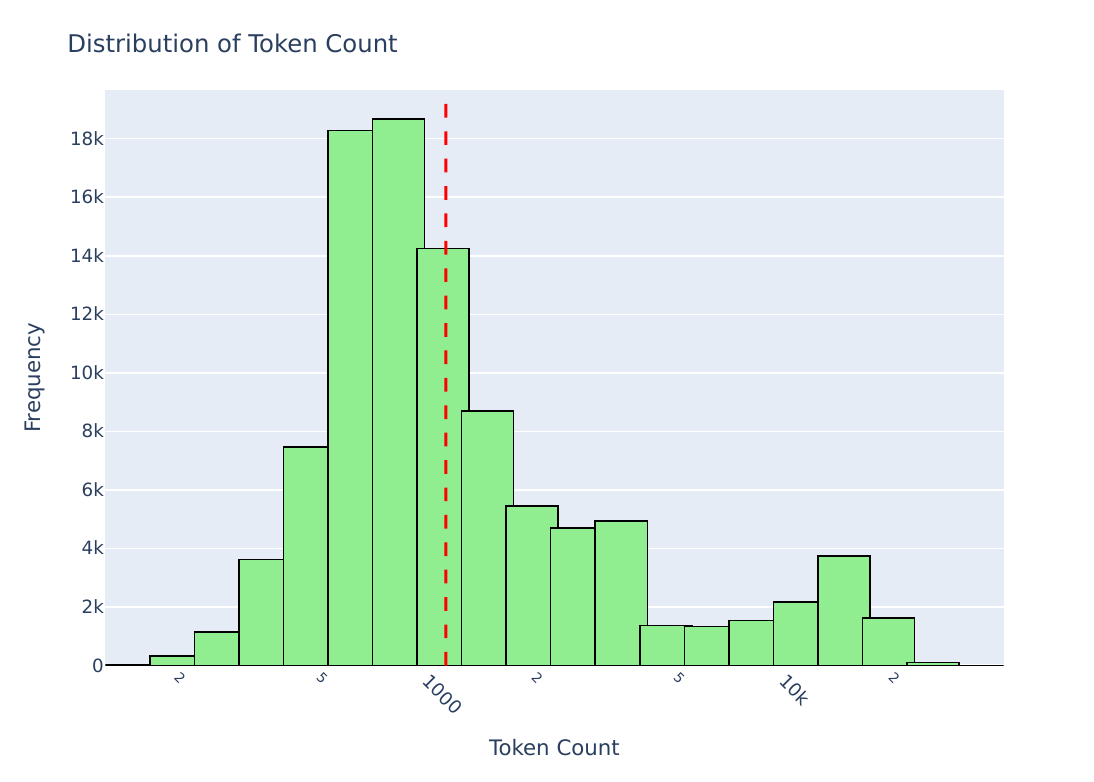}
\caption{Distribution of Qwen 2.5 VL tokens in the gold text. Log-scale x-axis. Red dashed line shows the median.}
\label{fig:dataset-token-count-distribution}
\end{figure}

\begin{figure}[ht!]
\centering
\includegraphics[width=\linewidth, trim={0.5cm 0cm 1cm 2cm}, clip]{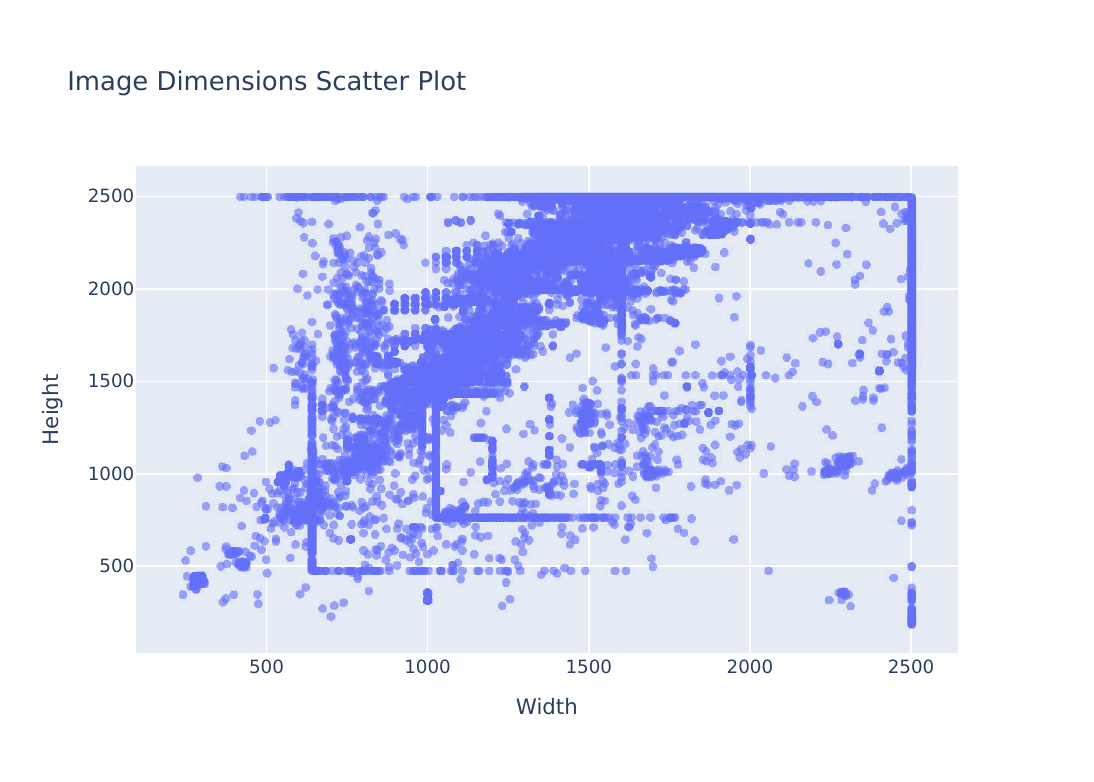}
\caption{Plot of image dimensions in \dataset. Each point shows the width and height of a single image. Note that we resize large images to fit inside a 2500 $\times$ 2500 pixel box while maintaining their original aspect ratio.}
\label{fig:scatter-plot-dimensions}
\end{figure}

Figures~\ref{fig:dataset-token-count-distribution} and \ref{fig:scatter-plot-dimensions} show the distribution of the length of the gold text labels, and the dimension of images in \dataset respectively.

\subsection{Language and Script Statistics}
Tables~\ref{tab:language_stats} and \ref{tab:dataset-script-statistics} show the number of examples per language and script, in \dataset.

\begin{table*}[ht]
\centering
\small
\begin{tabular}{|c|c|c|r|r|r|}
\hline

\textbf{Document Type} & \textbf{Cluster} & \textbf{Total} & \textbf{Train} & \textbf{Dev} & \textbf{Test} \\
\hline
\multirow{36}{*}{print} & German & 21,024 & 20,964 & 30 & 30 \\
& French & 14,648 & 14,588 & 30 & 30 \\
& Spanish & 8,426 & 8,366 & 30 & 30 \\
& Latin & 5,158 & 5,098 & 30 & 30 \\
& Czech & 3,145 & 3,085 & 30 & 30 \\
& Dutch & 3,016 & 2,956 & 30 & 30 \\
& Polish & 2,781 & 2,721 & 30 & 30 \\
& Slovenian & 2,723 & 2,663 & 30 & 30 \\
& English & 2,504 & 2,444 & 30 & 30 \\
& Japanese & 1,954 & 1,894 & 30 & 30 \\
& Bulgarian & 737 & 677 & 30 & 30 \\
& Sanskrit & 543 & 483 & 30 & 30 \\
& Finnish & 283 & 223 & 30 & 30 \\
& Swedish & 270 & 210 & 30 & 30 \\
& Romanian & 154 & 94 & 30 & 30 \\
& Hindi & 154 & 94 & 30 & 30 \\
& Chinese & 66 & 6 & 30 & 30 \\
& Bangla & 62 & 2 & 30 & 30 \\
& Italian & 50 & 50 & 0 & 0 \\
& Serbian & 48 & 48 & 0 & 0 \\
& Malayalam & 48 & 48 & 0 & 0 \\
& Estonian & 47 & 47 & 0 & 0 \\
& Slovak & 42 & 42 & 0 & 0 \\
& Latvian & 40 & 40 & 0 & 0 \\
& Indonesian & 38 & 38 & 0 & 0 \\
& Corsican & 35 & 35 & 0 & 0 \\
& Portuguese & 14 & 14 & 0 & 0 \\
& Catalan & 13 & 13 & 0 & 0 \\
& Russian & 9 & 9 & 0 & 0 \\
& Greek & 9 & 9 & 0 & 0 \\
& Occitan & 5 & 5 & 0 & 0 \\
& Yiddish & 3 & 3 & 0 & 0 \\
& Hungarian & 2 & 2 & 0 & 0 \\
& Malay & 1 & 1 & 0 & 0 \\
& Javanese & 1 & 1 & 0 & 0 \\
& Croatian & 1 & 1 & 0 & 0 \\
\hline

\multirow{28}{*}{handwriting} & Chinese & 5,113 & 5,053 & 30 & 30 \\
& Spanish & 4,512 & 4,452 & 30 & 30 \\
& French & 3,627 & 3,567 & 30 & 30 \\
& English & 3,439 & 3,379 & 30 & 30 \\
& Dutch & 3,262 & 3,202 & 30 & 30 \\
& Arabic & 2,367 & 2,307 & 30 & 30 \\
& German & 2,358 & 2,298 & 30 & 30 \\
& Latin & 1,596 & 1,536 & 30 & 30 \\
& Japanese & 892 & 832 & 30 & 30 \\
& Khmer & 622 & 562 & 30 & 30 \\
& Swedish & 552 & 492 & 30 & 30 \\
& Norwegian & 441 & 381 & 30 & 30 \\
& Italian & 430 & 370 & 30 & 30 \\
& Greek & 425 & 365 & 30 & 30 \\
& Persian & 417 & 357 & 30 & 30 \\
& Vietnamese & 416 & 356 & 30 & 30 \\
& Turkish & 237 & 177 & 30 & 30 \\
& Sanskrit & 182 & 122 & 30 & 30 \\
& Hebrew & 155 & 95 & 30 & 30 \\
& Catalan & 145 & 85 & 30 & 30 \\
& Portuguese & 72 & 12 & 30 & 30 \\
& Czech & 59 & 59 & 0 & 0 \\
& Finnish & 45 & 45 & 0 & 0 \\
& Newari & 19 & 19 & 0 & 0 \\
& Russian & 2 & 2 & 0 & 0 \\
& Pali & 2 & 2 & 0 & 0 \\
& Danish & 2 & 2 & 0 & 0 \\
& Yiddish & 1 & 1 & 0 & 0 \\

\hline
\end{tabular}
\caption{Language statistics in \dataset.}
\label{tab:language_stats}
\end{table*}

\begin{table*}[ht]
\centering
\resizebox{0.95\textwidth}{!}{
\begin{tabular}{|l|c|c|c|c|r|}
\hline
\textbf{Script Name} & \textbf{Script Family} & \textbf{Script Type} & \textbf{Writing Direction} & \textbf{ISO 15924 Code} & \textbf{Count in \dataset} \\
\hline

Latin & European & alphabet & LTR & Latn & 71,749 \\
Latin (Fraktur variant) & European & alphabet & LTR & Latf & 12,987 \\
Latin (Gaelic variant) & European & alphabet & LTR & Latg & 117 \\
Cyrillic & European & alphabet & LTR & Cyrl & 952 \\
Greek & European & alphabet & LTR & Grek & 434 \\
Hebrew & Middle Eastern & abjad & RTL (some bidirectional) & Hebr & 159 \\
Arabic & Middle Eastern & abjad & RTL (bidirectional) & Arab & 3,021 \\
Devanagari (Nagari) & Indic & abugida & LTR & Deva & 698 \\
Bengali (Bangla) & Indic & abugida & LTR & Beng & 64 \\
Malayalam & Indic & abugida & LTR & Mlym & 50 \\
Newa & Indic & abugida & LTR & Newa & 195 \\
Khmer & Southeast Asian & abugida & LTR & Khmr & 624 \\
Japanese (alias for Han + Hiragana + Katakana) & East Asian & logo-syllabary & vertical (RTL) and horizontal (LTR) & Jpan & 2,779 \\
Han (Hanzi, Kanji, Hanja) & East Asian & logo-syllabary & vertical (RTL) and horizontal (LTR) & Hani & 5,662 \\

\hline
\end{tabular}
}
\caption{Script statistics in \dataset. Script family, type and writing direction are from Glottolog 5.1~\cite{glottolog}.}
\label{tab:dataset-script-statistics}
\end{table*}

\section{System Details}

\subsection{Model IDs}
\label{appendix:model-ids}
Table~\ref{tab:model-ids} lists the exact model IDs used in our experiments. For closed models, the IDs follow the providers' APIs: Google Vertex AI and Azure OpenAI. For open-weight models, the IDs are taken from the HuggingFace hub. Throughout this paper, we refer to models by their total parameter counts to clarify the model sizes used. In cases where the official model ID differs (e.g., olmOCR), we use the total parameter count for consistency.

\begin{table*}[ht!]
\centering
\small
\begin{tabularx}{\textwidth}{l p{0.3\textwidth} ll}
\toprule
\textbf{Model Name} & \textbf{Model ID} & \textbf{Reasoning Budget} & \textbf{How Accessed} \\
\midrule
GPT-4o                      & \texttt{gpt-4o-2024-11-20}                 &                                   & \multirow{11}{=}{Azure OpenAI API} \\
GPT-4o Mini                 & \texttt{gpt-4o-mini-2024-07-18}            &                                   & \\
GPT-4.1                     & \texttt{gpt-4.1-2025-04-14}                &                                   & \\
GPT-4.1 Mini                & \texttt{gpt-4.1-mini-2025-04-14}           &                                   & \\
GPT-4.1 Nano                & \texttt{gpt-4.1-nano-2025-04-14}           &                                   & \\
O1                          & \texttt{o1-2024-12-17}                      & medium          & \\
O3                          & \texttt{o3-2025-04-16}                      & medium          & \\
O4 Mini                     & \texttt{o4-mini-2025-04-16}                & medium          & \\
GPT-5                       & \texttt{gpt-5-2025-08-07}                   & medium          & \\
GPT-5 Mini                  & \texttt{gpt-5-mini-2025-08-07}              & medium          & \\
GPT-5 Nano                  & \texttt{gpt-5-nano-2025-08-07}              & medium          & \\

\midrule
Claude Sonnet 3.7           & \texttt{claude-3-7-sonnet-20250219}        & 2048 tokens      & \multirow{5}{=}{Google Vertex AI API} \\
Claude Sonnet 4             & \texttt{claude-sonnet-4-20250514}        & 2048 tokens      & \\
Claude Opus 4.1             & \texttt{claude-opus-4-1-20250805}        & 2048 tokens      & \\
Gemini 2.5 Flash            & \texttt{gemini-2.5-flash}                   & 2048 tokens      & \\
Gemini 2.5 Pro              & \texttt{gemini-2.5-pro}                     & 2048 tokens      & \\

\midrule
Qwen 2.5 VL 3B              & \texttt{Qwen/Qwen2.5-VL-3B-Instruct}        &                                   & \multirow{12}{=}{Local GPU using vLLM v0.10.2} \\
Qwen 2.5 VL 72B             & \texttt{Qwen/Qwen2.5-VL-72B-Instruct}       &                                   & \\
Phi 4 Multimodal            & \texttt{microsoft/Phi-4-multimodal-instruct}&                                   & \\
InternVL 3.5 (30B)          & \texttt{OpenGVLab/InternVL3\_5-30B-A3B}     &                                   & \\
MiMo VL (7B)                & \texttt{XiaomiMiMo/MiMo-VL-7B-RL-2508}      &                                   & \\
Nanonets OCR (4B)           & \texttt{nanonets/Nanonets-OCR-s}            &                                   & \\
Nemotron Nano VL (8B)       & \texttt{nvidia/Llama-3.1-Nemotron-Nano-VL-8B-V1} &                             & \\
NuMarkdown (8B)             & \texttt{numind/NuMarkdown-8B-Thinking}      &                                   & \\
Gemma 3 (27B)               & \texttt{google/gemma-3-27b-it}              &                                   & \\
R (4B)                      & \texttt{YannQi/R-4B}                        &                                   & \\
Skywork R1v3 (38B)          & \texttt{Skywork/Skywork-R1V3-38B}           &                                   & \\
RolmOCR (8B)                & \texttt{reducto/RolmOCR}                    &                                   & \\
olmOCR (8B)                 & \texttt{allenai/olmOCR-7B-0825 }            &                                   & \\
\bottomrule
\end{tabularx}
\caption{Model names, IDs, reasoning budget, and access methods.}
\label{tab:model-ids}
\end{table*}

\subsection{Hyperparameters}
\label{appendix:hyperparameters}

For all VLMs, we generate outputs with a temperature of 0 (i.e., greedy decoding). All results are reported from a single run.

We set the maximum number of generated tokens to 20,000 for non-reasoning models and 40,000 for reasoning models to accommodate the additional reasoning tokens they generate. The value 20,000 is chosen based on Figure~\ref{fig:dataset-token-count-distribution} to allow for the generation of all gold outputs.
Reasoning parameters for reasoning models are included in Table~\ref{tab:model-ids}.

To create \model, we fine-tuned Qwen 2.5 VL (3B) for 5 epochs on 32 NVIDIA H100 GPUs. Using gradient accumulation and a total effective batch size of 128, fine-tuning took approximately 25 hours.
We used a learning rate of $5 \times 10^{-5}$ with a cosine schedule. Overall, the computation for this paper, including fine-tuning and inference for all open-weight models, totaled about 6,000 GPU-hours on NVIDIA H100 GPUs.

The tokenizer of Qwen 2.5 VL outputs a variable number of tokens depending on the resolution of the input image. We resized all input images to a maximum of 5,120 image patches, each of size $28 \times 28$ pixels, for all fine-tuning experiments. This limits the total number of image and text tokens per example to under 25,000.

\subsection{Prompt Used for Zero-Shot Evaluation of VLMs}
\label{appendix:zeroshot-prompt}

The prompt used to evaluate zero-shot VLMs is shown in Table~\ref{tab:system_prompt}.

\begin{table*}[ht!]
\centering
\begin{tabular}{@{}p{0.95\textwidth}@{}}
\begin{tcolorbox}[colback=white!95!gray, colframe=black!80!black,
  boxrule=0.5pt, arc=2mm, auto outer arc, boxsep=4pt,
  left=4pt, right=4pt, top=4pt, bottom=4pt]

You are an expert in diplomatic transcription of historical documents from various languages. Your task is to extract the full text from a given page. Only output the transcribed text between \texttt{<answer>} and \texttt{</answer>} tags.\\

Follow these instructions:\\

1. You will be provided with a scanned document page.\\

2. Perform transcription on the entirety of the page, converting all visible text into the following format. Include handwritten and print text, if any. Include tables, captions, headers, main text and all other visible text.\\

3. If you encounter any non-text elements, simply skip them without attempting to describe them.

4. Do not modernize or standardize the text. For example, if the transcription is using ``U+017F'' instead of ``s'' or ``U+0430'' instead of ``a'', keep it that way.\\

5. When you come across text in languages other than English, transcribe it as accurately as possible without translation.\\

6. Output the OCR result in the following format:\\

\texttt{<answer>}

extracted text here

\texttt{</answer>}\\

Remember, your goal is to accurately transcribe the text from the scanned page as much as possible. Process the entire page, even if it contains a large amount of text, and provide clear, well-formatted output. Pay attention to the appropriate reading order and layout of the text.

\end{tcolorbox}
\end{tabular}
\vspace{-0.6cm}
\caption{The prompt used for all zero-shot experiments. ``U+017F'' is the Unicode for long S, used in 18th and early 19th century English. ``U+0430'' is the Unicode for cyrillic small letter ``a''. We added these characters to the prompt after we noticed most VLMs try to normalize the text to modern English characters.}
\label{tab:system_prompt}
\end{table*}

\section{Additional Experiments}
\begin{figure*}[htbp]
\centering
\includegraphics[width=0.47\textwidth,trim=0 0 0 60,clip]{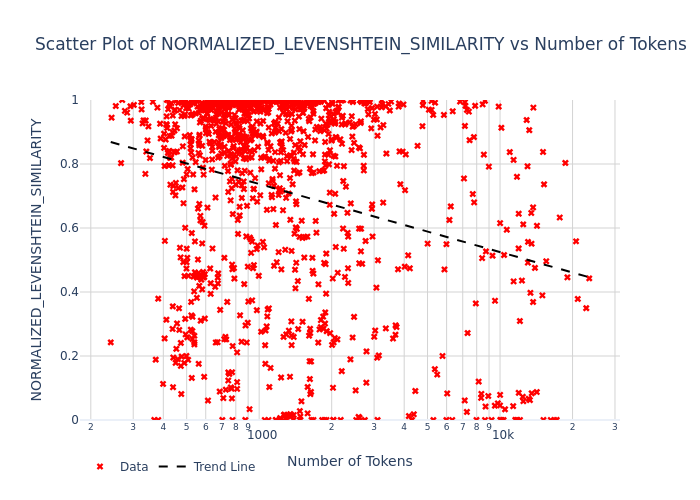}
\hfill
\includegraphics[width=0.47\textwidth,trim=0 0 0 60,clip]{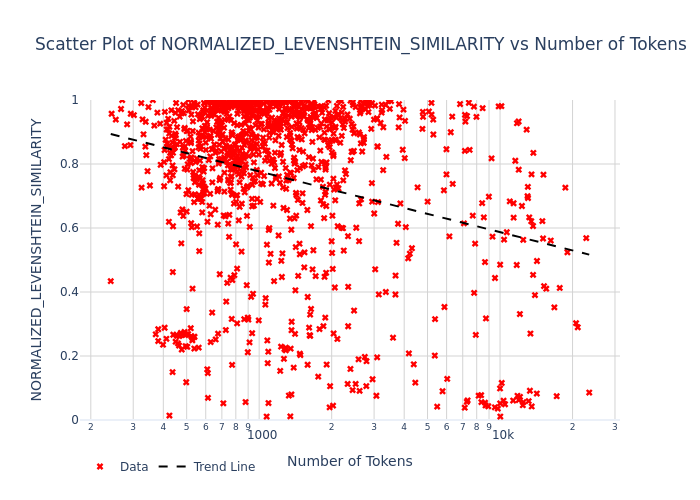}
\caption{The effect of document length on normalized Levenshtein similarity. Here we show this effect for Gemini 2.5 Pro on the left and \model on the right.}
\label{fig:length-vs-metric}
\end{figure*}

Figure~\ref{fig:length-vs-metric} plots normalized Levenshtein similarity against the number of tokens in the gold transcription for each page. We observe that, in general, model accuracy decreases as the amount of text on the page increases.

\section{Example System Outputs}
\label{appendix:example_output}
We present additional examples illustrating common error categories, with the correct gold output shown side by side with the outputs of the zero-shot Qwen 2.5 VL (3B) model and its fine-tuned counterpart, \model.

\subsection{Reading Order-Related Errors in Vertical Script Layouts}

\begin{figure*}[htbp]
\centering
\centering    \includegraphics[width=\linewidth]{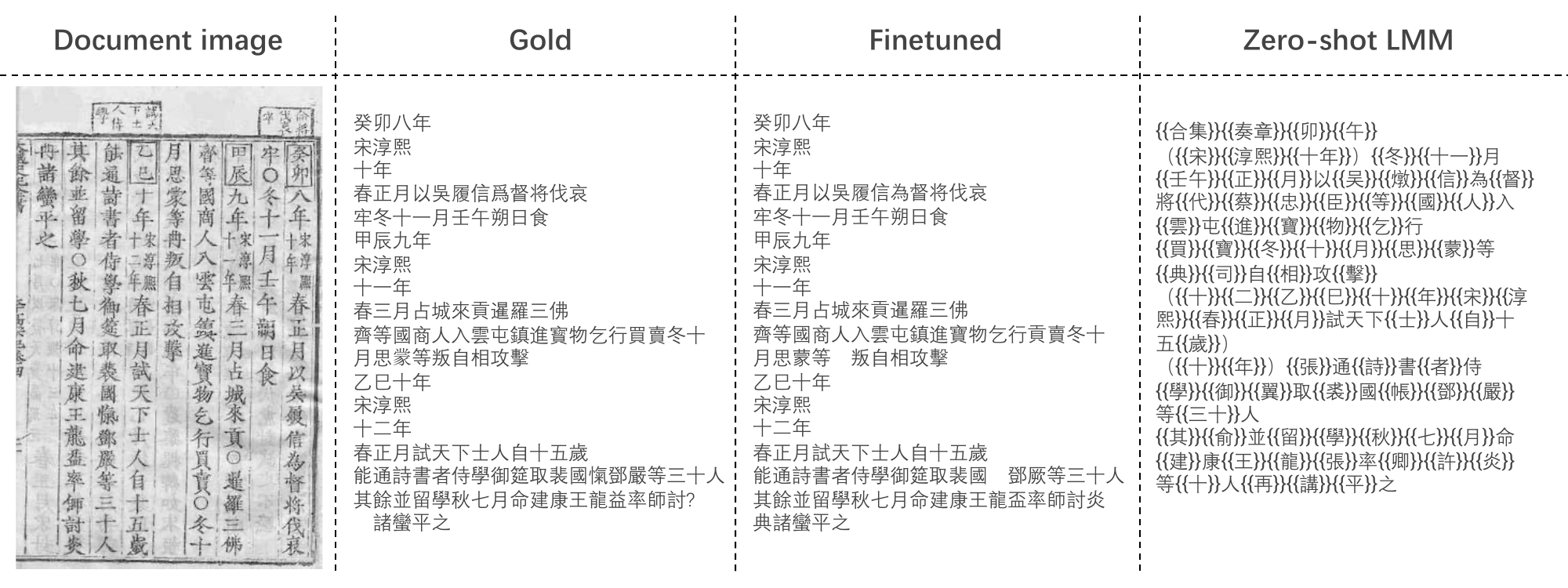}
\vspace{-1.2em}
\caption{Document example where the unusual reading order causes the zero-shot VLM to make major mistakes.\protect\footnotemark}
\label{fig:chinese-reading-order}
\end{figure*}

Figure~\ref{fig:chinese-reading-order} presents a Classical Chinese document that follows traditional layout conventions, where text is arranged vertically from top to bottom and columns are ordered from right to left.

The fine-tuned \model model produces a transcription that closely matches the ground truth. Most discrepancies are minor, such as normalization of historical glyph variants or the substitution of rare characters not fully represented in modern Unicode. Importantly, the reading order is preserved, and the model correctly segments lines and parses content without inserting hallucinated tokens.

In contrast, the zero-shot VLM fails to generate a coherent transcription. Its output contains hallucinated markup symbols, fragmented character groupings, and widespread omissions or rearrangements. We attribute this to a fundamental misalignment in layout assumptions: the model appears to interpret the document as a modern layout with left-to-right, top-to-bottom reading order, rather than adhering to the traditional vertical, right-to-left format. This results in substantial degradation in both recognition accuracy and semantic structure.

\footnotetext{The English translation of the gold text is: ``In the eighth year of the Guimao cycle, the 10th year of the Song dynasty’s Chunxi reign, in the first month of spring, Wu Lüxin was appointed as general to lead a campaign against Ailao. In the eleventh month, on the day of Renwu, a solar eclipse occurred. In the ninth year of Jiachen, the 11th year of Chunxi, in the third month of spring, Champa sent tribute, including three items from Siam. Merchants from Qi and neighboring countries came to Yuntun Garrison to offer goods and request permission to trade. In the tenth month, the Meng people and others rebelled and fought among themselves. In the 10th year of Yisi, the 12th year of Chunxi, in the first month of spring, civil service examinations were held for scholars aged fifteen and above. Those proficient in poetry and classical texts were admitted to study at the imperial court; thirty candidates, including Pei Guokai and Deng Yan, were selected. The rest remained in school. In the seventh month of autumn, the King of Jiankang, Long Yi, was ordered to lead an expedition against the southern tribes, which were subsequently pacified.''}

\subsection{Recognition Challenges with Small Characters}
\begin{figure*}[ht!]
\centering
\centering    \includegraphics[width=\linewidth]{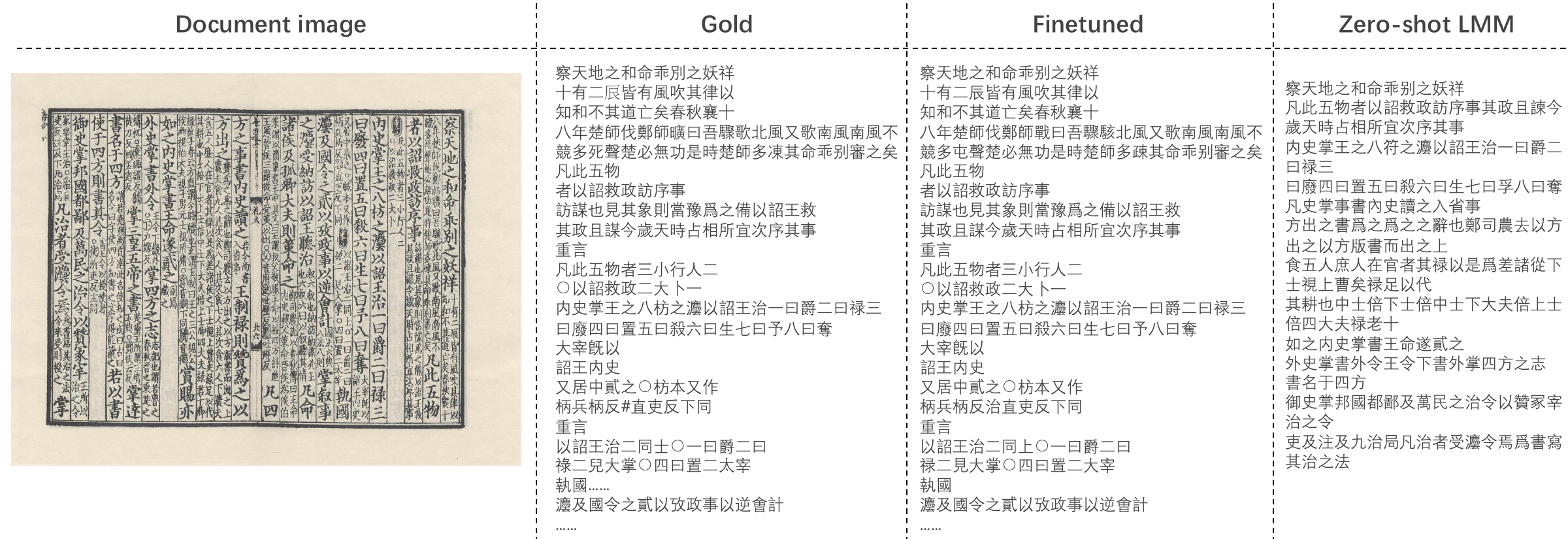}
\vspace{-1.2em}
\caption{A Classical Chinese document where the zero-shot VLM is not able to recognize the smaller text.
\protect\footnotemark
}
\label{fig:chinese-smaller-characters}
\end{figure*}

Figure~\ref{fig:chinese-smaller-characters} illustrates another challenging Classical Chinese document, characterized by a mix of large title-like characters and densely packed columns of smaller body text. Like the previous example, this page follows a vertical writing style with right-to-left column ordering.

\model transcribes the document with high accuracy. It preserves the reading order, successfully detects both large and small characters, and handles rare historical forms, with only minor mismatches due to character encoding limitations.

The zero-shot VLM, however, produces a significantly incomplete transcription. It primarily detects larger, visually prominent characters such as titles and headers, while failing to recognize most of the smaller text.

\footnotetext{The English translation of the gold text is: ``To understand the harmony of Heaven and Earth is to discern signs of disorder. Each of the twelve earthly branches has its corresponding wind, and when the winds deviate from their natural rhythm, it is a sign that order is failing. In the eighteenth year of Duke Xiang, the state of Chu attacked Zheng. The musician Shi Kuang first sang of the north wind, then of the south wind. The south wind lacked strength and coherence, and Shi Kuang declared that Chu would not succeed. Indeed, many Chu soldiers suffered from cold, confirming the signs of disrupted harmony. To address such disruptions and govern effectively, five roles are established to offer counsel and maintain order. These roles help interpret omens and seasonal patterns, ensuring that state affairs proceed in accordance with cosmic and political balance. Among them are three minor emissaries and two senior diviners, appointed to advise the king. The Inner Historian manages eight categories of royal decrees: granting rank, bestowing emoluments, dismissal, appointment, execution, birth, reward, and confiscation. The Grand Minister also conveys royal edicts and serves as a secondary authority, helping oversee the laws and administrative affairs of the state. These officials work together to uphold governance, interpret laws, and plan state policy.''}

\subsection{Reading Order in Western Languages}
\begin{figure*}[ht!]
\centering
\centering    \includegraphics[width=\linewidth]{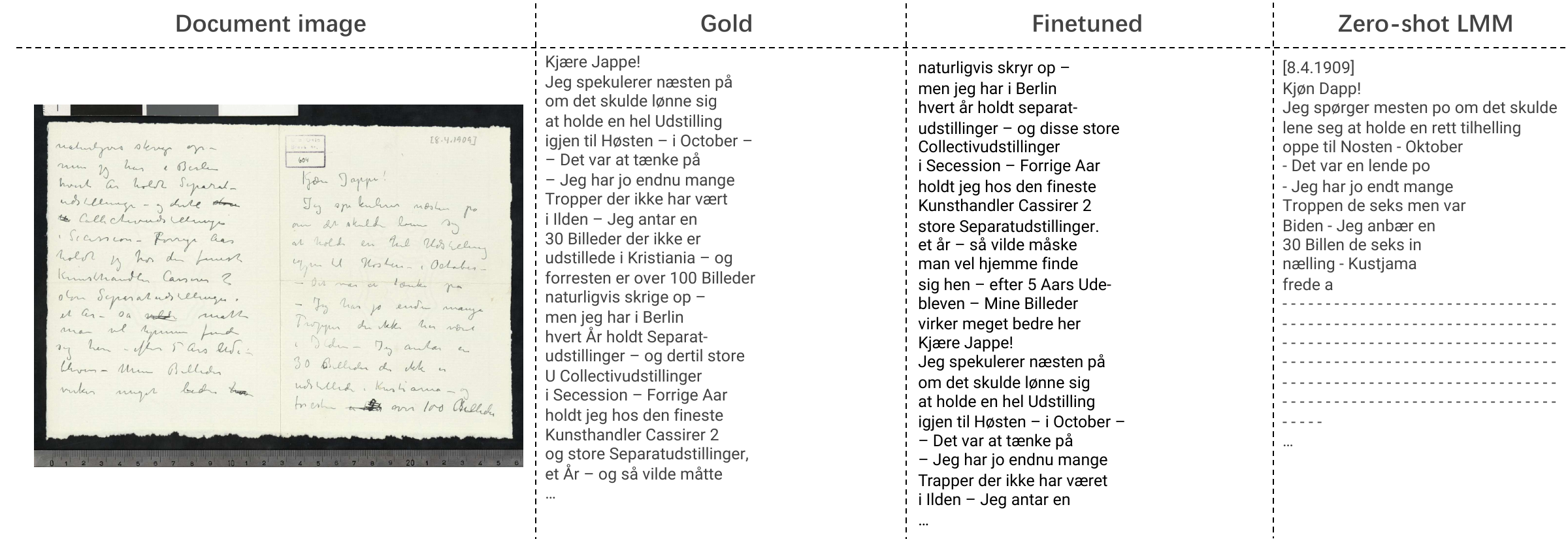}
\vspace{-1.2em}
\caption{A 20th-century Danish postcard that begins on the right side of the page, which causes mistakes even in \model predictions.}
\label{fig:danish}
\end{figure*}

The postcard in Figure~\ref{fig:danish} begins on the right side of the page, as is customary with postcards, and continues on the top left. \model recognizes the text mostly correctly, but it does not accurately identify the reading order. This leads to a prediction where the text order is reversed.

\subsection{Major Hallucinations from Stereotypes}
\begin{figure*}[ht!]
\centering
\centering    \includegraphics[width=\linewidth]{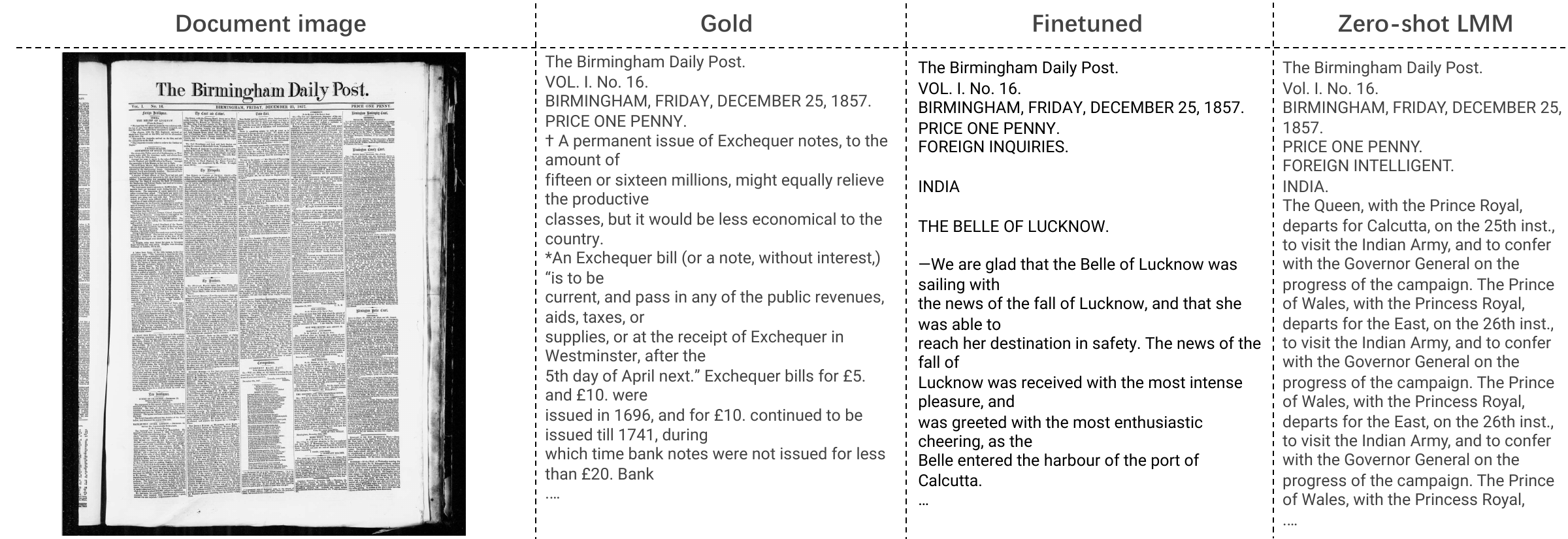}
\vspace{-1.2em}
\caption{19th-century English}
\label{fig:english-newspaper}
\end{figure*}

Some hallucinations due to misrecognition are based on stereotypes. In a mid-19th-century English newspaper (Figure~\ref{fig:english-newspaper}), the first column begins with a report from India. The second column mentions the Queen traveling with the Princess Royal. The VLM is unable to decipher this text but captures enough words---particularly India and queen---that it generates a stereotypical passage about the Queen traveling to Calcutta with the Prince Royal. While it is statistically likely that those traveling to India from the UK in the 19th century went to Calcutta in large numbers and royal princes traveled more than royal princesses, text generation based on stereotypes results in factual mistakes.

\subsection{Repetitive Degeneration}
\begin{figure*}[ht!]
\centering
\centering    \includegraphics[width=\linewidth]{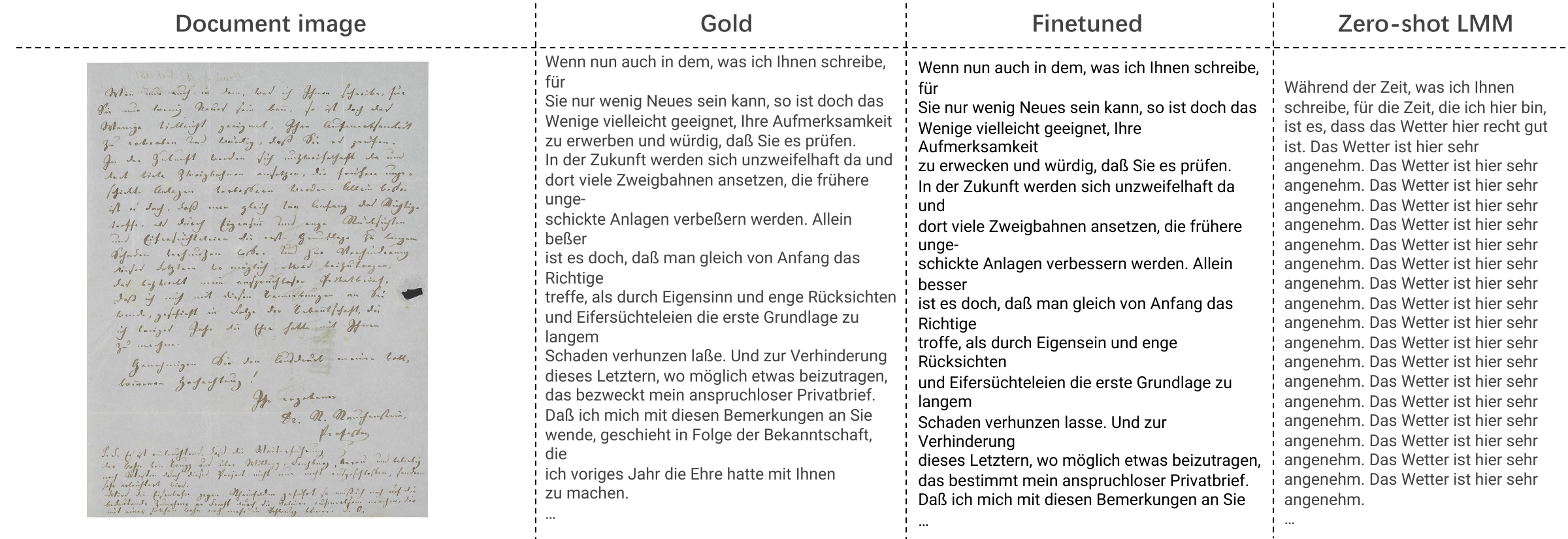}
\vspace{-1.2em}
\caption{A 19th-century German letter where the zero-shot VLM exhibits repetitive degeneration.}
\label{fig:german-letter}
\end{figure*}

Figure~\ref{fig:german-letter} shows an example from a 19th-century German letter. The zero-shot VLM, upon gathering from the first line that this is a letter, repeatedly predicts the same sentence over a dozen times: Das Wetter ist hier sehr angenehm,'' which translates to The weather here is really nice.'' This is a very standard opening line for a letter.

\begin{figure*}[ht!]
\centering
\centering    \includegraphics[width=\linewidth]{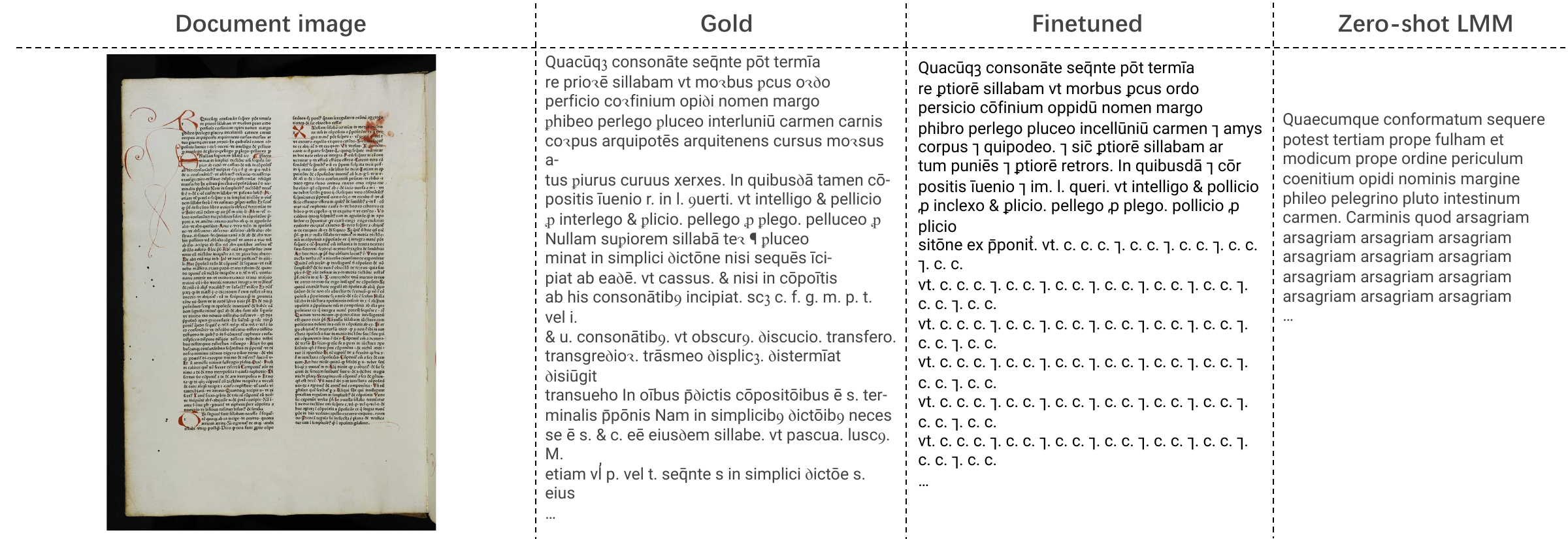}
\vspace{-1.2em}
\caption{A Latin manuscript where both models exhibit repetitive degeneration.}
\label{fig:latin-manuscript}
\end{figure*}

In the case of Figure~\ref{fig:latin-manuscript}, a Latin manuscript, the zero-shot VLM predicts the word ``arsagriam'' and, without another word to follow, repeatedly generates the same word until it runs out of generation tokens. This manuscript remains difficult even for \model. The fine-tuned VLM has several serious issues with recognizing the text and enters a degeneration loop with another phrase.

\subsection{Errors Due to Historical Script Changes}
\begin{figure*}[ht!]
\centering
\centering    \includegraphics[width=\linewidth]{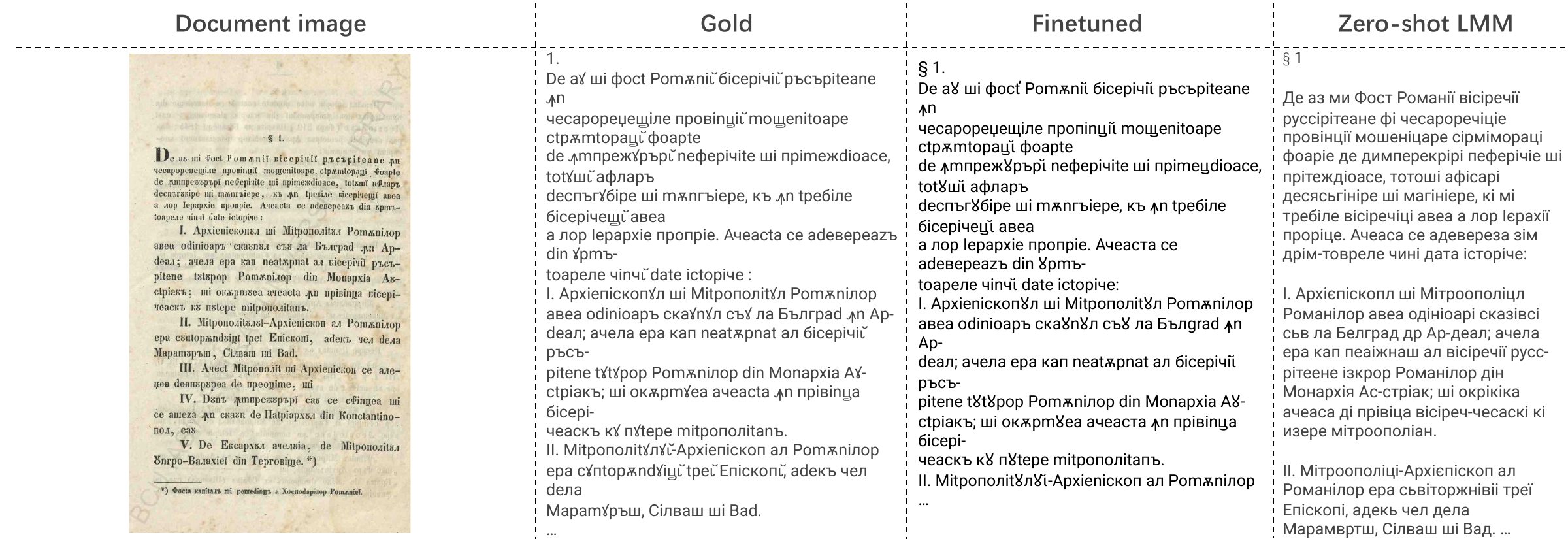}
\vspace{-1.2em}
\caption{A 19th-century transitional Romanian book where the zero-shot VLM is unable to recognize Latin characters.}
\label{fig:transitional-romanian}
\end{figure*}

Sometimes, in historical languages, it is not only the spelling that differs from contemporary usage but also the script itself. This is the case for many contemporary languages---for example, Azeri---which has historically been written in the Arabo-Persian script and continues to be written with that script in Iran. However, during the Soviet era in Azerbaijan, Azeri was written with the Cyrillic alphabet, and today it is written in the Latin alphabet. In our dataset, this diversity of scripts is showcased by transitional Romanian, a 19th-century alphabet that combined Cyrillic and Latin characters. Already a subject of computational linguistics research~\citep{frincu-etal-2023-challenges}, transitional Romanian poses challenges for VLMs. Figure~\ref{fig:transitional-romanian} shows a 19th-century printed Romanian book. Without fine-tuning, the VLM does not recognize the Latin characters properly and defaults to Cyrillic spellings. These mistakes are drastically reduced in \model.

\subsection{Strict Evaluation}
In Figure~\ref{fig:german-letter}, words like beßer'' or ausgeschloßen'' (the character representing the German sharp S) are transcribed as besser'' and ausgeschlossen'' (with two modern s characters), following orthographic changes in current standard German spelling. Such normalizations, while diverging from the text on paper, are not the same as actual recognition mistakes because they do not change the meaning of the text.

\begin{figure*}[ht!]
\centering
\centering    \includegraphics[width=\linewidth]{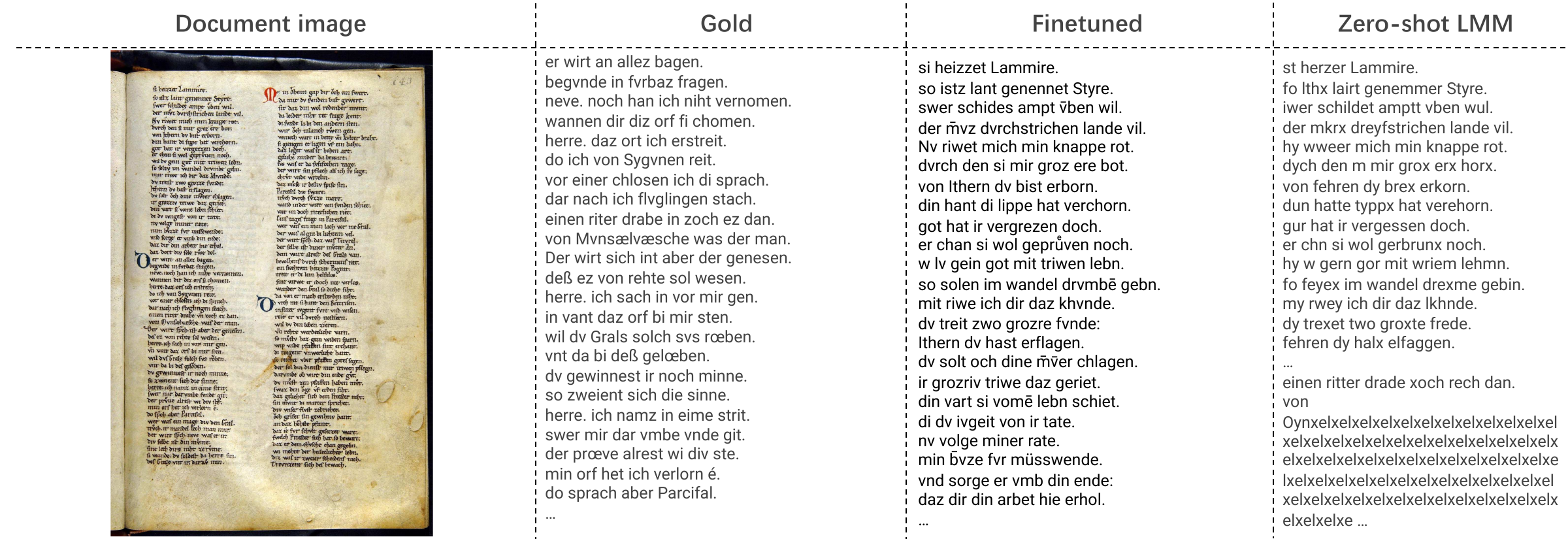}
\vspace{-1.2em}
\caption{Middle High German}
\label{fig:middle-high-german}
\end{figure*}

\begin{figure*}[ht!]
\centering
\centering    \includegraphics[width=\linewidth]{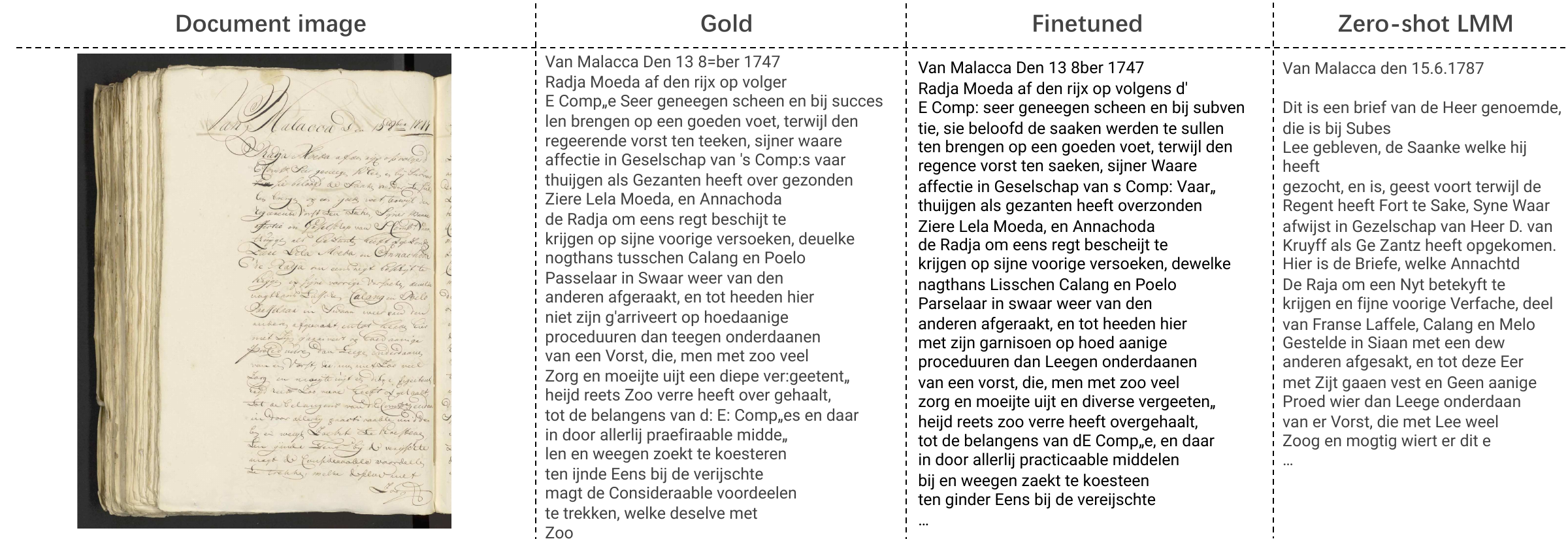}
\vspace{-1.2em}
\caption{18th-century Dutch letter, bound}
\label{fig:dutch-letter-bound}
\end{figure*}

\begin{figure*}[ht!]
\centering
\centering    \includegraphics[width=\linewidth]{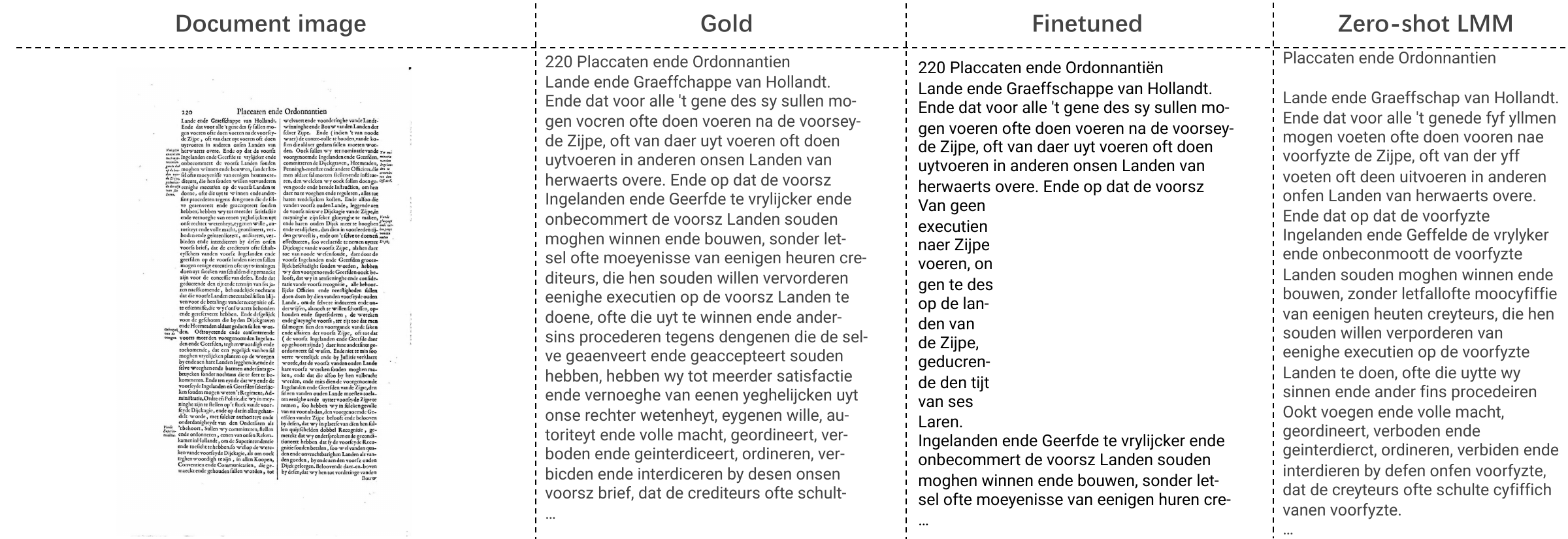}
\vspace{-1.2em}
\caption{17th-century Dutch print}
\label{fig:dutch-print}
\end{figure*}

\begin{figure*}[ht!]
\centering
\centering    \includegraphics[width=\linewidth]{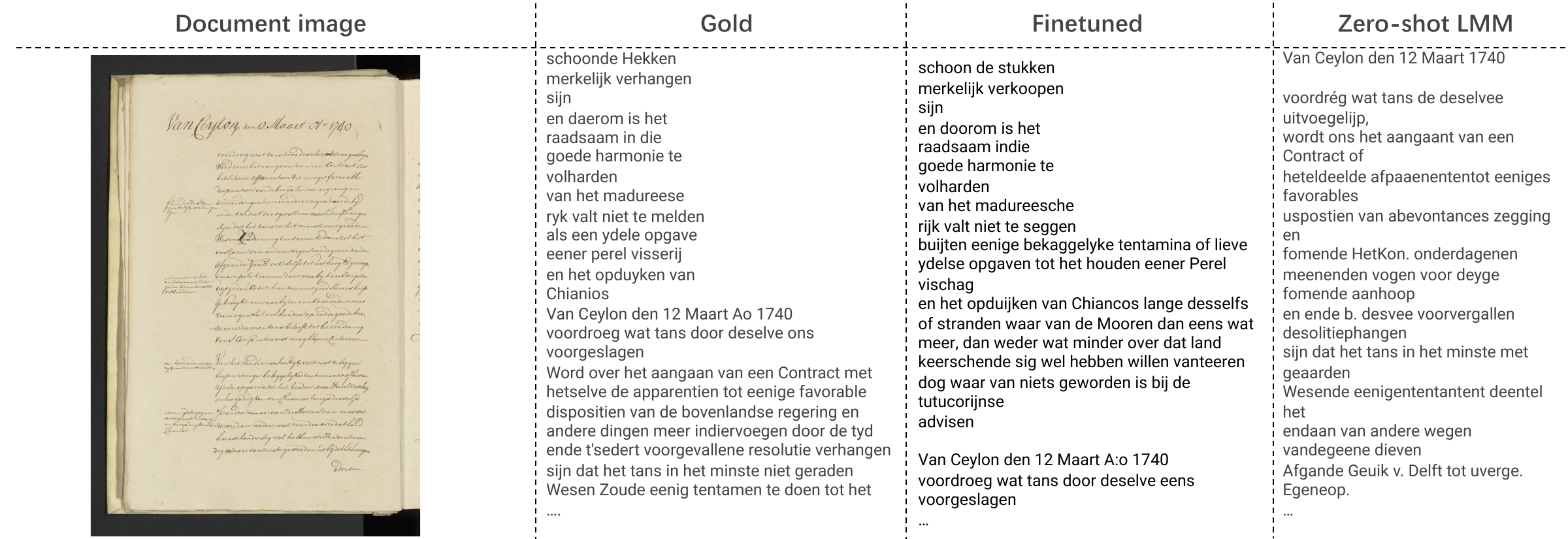}
\vspace{-1.2em}
\caption{18th-century Dutch letter with marginalia}
\label{fig:dutch-print-marginalia}
\end{figure*}

\begin{figure*}[ht!]
\centering
\centering    \includegraphics[width=\linewidth]{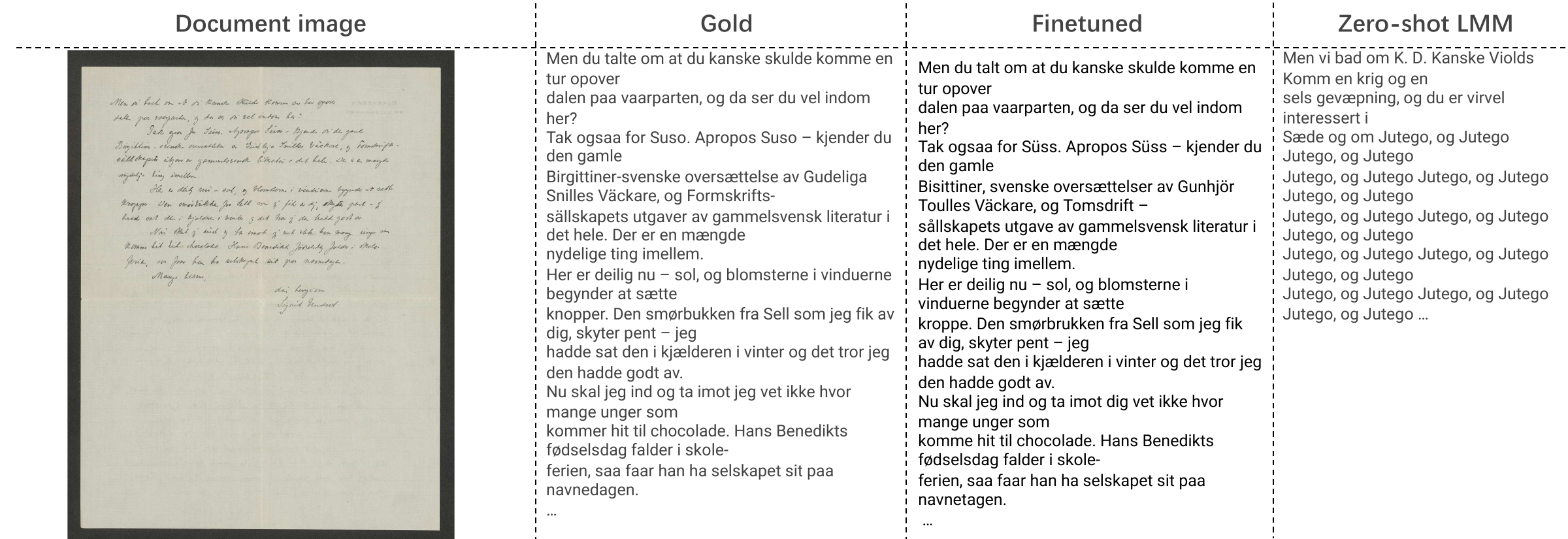}
\vspace{-1.2em}
\caption{19th–20th-century Norwegian letter}
\label{fig:norwegian-letter}
\end{figure*}

\section{List of 155 Sources Used to Create \dataset}
\label{appendix:dataset-list}

Here we list the datasets unified in order to build \dataset, alongside a short description of each. The quotations in the ``Description'' field are direct excerpts taken from the corresponding paper or dataset description, as cited.

\vspace{1em}
\textbf{DBNL OCR Data set
 \cite{dbnl}}\\
\textbf{Description:} ``A set of 220 books digitised by the Dutch DBNL
The texts range from 1776 to 1878 and are primarily in Dutch.''\\
\textbf{Link:} \url{https://zenodo.org/records/3239290#.XPfIqhYza70} \url{https://www.dbnl.org/titels/titels_pd.php}\\
\textbf{License:} Creative Commons Zero v1.0 Universal

\vspace{1em}
\textbf{Paris Bible Project \cite{gueville2021groundtruth}}\\
\textbf{Description:}  ``Latin Bibles throughout 13th and 14th century Europe. In the interest of diversifying the data we have opted for a wide variety of manuscript folios.''\\
\textbf{Link:} \url{https://github.com/parisbible/ground_truth/tree/main/PBP%201.0}\\
\textbf{License:} ``public domain''

\vspace{1em}
\textbf{POPP Datasets : Datasets for handwriting recognition from French population census \cite{constum2022popp}}\\
\textbf{Description:} \\
\textbf{Link:} \url{https://zenodo.org/record/6581158}\\
\textbf{License:} Creative Commons Attribution 4.0 International

\vspace{1em}
\textbf{Deutschen Textarchiv (DTA) \cite{dta2025}}\\
\textbf{Description:} \\
\textbf{Link:} \url{https://www.deutschestextarchiv.de/download}\\
\textbf{License:} CC BY-SA 4.0

\vspace{1em}
\textbf{Early Chinese Periodicals Online (ECPO) \cite{arnold2022ecpo}}\\
\textbf{Description:} \\
\textbf{Link:} \url{https://github.com/hcts-hra/ecpo-data} \url{https://heidata.uni-heidelberg.de/dataset.xhtml?persistentId=doi:10.11588/DATA/Z3J0DV}\\
\textbf{License:} CC BY 4.0

\vspace{1em}
\textbf{Atlas lingüístico diacrónico e interactivo de la Comunidad de Madrid \cite{aldicam}}\\
\textbf{Description:} \\
\textbf{Link:} \url{https://aldicam.corpuscodea.es/consultas.php}\\
\textbf{License:} permission granted for research use

\vspace{1em}
\textbf{Corpus of Spanish Documents Prior to 1700 (CODEA)~\cite{codea}}\\
\textbf{Description:} ``1,500 documents from various archives, covering all non-bilingual provinces of peninsular Spain and spanning from the 12th to the 17th centuries.'' \\
\textbf{Link:} \url{https://corpuscodea.es/corpus/corpus2022/consultas.php}\\
\textbf{License:} CC BY-NC-ND 4.0

\vspace{1em}
\textbf{VML-HD: The Historical Arabic Documents Dataset for Recognition Systems
 \cite{kassis2017vmlhd}}\\
\textbf{Description:} ``Based on five books written by different writers from the years 1088-1451.''\\
\textbf{Link:} \url{https://github.com/majeek/vml-hd?tab=readme-ov-file}\\
\textbf{License:} MIT License 

\vspace{1em}
\textbf{A Dataset of French Trade Directories from the 19th Century (FTD)
 \cite{abadie_dataset_22}}\\
\textbf{Description:} ``Composed of pages and entries extracted from French directories published between 1798 and 1861.''\\
\textbf{Link:} \url{https://zenodo.org/records/6394464}\\
\textbf{License:} The images were extracted from the original source https://gallica.bnf.fr, owned by the Bibliothèque nationale de France (French national library).
Original contents from the Bibliothèque nationale de France can be reused non-commercially, provided the mention ``Source gallica.bnf.fr / Bibliothèque nationale de France'' is kept.  
Researchers do not have to pay any fee for reusing the original contents in research publications or academic works.  
Original copyright mentions extracted from https://gallica.bnf.fr/edit/und/conditions-dutilisation-des-contenus-de-gallica on March 29, 2022.

The original contents were significantly transformed before being included in this dataset.
All derived content is licensed under the permissive Creative Commons Attribution 4.0 International license.

\vspace{1em}
\textbf{Information Extraction in Historical Handwritten Records (IEHHR 2017)
BH2M: the Barcelona Historical Handwritten Marriages database \cite{fernandezmota2014bh2m}}\\
\textbf{Description:} \\
\textbf{Link:} \url{https://rrc.cvc.uab.es/?ch=10&com=introduction} \url{http://dag.cvc.uab.es/the-historical-marriages-database/}\\
\textbf{License:} CC BY-NC-ND 4.0

\vspace{1em}
\textbf{Leopardi Dataset
 \cite{cascianelli2021learning}}\\
\textbf{Description:}  ``Collection of early 19th Century letters written in Italian by Giacomo Leopardi''\\
\textbf{Link:} \url{https://github.com/aimagelab/HWD/releases/tag/leopardi}\\
\textbf{License:} CC BY 4.0

\vspace{1em}
\textbf{Modern Magazine OCR Training Dataset (Created by CODH and NINJAL) \cite{codhmodernmagazineocr}}\\
\textbf{Description:} ``Data from the modern magazines Kokumin no Tomo (Issues 1--27) and Meiroku Zasshi (Issues 1--43)''\\
\textbf{Link:} \url{https://codh.rois.ac.jp/modern-magazine/dataset/}\\
\textbf{License:} CC BY 4.0

\vspace{1em}
\textbf{Japanese Classical Books \cite{codhjapaneseclassics}}\\
\textbf{Description:} ``In addition to the field of Japanese literature, this collection includes classical books from various fields such as medicine, science, and industry held by the National Institute of Japanese Literature. It also includes cookbooks and other materials held by the Ajinomoto Foundation for Dietary Culture, and features classical books photographed by the National Institute of Japanese Literature.''\\
\textbf{Link:} \url{https://codh.rois.ac.jp/pmjt/}\\
\textbf{License:} CC BY-SA 4.0

\vspace{1em}
\textbf{HisClima Dataset
 \cite{prhlt2022hisclima}}\\
\textbf{Description:} ``208 descriptive text page images of the Jeannette logbook and with 72 table images coming from different log books of a ship called Albatross.''\\
\textbf{Link:} \url{https://zenodo.org/records/7442971}\\
\textbf{License:} CC BY 4.0

\vspace{1em}
\textbf{StABS Ratsbücher O10, Urfehdenbuch X \cite{hodel2021stabs}}\\
\textbf{Description:} \\
\textbf{Link:} \url{https://zenodo.org/records/5153263}\\
\textbf{License:} CC-BY-NC-SA

\vspace{1em}
\textbf{The git-project-Boccace by the Boccace team \cite{vlachouefstathiou2022boccace}}\\
\textbf{Description:} ``it is composed of two documents: 1. A Middle French incunabulum, namely BnF Rés. J-845 (département Réserve des livres rares, Rés. J-845) written between 1498-1499,scanned in high definition in black and white (notice http://ark.bnf.fr/ark:/12148/cb30116914c). 2. The editio princeps of the work, namely Bibliothèque Mazarine Inc. 59 (notice https://www.sudoc.fr/135345618) scanned in high definition in color.''\\
\textbf{Link:} \url{https://github.com/PSL-Chartes-HTR-Students/HN2021-Boccace}\\
\textbf{License:} MIT License

\vspace{1em}
\textbf{19th-Century Romanian Transitional Script \cite{romanian_transitional}}\\
\textbf{Description:} ``Romanian texts written in the Romanian Transitional Script (RTS). RTS is a mix of Latin and Cyrillic characters that were used in the 19th century in the Romanian provinces to facilitate the transition from the Romanian Cyrillic Script to the modern Latin Script. The images cover the period between 1833 and 1864. The selected texts cover a diverse range of literary genres, including poems, novels, dramas, stories, newspapers, and religious texts.''\\
\textbf{Link:} \url{https://www.kaggle.com/datasets/mariuspenteliuc/rts-ocr}\\
\textbf{License:} ``You may use the dataset freely as long as you mention this page or the project below.''

\vspace{1em}
\textbf{NomNaOCR \cite{nomnaocr}}\\
\textbf{Description:} st\\
\textbf{Link:} \url{https://www.kaggle.com/datasets/quandang/nomnaocr}\\
\textbf{License:} CC BY-NC-SA 4.0

\vspace{1em}
\textbf{University of Denver Jewish Consumptives Relief Society Medical Records \cite{jcsr}}\\
\textbf{Description:} ``All individuals for whom records are provided have been deceased for at least 70 years, but were they still living today, these records would be recognized as being protected health information under the US Health Insurance Portability and Accountability Act of 1996 (HIPAA).''\\
\textbf{Link:} \url{https://zenodo.org/records/4243023}\\
\textbf{License:} CC BY 4.0

\vspace{1em}
\textbf{Vienna, Österreichische Nationalbibliothek - HTR Winter School 2022 \cite{wien}}\\
\textbf{Description:} ``This is ground truth for Vienna, Österreichische Nationalbibliothek, 2160, which is dated to the 3rd quarter of the 9thcentury (North Italy), probably in the vicinity of Rome (Kaiser, Liebs).''\\
\textbf{Link:} \url{https://zenodo.org/records/7537204}\\
\textbf{License:} CC BY 4.0

\vspace{1em}
\textbf{6000 ground truth of VOC and notarial deeds 3.000.000 HTR of VOC, WIC and notarial deeds \cite{voc}}\\
\textbf{Description:} ``17th and 18th century documents from the Dutch East-India Company (VOC). And 19th century notarial deeds from Noord-Hollands Archief and other archives in the provinces.''\\
\textbf{Link:} \url{https://zenodo.org/records/11209325}\\
\textbf{License:} CC BY 4.0

\vspace{1em}
\textbf{Incunabula Reichenau \cite{reichenau}}\\
\textbf{Description:} ``Gothic and Antiqua typefaces found in Latin incunabula and early prints. based on a collection of incunabula and post-incunabula of the former Reichenau monastery, now held at the Badische Landesbibliothek in Karlsruhe.''\\
\textbf{Link:} \url{https://zenodo.org/records/11046062}\\
\textbf{License:} CC-BY-SA 4.0

\vspace{1em}
\textbf{Towards a general open dataset and model for late medieval Castilian text recognition (HTR/OCR) \cite{castilian}}\\
\textbf{Description:} ``All manuscripts are known to have been produced in the 15th century. The incunabula was printed
in 1494. The corpus is a good basis for the production of a general corpus for the late medieval
period (14th-15th or even 13th-15th centuries), given its extension and the diversity of hands rep-
resented.''\\
\textbf{Link:} \url{https://zenodo.org/records/8340483}\\
\textbf{License:} CC BY-NC-SA 4.0

\vspace{1em}
\textbf{Ground truth of a sample Rudolph Gwalther's handwriting from ``Lateinische Gedichte'' \cite{gwalther}}\\
\textbf{Description:} ``Ruolph Gwalther's (1519-1586) handwriting taken from his book Lateinische Gedichte, where he accumulated writings between 1540 and 1580.''\\
\textbf{Link:} \url{https://zenodo.org/records/4780947}\\
\textbf{License:} CC BY-NC 4.0

\vspace{1em}
\textbf{ARletta: open-source handwritten text recognition models for historic Dutch \cite{arletta}}\\
\textbf{Description:} ``Manu: At the end of 2023, no models were publicly available in the kraken model repository (Kiessling, 2020) for the automated transcription of handwritten sources in historic Dutch, except the multilingual medieval CATMuS model that was partially trained on Middle Dutch (Clérice et al., 2023). Of all the models on HTR-united, Manu McFrench came closest to our Antwerp focus material because it was trained on French materials with a similar temporal scope (Chagué et al., 2023).

IJsberg: Two publicly available training sets created by the Dutch national archives seemed valuable to us because they share the same language, historical period, and administrative character (Keijser, 2020):
VOC: 4,735 pages of the 17th and 18th century East-Asia Company archive

Notarial: 1,615 pages of 19th-century Dutch notarial deeds
Antwerp incident books: We prepared a ground truth dataset from the Antwerp incident books. One expert (PhD student) annotated an initial subset of the data, consisting of 271 randomly selected pages.2 This subset was later supplemented with 3444 pages annotated by students.3 The student-contributed materials are much larger in scope than the materials annotated by the expert, but the former’s quality is unfortunately much less consistent.''\\
\textbf{Link:} \url{https://zenodo.org/records/11191457}\\
\textbf{License:} CC-BY-NC-SA

\vspace{1em}
\textbf{Diario del soldato Bruno Celestino \cite{celestino}}\\
\textbf{Description:} ``diary of Italian soldier Bruno Celestino in World War I''\\
\textbf{Link:} \url{https://zenodo.org/records/13760586}\\
\textbf{License:}  CC0 1.0

\vspace{1em}
\textbf{Dagboek Ernest Clarysse \cite{clarysse}}\\
\textbf{Description:} ``diary of Belgian citizen Ernest Clarysse in World War I''\\
\textbf{Link:} \url{https://zenodo.org/records/13769222}\\
\textbf{License:}  CC0 1.0

\vspace{1em}
\textbf{Mémoire sur St Domingue par H ? M. Michel \cite{domingue}}\\
\textbf{Description:} ``Transcription of monography Mémoire sur St Domingue par H ? M. Michel, dating from 1797 and dealing on slavery in Haiti''\\
\textbf{Link:} \url{https://zenodo.org/records/13784411}\\
\textbf{License:}  CC0 1.0

\vspace{1em}
\textbf{TKH and MTH Datasets
MTHv2 \cite{tkh-and-mth}}\\
\textbf{Description:} ``The Tripitaka Koreana in Han (TKH) Dataset and the Multiple Tripitaka in Han (MTH) Dataset for the research of Chinese character detection and recognition in historical documents''\\
\textbf{Link:} \url{https://github.com/HCIILAB/TKH_MTH_Datasets_Release} \url{https://github.com/HCIILAB/MTHv2_Datasets_Release}\\
\textbf{License:} ``The TKH dataset and MTH dataset can only be used for non-commercial research purpose.''

\vspace{1em}
\textbf{M5HisDoc \cite{m5hisdoc}}\\
\textbf{Description:} ``M5HisDoc consists of 5,000 historical Chinese handwritten document images collected from ancient books and manuscripts. The documents are typically structured with form-like layouts''\\
\textbf{Link:} \url{https://github.com/HCIILAB/M5HisDoc} \\
\textbf{License:} CC BY-NC-ND 4.0

\vspace{1em}
\textbf{The Pinkas Dataset \cite{pinkas}}\\
\textbf{Description:} ``Pinkas dataset is created from a historical Hebrew manuscript that contains records of Jewish communities in Europe in the early modern period (c. 1500-1800). The dataset consists of 30 pages digitized by full color digital images in JPG format with high resolution. The pages exhibit numerous degradations, complex layout and different handwritings. They are written in a mixture of Medieval Hebrew by different writers. The writers were not professional scribes. This adds additional challenge, since very often the same letters are written in different shapes.''\\
\textbf{Link:} \url{https://zenodo.org/records/3569694}\\
\textbf{License:} CC BY 4.0

\vspace{1em}
\textbf{Saint Gall database \cite{saintgall}}\\
\textbf{Description:} ``The Saint Gall database presented in [1] contains a handwritten historical manuscript with following characteristics:

9th century,
Latin language,
single writer,
Carolingian script,
ink on parchment''\\
\textbf{Link:} \url{https://fki.tic.heia-fr.ch/databases/saint-gall-database}\\
\textbf{License:} ``The Saint Gall database may be used for non-commercial research and teaching purposes only. If you are publishing scientific work based on the Saint Gall database, we request you to include a reference to our paper [1] A. Fischer, V. Frinken, A. Fornés, and H. Bunke: "Transcription Alignment of Latin Manuscripts using Hidden Markov Models," in Proc. 1st Int. Workshop on Historical Document Imaging and Processing, pages 29-36, 2011.

With kind permission of Prof. Ernst Tremp from the Abbey Library of Saint Gall, the original manuscript images [3] provided by e-codices can be used for non-commercial research and teaching purposes explicitly as follows:

Show and print sample manuscript images in scientific publications
Show sample manuscript images during talks
Show sample manuscript images online
For any purposes other than non-commercial research and teaching, the Abbey Library of Saint Gall has to be contacted first.

With kind permission of Max Bänziger from the Monumenta project, the aligned text edition [4] is also included in the Saint Gall database.''

\vspace{1em}
\textbf{Parzival Database
 \cite{parzival}}\\
\textbf{Description:} ``The Parzival database described in [4] contains a handwritten historical manuscript with following characteristics:

13th century
Medieval German language
three writers
Gothic script
ink on parchment''\\
\textbf{Link:} \url{https://fki.tic.heia-fr.ch/databases/parzival-database}\\
\textbf{License:} ``The Parzival database may be used for non-commercial research and teaching purposes only. If you are publishing scientific work based on the Parzival database, we request you to include a reference to [1] A. Fischer, A. Keller, V. Frinken, and H. Bunke: "Lexicon-Free Handwritten Word Spotting Using Character HMMs," in Pattern Recognition Letters, Volume 33(7), pages 934-942, 2012.

With kind permission of Prof. Ernst Tremp from the Abbey Library of Saint Gall, the original manuscript images can be used for non-commercial research and teaching purposes explicitly as follows:

Show and print sample manuscript images in scientific publications
Show sample manuscript images during talks
Show sample manuscript images online
For any purposes other than non-commercial research and teaching, the Abbey Library of Saint Gall has to be contacted first.''

\vspace{1em}
\textbf{The ``ScribbleLens'' Dutch Historical Handwriting Corpus \cite{scribble}}\\
\textbf{Description:} ``one of the first manuscript recognition corpus for Early
Modern Dutch, 400 years old material from the Dutch East
India company (VOC) with ship journals and other written
Materia'' ``80 writers'' ``The manuscripts span over 150 years of significant journeys by captains and traders from the Vereenigde Oost-indische Company (VOC) such as Tasman, Brouwer and Van Neck''\\
\textbf{Link:} \url{https://openslr.org/84/}\\
\textbf{License:} CC-BY-NC-ND

\vspace{1em}
\textbf{SLOVAK Supermodel print \& typewriter (SSPT1) \cite{slovensky}}\\
\textbf{Description:} ``92 s. GT of historical printed book J.A. Comenius' Orbis Pictus (1798 edition). the book has many illustrations, it is written in 4 languages (Latin, Hungarian, German, Czech), in addition in the form of tables and in antique and Swabian script

a whole series of historical newspapers, magazines and books from the 19th and early 20th centuries (Moravské noviny (1849), Program bulletins of the Slovak Philharmonic (1849-1970), Opavský Besedník (1863), Jitrenka (1840), I. Palugya: Kde jest pravda (1854), Lusatian Serbian magazine Lužica (1909), Šlabikár (1872), J.M. Hurban: Cirkev Ewanjelicko- Lutheránska (1861), J.N. Bobula: Jánošík (1862), D. Lichard: Obzor (1866) and others

typewritten documents, using samples of various fonts in Slovak, Czech, and German languages (ca. 150 pp.)

ca. 80 s. of GT transcriptions from various historical prints from the 18th and 19th centuries written in Czech (Svabian).''\\
\textbf{Link:} \url{https://zenodo.org/records/11218527}\\
\textbf{License:} CC BY 4.0

\vspace{1em}
\textbf{konzilsprotokolle \cite{konzils}}\\
\textbf{Description:} ``this dataset contains 8770 transcribed text lines of handwritten historical documents from the late 18th century.''\\
\textbf{Link:} \url{https://zenodo.org/records/215383}\\
\textbf{License:} CC BY 4.0

\vspace{1em}
\textbf{Charters and Records of Königsfelden Abbey and Bailiwick \cite{charters}}\\
\textbf{Description:} During its existence (1308-1528), the double monastery of Königsfelden was one of the central clerical institutions of Aargau and continued to exert a great attraction on the people of the region and beyond for centuries. The project makes the rich tradition from the Middle Ages and the early modern period accessible to researchers and interested laypeople in the form of a digital edition.\\
\textbf{Link:} \url{https://zenodo.org/records/5179361}\\
\textbf{License:} CC BY 4.0

\vspace{1em}
\textbf{NorHand v3 \cite{norhand}}\\
\textbf{Description:} ``Norwegian letter and diary documents from 19th and early 20th century''\\
\textbf{Link:} \url{https://zenodo.org/records/10255840}\\
\textbf{License:} CC BY 4.0

\vspace{1em}
\textbf{The Finnish Court Records Dataset \cite{fcr}}\\
\textbf{Description:} 600 random sample of ``The Finnish Court Records (FCR) collection encompasses 785 manuscripts from the ``Renovated District Court Records'' held by the National Archives of Fin land. Many of these manuscripts were scanned into double-page images, amount ing to 630,388 images of about one million pages in total. The manuscripts date from the 18th century and consist of records of deeds, mortgages, traditional life annuity, among others. They were written by many hands mostly in Swedish''\\
\textbf{Link:} \url{https://zenodo.org/records/4767732}\\
\textbf{License:} CC BY 4.0

\vspace{1em}
\textbf{ICDAR 2015 Competition HTRtS: Handwritten Text Recognition on the tranScriptorium Dataset \cite{icdar2015}}\\
\textbf{Description:} ``drawn from the English ``Bentham collection'' dataset used in the TRAN SCRIPTORIUM project. The selected data has been written by several hands and entails significant variabilities and difficulties regarding the quality of text images, writing styles and crossed-out text.''\\
\textbf{Link:} \url{https://zenodo.org/records/248733#.WH3zMczhBTY}\\
\textbf{License:} CC BY 4.0

\vspace{1em}
\textbf{The e-NDP project : collaborative digital edition of the Chapter registers of Notre-Dame of Paris \cite{endp}}\\
\textbf{Description:} According to our estimates no fewer than 18 main hands were involved in the writing of the registers during the medieval period. 
More than 98\% of the content of the registers was written in Latin, the rest in French. The exact percentage is hard to estimate because the vernacular language is often used in formulae, notes and comments. It is rare to find entire pages or blocks written in French. 

Script family : The registers were written using a Cursive script (ca. late XIIIe - XVIe).

Documental typology : The volumes containing the chapter conclusions were conceived to serve as memorial records, but above all as documents for regular use and consultation in the daily practice of administration and management. In diplomatics the notion of "documentary manuscripts" is used to describe this kind of sources also by opposition to books and litterary or normative manuscripts.\\
\textbf{Link:} \url{https://zenodo.org/records/7575693}\\
\textbf{License:} CC BY 4.0

\vspace{1em}
\textbf{Faithful Transcriptions Data Set: TEI/XML-encoded Transcriptions of Medieval Theological Manuscripts \cite{faithful}}\\
\textbf{Description:} ``181 pages with 8.952 text lines from 12 manuscripts in German, Dutch, and Latin. The medieval scripts include Textura, Textualis, Gothic Cursiva, and Bastarda''\\
\textbf{Link:} \url{https://zenodo.org/records/5582483}\\
\textbf{License:}  CC0 1.0

\vspace{1em}
\textbf{A Benchmark and Dataset for Post-OCR text correction in Sanskrit \cite{postocr_sanskrit}}\\
\textbf{Description:} ``a multi-domain dataset from 30 different books and have 218,000 manually verified
sentences in it''
``We consider printed versions of these books, most of them
reprinted in the first half of the twentieth century.
While, these books are widely accessible to the
public via libraries and academic institutions, we
manually had to scan several of them as part of
its digitisation process. These books vary widely
in their vocabulary and stylistic usage owing to the
differences in the domain and the original time period of publication, where the latter can be as old
as the fifth century AD.''\\
\textbf{Link:} \url{https://github.com/ayushbits/pe-ocr-sanskrit}\\
\textbf{License:} N/A

\vspace{1em}
\textbf{ICDAR2017 Competition on Handwritten Text Recognition on the READ Dataset \cite{icdar2017}}\\
\textbf{Description:} ``Most of the
dataset was taken from the Alfred Escher Letter Collection
(AEC) 5 which is written in German but it also has pages
in French and Italian. The selected dataset also included
handwritten images drawn from other German collections,
and thus characterized by being written by several hands.''\\
\textbf{Link:} \url{https://zenodo.org/records/835489}\\
\textbf{License:} CC BY 4.0

\vspace{1em}
\textbf{ICFHR2016 Competition on Handwritten Text Recognition on the READ Dataset \cite{icfhr2016}}\\
\textbf{Description:} ``a subset of documents from the Ratsprotokolle collection composed of minutes of the council meetings held from 1470 to 1805 (about 30.000 pages), which will be used in the READ project. This dataset is written in Early Modern German. The number of writers is unknown. Handwriting in this collection is complex enough to challenge the HTR software.''\\
\textbf{Link:} \url{https://zenodo.org/records/218236}\\
\textbf{License:} CC BY 4.0

\vspace{1em}
\textbf{RASAM 1 and 2 \cite{rasam1}}\\
\textbf{Description:} ``a diverse collection of Maghrebi Arabic manuscripts from the BULAC Library, featuring a wide variety of handwriting styles, layouts, states of preservation, and other characteristics representative of Arabic Maghrebi manuscript production.''\\
\textbf{Link:} \url{https://github.com/calfa-co/rasam-dataset}\\
\textbf{License:} Apache 2.0

\vspace{1em}
\textbf{Fraktur-GT \cite{fraktur}}\\
\textbf{Description:} ``Ground Truth (GT) data for Fraktur/Gothic prints from the 19th Century, released by UB, Uni-Tübingen''\\
\textbf{Link:} \url{https://github.com/ubtue/gt-fraktur}\\
\textbf{License:} CC0 1.0

\vspace{1em}
\textbf{Ground Truth transcriptions for training OCR of historical Arabic handwritten texts
- RASM2018 and RASM2019 \cite{rasm2019}}\\
\textbf{Description:} ``drawn from a selection of historical Arabic scientific manuscripts (10th-19th century) digitised through the British Library Qatar Foundation Partnership''\\
\textbf{Link:} \url{https://www.primaresearch.org/RASM2018/resources}\\
\textbf{License:} ``All ground truth resources created for RASM competitions are freely available under an open license''

\vspace{1em}
\textbf{Prima Europeana \cite{europeana}}\\
\textbf{Description:} Newspapers from 17th to 20th century\\
\textbf{Link:} \url{https://www.primaresearch.org/datasets}\\
\textbf{License:} ``Resources are available exclusively for personal research, if intended to use any software or data in the scope of commercial activities, please contact enquiries@primaresearch.org to discuss individual licensing options.''

\vspace{1em}
\textbf{Prima Impact \cite{impact}}\\
\textbf{Description:} ``texts from as early as 1500, and containing material from newspapers, books, pamphlets and typewritten notes''\\
\textbf{Link:} \url{https://www.primaresearch.org/datasets}\\
\textbf{License:} ``Resources are available exclusively for personal research, if intended to use any software or data in the scope of commercial activities, please contact enquiries@primaresearch.org to discuss individual licensing options.''

\vspace{1em}
\textbf{BLN600: A Parallel Corpus of Machine/Human Transcribed Nineteenth Century Newspaper Texts \cite{bln}}\\
\textbf{Description:} ``nineteenth-century newspaper text focused on crime in London, derived
from the Gale British Library Newspapers corpus parts 1 and 2''\\
\textbf{Link:} \url{https://orda.shef.ac.uk/articles/dataset/BLN600_A_Parallel_Corpus_of_Machine_Human_Transcribed_Nineteenth_Century_Newspaper_Texts/25439023}\\
\textbf{License:} CC BY-NC-ND 4.0

\vspace{1em}
\textbf{REID2019 ICDAR Competition on Recognition of Early Indian printed Documents \cite{reid2019}}\\
\textbf{Description:} ``printed books written in Bengali (1713-1914)''\\
\textbf{Link:} \url{https://www.primaresearch.org/datasets}\\
\textbf{License:} ``Resources are available exclusively for personal research, if intended to use any software or data in the scope of commercial activities, please contact enquiries@primaresearch.org to discuss individual licensing options.''

\vspace{1em}
\textbf{ICDAR2017 Competition on Recognition of Early Indian Printed Documents , REID2017 \cite{reid2017}}\\
\textbf{Description:} ``Books in Bengali language dating between 1785 and
1909. For the most part, the scanned images contain single
column lines of text, with a small amount containing illustrations as well as text. Some pages contain marginal data
such as numbers, handwritten notes, and decorative frames.''\\
\textbf{Link:} \url{https://www.primaresearch.org/datasets}\\
\textbf{License:} ``Resources are available exclusively for personal research, if intended to use any software or data in the scope of commercial activities, please contact enquiries@primaresearch.org to discuss individual licensing options.''

\vspace{1em}
\textbf{Gado2: multilingual newspapers from the Netherlands Indies \cite{gado}}\\
\textbf{Description:} \\
\textbf{Link:} \url{https://zenodo.org/records/4980170}\\
\textbf{License:} CC BY 4.0

\vspace{1em}
\textbf{Klosterneuburg, Stiftsbibl., Cod. 48 \cite{klostern}}\\
\textbf{Description:} ``collection of sermons of Nikolaus von Dinkelsbühl (ca. 1360 to 17th March 1433), translated and reorganised by a German redactor, from the 15th century has never been edited until now. It consists of 361 folios of parchment and paper. The text speaks about various topics such as fasting and other religious practices''\\
\textbf{Link:} \url{https://zenodo.org/records/7466928}\\
\textbf{License:} CC BY 4.0

\vspace{1em}
\textbf{The Datafication of Early Modern
Ordinances \cite{entangled}}\\
\textbf{Description:} ``All books have been published between 1532 and 1789, within
the seventeen federation-states of the Habsburg Netherlands and the Dutch Republic''\\
\textbf{Link:} \url{https://doi.org/10.5281/zenodo.3556795} \url{https://doi.org/10.5281/zenodo.3556788} \url{https://doi.org/10.5281/zenodo.3552470} \url{https://doi.org/10.5281/zenodo.3612505} \url{https://doi.org/10.5281/zenodo.3612561} \url{https://doi.org/10.5281/zenodo.3612571} \url{https://doi.org/10.5281/zenodo.3612577} \url{https://doi.org/10.5281/zenodo.3612606} \url{https://doi.org/10.5281/zenodo.3612613} \url{https://doi.org/10.5281/zenodo.3613585} \url{https://doi.org/10.5281/zenodo.3613937} \url{https://doi.org/10.5281/zenodo.3624664} \url{https://doi.org/10.5281/zenodo.3614031} \url{https://doi.org/10.5281/zenodo.3625184} \url{https://doi.org/10.5281/zenodo.3625615} \url{https://doi.org/10.5281/zenodo.3625186} \url{https://doi.org/10.5281/zenodo.3625188} \url{https://doi.org/10.5281/zenodo.3625763} \url{https://doi.org/10.5281/zenodo.3614025} \url{https://doi.org/10.5281/zenodo.3613987} \url{https://doi.org/10.5281/zenodo.3625168} \url{https://doi.org/10.5281/zenodo.3613916} \url{https://doi.org/10.5281/zenodo.3613893} \url{https://doi.org/10.5281/zenodo.3625073} \url{https://doi.org/10.5281/zenodo.3625117} \url{https://doi.org/10.5281/zenodo.3625101} \url{https://doi.org/10.5281/zenodo.3625119} \url{https://doi.org/10.5281/zenodo.3624700} \url{https://doi.org/10.5281/zenodo.3625079} \url{https://doi.org/10.5281/zenodo.3625081} \url{https://doi.org/10.5281/zenodo.3625121} \url{https://doi.org/10.5281/zenodo.3625653} \url{https://doi.org/10.5281/zenodo.3625659} \url{https://doi.org/10.5281/zenodo.3625661} \url{https://doi.org/10.5281/zenodo.3613895} \url{https://doi.org/10.5281/zenodo.3625097} \url{https://doi.org/10.5281/zenodo.3625767} \url{https://doi.org/10.5281/zenodo.3625105} \url{https://doi.org/10.5281/zenodo.3625109} \url{https://doi.org/10.5281/zenodo.3625111} \url{https://doi.org/10.5281/zenodo.3625113} \url{https://doi.org/10.5281/zenodo.3625790} \url{https://doi.org/10.5281/zenodo.3625115} \url{https://doi.org/10.5281/zenodo.3624770} \url{https://doi.org/10.5281/zenodo.3624882} \url{https://doi.org/10.5281/zenodo.3625107} \url{https://doi.org/10.5281/zenodo.3625513} \url{https://doi.org/10.5281/zenodo.3625655} \url{https://doi.org/10.5281/zenodo.3625657} \url{https://doi.org/10.5281/zenodo.3625085} \url{https://doi.org/10.5281/zenodo.3624902} \url{https://doi.org/10.5281/zenodo.3625083} \url{https://doi.org/10.5281/zenodo.3625093} \url{https://doi.org/10.5281/zenodo.3625497} \url{https://doi.org/10.5281/zenodo.3613994} \url{https://doi.org/10.5281/zenodo.3614003} \url{https://doi.org/10.5281/zenodo.3624898} \url{https://doi.org/10.5281/zenodo.3625091} \url{https://doi.org/10.5281/zenodo.3625772} \url{https://doi.org/10.5281/zenodo.3613925} \url{https://doi.org/10.5281/zenodo.3613908} \url{https://doi.org/10.5281/zenodo.3613914} \url{https://doi.org/10.5281/zenodo.3614019} \url{https://doi.org/10.5281/zenodo.3624563} \url{https://doi.org/10.5281/zenodo.3614005} \url{https://doi.org/10.5281/zenodo.3625172} \url{https://doi.org/10.5281/zenodo.3625174} \url{https://doi.org/10.5281/zenodo.3625170} \url{https://doi.org/10.5281/zenodo.3625178} \url{https://doi.org/10.5281/zenodo.3625509} \url{https://doi.org/10.5281/zenodo.3625087} \url{https://doi.org/10.5281/zenodo.3625176} \url{https://doi.org/10.5281/zenodo.3625075} \url{https://doi.org/10.5281/zenodo.3556788} \url{https://doi.org/10.5281/zenodo.3625645} \url{https://doi.org/10.5281/zenodo.3625649} \url{https://doi.org/10.5281/zenodo.3625651} \url{https://doi.org/10.5281/zenodo.3625507} \url{https://doi.org/10.5281/zenodo.3625511} \url{https://doi.org/10.5281/zenodo.3625505} \url{https://doi.org/10.5281/zenodo.3625647} \url{https://doi.org/10.5281/zenodo.3625099} \url{https://doi.org/10.5281/zenodo.3625089} \url{https://doi.org/10.5281/zenodo.3613523} \url{https://doi.org/10.5281/zenodo.3613601} \url{https://doi.org/10.5281/zenodo.3613617} \url{https://doi.org/10.5281/zenodo.3613663} \url{https://doi.org/10.5281/zenodo.3613927} \url{https://doi.org/10.5281/zenodo.3613555} \url{https://doi.org/10.5281/zenodo.3613518} \url{https://doi.org/10.5281/zenodo.3624906} \url{https://doi.org/10.5281/zenodo.3624908} \url{https://doi.org/10.5281/zenodo.3624910} \url{https://doi.org/10.5281/zenodo.3624912} \url{https://zenodo.org/record/3624914#.XiicgmhKhPY} \url{https://doi.org/10.5281/zenodo.3624916} \url{https://doi.org/10.5281/zenodo.3624918} \url{https://doi.org/10.5281/zenodo.3624920} \url{https://doi.org/10.5281/zenodo.3624922} \url{https://doi.org/10.5281/zenodo.3613553} \url{https://doi.org/10.5281/zenodo.3625103} \url{https://doi.org/10.5281/zenodo.3613974} \url{https://doi.org/10.5281/zenodo.3614027} \url{https://doi.org/10.5281/zenodo.3625182} \url{https://doi.org/10.5281/zenodo.3625501} \url{https://doi.org/10.5281/zenodo.3625503} \url{https://doi.org/10.5281/zenodo.3625190} \url{https://doi.org/10.5281/zenodo.3625643}\\
\textbf{License:} CC BY 4.0

\vspace{1em}
\textbf{NewsEye / READ OCR training dataset from Swedish Newspapers (18th, 19th, early 20th C.) \cite{newseye_swedish}}\\
\textbf{Description:} ``The dataset comprises swedish newspaper pages from late 18th till early 20th century with carefully corrected text. The page images were provided by the National Library Finland (NLF)''\\
\textbf{Link:} \url{https://zenodo.org/records/4599624}\\
\textbf{License:} CC BY 4.0

\vspace{1em}
\textbf{NewsEye / READ OCR training dataset from Austrian Newspapers (19th C.) \cite{newseye_austrian}}\\
\textbf{Description:} ``The dataset comprises Austrian newspaper pages from 19th and early 20th century with carefully corrected text. The page images were provided by the Austrian National Library``\\
\textbf{Link:} \url{https://zenodo.org/records/3387369}\\
\textbf{License:} CC BY 4.0

\vspace{1em}
\textbf{NewsEye - READ OCR training dataset from French Newspapers (18th, 19th, early 20th C)
NewsEye - READ AS training dataset from French Newspapers (19th, early 20th C) \cite{newseye_french}}\\
\textbf{Description:} ``The dataset comprises French newspaper pages from 18th, 19th and early 20th century with carefully corrected text. The page images were provided by the French National Library''\\
\textbf{Link:} \url{https://zenodo.org/records/4293602} \url{https://zenodo.org/records/5654841}\\
\textbf{License:} CC BY 4.0

\vspace{1em}
\textbf{NewsEye / READ OCR training dataset from Finnish Newspapers (18th, 19th, early 20th C.)
NewsEye / READ AS training dataset from Finnish Newspapers (19th C.) \cite{newseye_finnish}}\\
\textbf{Description:} ``The dataset comprises finnish newspaper pages from late 18th till early 20th century with carefully corrected text. The page images were provided by the National Library Finland (NLF)''\\
\textbf{Link:} \url{https://zenodo.org/records/4599472} \url{https://zenodo.org/records/5654858}\\
\textbf{License:} CC BY 4.0

\vspace{1em}
\textbf{Ground truth for Neue Zürcher Zeitung black letter period \cite{nzz}}\\
\textbf{Description:} ``The Neue Zürcher Zeitung (NZZ) has been publishing in black letter from its very first issue in 1780 until 1947. From this time period, we randomly sampled one frontpage per year, resulting in a total of 167 pages. We chose frontpages because they typically contain highly relevant material and because we want to make sure not to sample pages containing exclusively advertisements or stock information. During certain periods, the NZZ was published several times a day, and there were supplements, too. Due to incomplete metadata, the sampling included frontpages from supplements.''\\
\textbf{Link:} \url{https://github.com/impresso/NZZ-black-letter-ground-truth}\\
\textbf{License:} CC BY-NC 4.0

\vspace{1em}
\textbf{OCR17+ - Layout analysis and text recognition for 17th c. French prints \cite{ocr17plus}}\\
\textbf{Description:} ``17th c. French print''\\
\textbf{Link:} \url{https://github.com/Heresta/OCR17plus}\\
\textbf{License:} Data is CC-BY, except images which come from Gallica (cf. conditions d'utilisation).

\vspace{1em}
\textbf{The OCR-D project \cite{ocrd}}\\
\textbf{Description:} ``Union Catalogue of Books of the 16th--18th century (VD 16, VD 17, VD 18) published in the German-speaking countries''\\
\textbf{Link:} \url{https://github.com/OCR-D/gt_structure_text}\\
\textbf{License:} CC-BY-SA-4.0

\vspace{1em}
\textbf{A dataset of Spanish notarial deeds (18th Century) for Handwritten Text Recognition and Layout Analysis of historical documents \cite{notarial}}\\
\textbf{Description:} ``a subset of 596 documents from the Registre d'Hipoteques de Girona of 1769 collection, guarded by the Arxiu Històric de Girona. This collection, is composed by hundreds of thousands of notarial deeds from the XVIII-XIX century (1768-1862). Sales, redemption of censuses, inheritance and matrimonial chapters are among the most common documentary typologies in the collection.''\\
\textbf{Link:} \url{https://zenodo.org/records/1322666#.Ypi6Ty8RoUE}\\
\textbf{License:} CC BY-NC 4.0

\vspace{1em}
\textbf{Chronicling Germany \cite{chronicling_germany}}\\
\textbf{Description:} ``693 annotated historical newspaper pages from the time period between 1852 and 1924''\\
\textbf{Link:} \url{https://github.com/Digital-History-Bonn/Chronicling-Germany-Code/blob/master/script/download.py}\\
\textbf{License:} EUROPEAN UNION PUBLIC LICENCE v. 1.2

\vspace{1em}
\textbf{Manuscripts of Handwritten Arabic dataset (Muharaf) for cursive text recognition \cite{muharaf}}\\
\textbf{Description:} ``The Muharaf dataset consists of a diverse set of images, ranging from individual personal letters, poems, and dialogues to legal consensus records, correspondences, and church records. The manuscripts
date from the early 19th century to the early 21st century. The quality of page images varies, from writing on a clean white background to illegible sentences on creased pages with ink bleeds''\\
\textbf{Link:} \url{https://zenodo.org/records/11492215}\\
\textbf{License:} CC BY-NC-SA 2.0

\vspace{1em}
\textbf{A scarce dataset for ancient Arabic
handwritten text recognition \cite{scarce}}\\
\textbf{Description:} ``The books and
their authors, orderly, as appearing in the dataset, are the following: Rafaa Al Niqab An Kitab Al
Shahab, Al Husain Al Shushawi; Nuzul Al Saereen Ela Allah Rabb Al Aalamin Fee Ahadith Said Al
Mursalin, Mahmoud Al Darkazini; Kitab Al Azamah, Ibn Hayyan Al Asbahani; Al Rawd Al Nad-
heer, Muhammad Mutawalli; Sirr Al Fosoos, Muhammad Abdulbaqi; Al Juzz Alkhamis Min Jamii
Al Masanid Wa Al Sunan Al Hadi Li Aqwam Sunan, Ismail Ibn Kathir; Kitab Tareekh Madinat
Dimashq Wa Mn Banaha Min Al Mutaqaddimin, Ali Al Rabiy; Dirham Al Suarrah Fee Wad Al
Yadain Taht Al Surrah, Muhammad Hashim Al Sindi.''\\
\textbf{Link:} \url{https://data.mendeley.com/datasets/xz6f8bw3w8/1}\\
\textbf{License:} CC BY 4.0

\vspace{1em}
\textbf{Caroline Minuscule \cite{whitecarolineminuscule}}\\
\textbf{Description:} Manuscripts from 800 to 1100\\
\textbf{Link:} \url{https://github.com/rescribe/carolineminuscule-groundtruth}\\
\textbf{License:} ``The ground truth contained in this repository should be considered Public Domain, or licensed under Apache License 2.0, whichever suits your needs better.''

The metadata.txt files in each manuscript directory describes its provenance and the licensing of the images. We have only used manuscript images which are freely redistributable and reusable.

\vspace{1em}
\textbf{ \cite{safavi}}\\
\textbf{Description:} Transcription is adopted from the book ``Qāsim Beg \d{H}ayātī Tabrīzī: A Chronicle of the Early Safavids and the Reign of Shah Ismā\textsuperscript{`}īl (907--930/1501--1524): Persian Edition and Introduction'', then aligned at the page-level with the manuscript images by the authors of this paper.\\
\textbf{Link:} \\
\textbf{License:} permission granted for research use

\vspace{1em}
\textbf{Ground Truth data for printed Devanagari \cite{merkelhilf2022devanagari}}\\
\textbf{Description:} \\
\textbf{Link:} \url{https://heidata.uni-heidelberg.de/dataset.xhtml?persistentId=doi:10.11588/data/EGOKEI}\\
\textbf{License:} CC BY 4.0

\vspace{1em}
\textbf{Les Papiers Barye \cite{claass2021papiersbarye}}\\
\textbf{Description:} a collection of correspondence and documents relating to the activity of the sculptor and painter Antoine-Louis Barye (1795-1875). These documents are kept in the Antoine-Louis Barye archive of the Library of the Institut National d'Histoire de l'Art, Jacques Doucet collections. The collection is composed of 350 documents (918 pages) and organized into 6 collections:

Correspondence during the lifetime of Antoine-Louis Barye
Anonymous documentation on Antoine-Louis Barye
Documents relating to Barye's activity during his lifetime
Biographical documentation on Antoine-Louis Barye
Documents relating to the trade and exhibition of Barye's works after his death
Iconographic documentation\\
\textbf{Link:} \url{https://gitlab.inha.fr/snr/LesPapiersBarye}\\
\textbf{License:} CC-BY 4.0

\vspace{1em}
\textbf{Padeřov-Bible Handwriting Ground Truth \cite{michalcova2022paderov}}\\
\textbf{Description:} the Padeřov Bible (Vienna, Austrian National Library, shelfmark Cod. 1175, 1432--1435), the bible of the third redaction of the Old Czech Bible translation\\
\textbf{Link:} \url{https://zenodo.org/records/7467034#.Y6LQZBWZM2w}\\
\textbf{License:} CC BY 4.0

\vspace{1em}
\textbf{TRANSCRIPTIONS OF THE INTERNATIONAL CONGRESS OF ETHNOGRAPHIC SCIENCES (PARIS, 1878) \cite{christensen2022exposition1878}}\\
\textbf{Description:} The International Congress of Ethnographic Sciences of 1878 took place on the occasion of the 1878 Universal Exposition in Paris. Edited in 1881 by the National Printing Office, the proceedings of this congress have been made available by the Digital Conservatory of Arts and Crafts.
These proceedings allow us to revisit the often problematic beginnings of several scientific disciplines: ethnography, anthropology, and prehistory. They also bear witness to the development of archaeology, folklore studies, and French popular culture.\\
\textbf{Link:} \url{https://github.com/PSL-Chartes-HTR-Students/TNAH-2021-Expositions_Universelles}\\
\textbf{License:} CC-BY 4.0

\vspace{1em}
\textbf{Project for Transcribing the Active Correspondence of Hector Berlioz to His Sister Nanci Berlioz \cite{ceard2022berlioz}}\\
\textbf{Description:} ``the active correspondence of Hector Berlioz addressed to his sister Anne-Marguerite "Nanci" Berlioz''.\\
\textbf{Link:} \url{https://github.com/PSL-Chartes-HTR-Students/TNAH-2021-Projet-Correspondance-Berlioz}\\
\textbf{License:} CC-BY 4.0

\vspace{1em}
\textbf{TAPUSCORPUS \cite{chague2021tapuscorpus}}\\
\textbf{Description:} Ground Truth for French 20th century typewritten documents collected on Gallica and Europeana\\
\textbf{Link:} \url{https://github.com/HTR-United/tapuscorpus}\\
\textbf{License:} CC BY 4.0

\vspace{1em}
\textbf{TIMEUS CORPUS \cite{Chague_Time_Us_Corpus}}\\
\textbf{Description:} Registers from the Prud'hommes Court for the Textile Industry in Paris, january to june 1858
Registers from the Prud'hommes Court for the Textile Industry in Paris, january 1878
\\
\textbf{Link:} \url{https://github.com/HTR-United/timeuscorpus}\\
\textbf{License:} CC BY 4.0

\vspace{1em}
\textbf{Notaires de Paris - Bronod
 \cite{limonbonnet2024bronod}}\\
\textbf{Description:} The lectaurep-bronod corpus consists of 100 pages from the register of Maître Louis Bronod (1719--1765), a notary in Paris from December 13, 1719, to July 23, 1765. The selected pages were written between the years 1742 and 1745.\\
\textbf{Link:} \url{https://github.com/HTR-United/lectaurep-bronod}\\
\textbf{License:} CC BY 4.0

\vspace{1em}
\textbf{Lectaurep-Mariages-et-Divorces, ground truth for the Registres des Contrats de Mariages et des Séparations et Divorces (French 19th century)~\cite{rostaing2024mariages}}\\
\textbf{Description:} \\
\textbf{Link:} \url{https://github.com/HTR-United/lectaurep-mariages-et-divorces}\\
\textbf{License:} CC BY 4.0

\vspace{1em}
\textbf{Ground truth for various Parisian notary's repertoires (French 19th and 20th century)~\cite{lectaurep2021repertoires}}\\
\textbf{Description:} \\
\textbf{Link:} \url{https://github.com/HTR-United/lectaurep-repertoires}\\
\textbf{License:} CC BY 4.0

\vspace{1em}
\textbf{ \cite{alba2023htromance}}\\
\textbf{Description:} \\
\textbf{Link:} \url{https://github.com/HTRomance-Project/medieval-italian}\\
\textbf{License:} CC-BY-4.0

\vspace{1em}
\textbf{ \cite{glaise2023htromance}}\\
\textbf{Description:} \\
\textbf{Link:} \url{https://github.com/HTRomance-Project/medieval-latin}\\
\textbf{License:} CC-BY 4.0

\vspace{1em}
\textbf{ \cite{bordier2023htromance}}\\
\textbf{Description:} \\
\textbf{Link:} \url{https://github.com/HTRomance-Project/middle-ages-in-spain}\\
\textbf{License:} CC-BY 4.0

\vspace{1em}
\textbf{ \cite{norindr2023htromance}}\\
\textbf{Description:} \\
\textbf{Link:} \url{https://github.com/HTRomance-Project/modern-roman-languages}\\
\textbf{License:} CC-BY 4.0

\vspace{1em}
\textbf{Digital Peraire \cite{chague2023peraire}}\\
\textbf{Description:} The documents are handwritten, dating from the second half of the 20th century, written in French with a blue ink pen or, more frequently, with a blue pencil. Occasional marginal notes appear in red.\\
\textbf{Link:} \url{https://github.com/alix-tz/peraire-ground-truth}\\
\textbf{License:}  CC-BY 4.0

\vspace{1em}
\textbf{Ground truth for German newspaper Deutscher Reichsanzeiger und Preußischer Staatsanzeiger (1819--1945) \cite{kamlah2024reichsanzeiger}}\\
\textbf{Description:} ``Ground truth for German newspaper "Deutscher Reichsanzeiger und Preußischer Staatsanzeiger" (German Imperial Gazette and Prussian Official Gazette), which was published under changing names from 1819 to 1945''\\
\textbf{Link:} \url{https://github.com/UB-Mannheim/reichsanzeiger-gt}\\
\textbf{License:} CC0-1.0 license

\vspace{1em}
\textbf{Données du recensement du Valais \cite{dubois2024valais}}\\
\textbf{Description:} \\
\textbf{Link:} \url{https://github.com/PonteIneptique/valais-recensement}\\
\textbf{License:} CC BY-NC 4.0.

\vspace{1em}
\textbf{Ground Truth for ÖNB, Cod. 3891 \cite{ainonen2022onb3891}}\\
\textbf{Description:} \\
\textbf{Link:} \url{https://zenodo.org/record/7467249}\\
\textbf{License:} CC-BY 4.0

\vspace{1em}
\textbf{Stavronikita Monastery Greek handwritten document Collection \cite{pratikakis2021stavronikita}}\\
\textbf{Description:} It comprises manuscripts made of paper, written in the 16th century and its dimensions are 220X165 mm. The manuscript is embellished with epititles and red initials. Tachygraphical symbols and abbreviations are encountered in the manuscript as well
The collection is one of the oldest Stavronikita Monastery on Mount Athos. It is a parchment, four-gospel manuscript which has been written between 1301 and 1350. It comprises 54 pages with dimensions that are approximately 250x185 mm. The script is elegant minuscule and the use of majuscule letters is rare. Tachygraphical symbols and abbreviations are encountered in the manuscript as well. Furthermore, the manuscript is enriched with chrysography, elegant epititles and initials.
It comprises manuscripts made of paper, written at the end of the 15th century and its dimensions are 218X150 mm. In various pages, we find red initials and epititles which enrich the manuscript’s decoration. \\
\textbf{Link:} \url{https://zenodo.org/records/5578251} \url{https://zenodo.org/records/5595669} \url{https://zenodo.org/records/5578136}\\
\textbf{License:} CC-BY 4.0

\vspace{1em}
\textbf{ \cite{vlachouefstathiou2024eutyches}}\\
\textbf{Description:} Eutychès was a Latin Grammarian, active in the mid-6th century AD Constantinople, a disciple of Priscianus Caesariensis. His contribution to the field of grammar consists mainly of a treatise called De uerbo, addressed to his diligentissimum discipulorum Craterus.The ars elaborates on the criteria for the classification of the conjugation of verbs, in short, if the genitive of dico is dicas or dicis. Itself a rather arid development of principles that do not escape the general rule, nevertheless thanks to De uerbo 141 citations of classical authors have been passed down to us -some passages being otherwise lost-, mostly early Augustan poets. Adding to this, exhaustive lists of examples of verbs and derived nouns make an integral part of the treatise, that have been recycled from posterior Grammatici.

De uerbo was last edited at the end of the 19th century by Heinrich Keil in his fifth (out of seven) tome of the Grammatici Latini, alongside an enormous corpus of artes dating from the 4th to the 7th century. This monumental edition lacks methodological rigor and exhaustive paleographical research, is, for the most part, outdated, leaving a substantial amount of grammatical works in need of critical editions, including Eutychès' own contribution.

The manuscript tradition of the work, as reported by the excellent work of Colette Jeudy, is rather modest, as it is comprised of 32 manuscripts transmitting partially on entirely the text, and span from the end of the 8th to the 11th century AD. De uerbo enjoyed a fulgurant posterity, that inspired two medieval scholars, namely Sedulius Scottus and Remigius of Auxerre to make a commentary rendition out of it, attesting to its importance in the educational milieu.

The typology of the manuscripts comes with two interesting features that pose at the same time the biggest difficulty for perspective editors (and, as we suppose, for Keil). Firstly, as many other works on grammar, actively used in the "medieval classroom" for the teaching of Latin, lists of examples are incorporated in the narrative between bits of theory, forming columns or/and tabular boards, a mise-en-page apt for the memorization of these exceptions. At first, this feature may not seem as big of a deal, but a thorough examination of these tables and of the posterior works that use them indicate that it yields important information for the understanding of the manuscript tradition (see especially Conduché 2019 "La mise en page d'Eutychès"). Secondly, several witnesses are richly annotated with multiple layers of interlinear and marginal notes, the progressive accumulation of which finds its climax in Remigius of Auxerre's commentary. Gloses in grammatical manuscripts have only recently (better late than ever!) started to interest editors (see Monella/Rosselini for Priscian and Evina Steinovà for Isidore of Seville's Etymologies)most of them making use of digital tools. And that is because a typology as complex as that of grammatical glossed manuscripts cannot be easily handled and, better, be utilized for research purposes without the flexibility, absence of spatial constraints and ability of to handle of big and multilayered data that digital tools offer.\\
\textbf{Link:} \url{https://github.com/malamatenia/Eutyches}\\
\textbf{License:} Apache-2.0 license

\vspace{1em}
\textbf{ \cite{gabay2023fondue}}\\
\textbf{Description:} \\
\textbf{Link:} \url{https://github.com/FoNDUE-HTR/FONDUE-EN-PRINT-20}\\
\textbf{License:} Annotation is CC-BY. Images belong to the digital libraries.

\vspace{1em}
\textbf{ \cite{fondue-es-print-19}}\\
\textbf{Description:} \\
\textbf{Link:} \url{https://github.com/FoNDUE-HTR/FONDUE-ES-PRINT-19}\\
\textbf{License:} CC BY 4.0

\vspace{1em}
\textbf{ \cite{fondue-fr-mss-18}}\\
\textbf{Description:} \\
\textbf{Link:} \url{https://github.com/FoNDUE-HTR/FONDUE-FR-MSS-18}\\
\textbf{License:} CC BY 4.0

\vspace{1em}
\textbf{ \cite{fondue-fr-print-16}}\\
\textbf{Description:} \\
\textbf{Link:} \url{https://github.com/FoNDUE-HTR/FONDUE-FR-PRINT-16}\\
\textbf{License:} CC BY 4.0

\vspace{1em}
\textbf{ \cite{fondue-fr-print-20}}\\
\textbf{Description:} \\
\textbf{Link:} \url{https://github.com/FoNDUE-HTR/FONDUE-FR-PRINT-20}\\
\textbf{License:} CC BY 4.0

\vspace{1em}
\textbf{ \cite{fondue-it-print-20}}\\
\textbf{Description:} \\
\textbf{Link:} \url{https://github.com/FoNDUE-HTR/FONDUE-IT-PRINT-20}\\
\textbf{License:} CC BY 4.0

\vspace{1em}
\textbf{Swiss (German and French) art catalogues of the 19th c. \cite{joyeuxprunel2023fondue}}\\
\textbf{Description:} Catalogues from Swiss art exhibitions from the Turnus period (1842-1961) published by the Société suisse des beaux-arts (SSBA). More information on the website of the SIK-ISEA.\\
\textbf{Link:} \url{https://github.com/FoNDUE-HTR/FONDUE-MLT-ART}\\
\textbf{License:} The catalogues are in the public domain, images are made available by SIK-ISEA and transcriptions are CC-BY.

\vspace{1em}
\textbf{Datasets for catalogs OCR and segmentation \cite{pradier2022fonduecat}}\\
\textbf{Description:} transcription of catalogues, printed mainly in the 19th c. but not only\\
\textbf{Link:} \url{https://github.com/FoNDUE-HTR/FONDUE-MLT-CAT}\\
\textbf{License:} CC BY 4.0

\vspace{1em}
\textbf{HTR data sets from medieval manuscripts (13th-14th c.) collecting "fabliaux" \cite{pinche2023fabliaux}}\\
\textbf{Description:} medieval manuscripts (13th-14th c.) collecting "fabliaux" in Old French\\
\textbf{Link:} \url{https://github.com/CIHAM-HTR/Fabliaux}\\
\textbf{License:} CC BY 4.0

\vspace{1em}
\textbf{ \cite{carta2022fondue}}\\
\textbf{Description:} 374 Spanish chapbooks published by the same printer (José María Moreno) during the 19th century.
5 pliegos from the Varios corpus (32 pages)\\
\textbf{Link:} \url{https://github.com/DesenrollandoElCordel/FoNDUE-Spanish-chapbooks-Dataset} \url{https://github.com/DesenrollandoElCordel/Moreno-OCR-files}\\
\textbf{License:} CC0-1.0 license

\vspace{1em}
\textbf{Ground Truth data for printed Malayalam \cite{tuebingen2023malayalam}}\\
\textbf{Description:} \\
\textbf{Link:} \url{https://heidata.uni-heidelberg.de/dataset.xhtml?persistentId=doi:10.11588/data/L2KRZO}\\
\textbf{License:} CC BY 4.0

\vspace{1em}
\textbf{Ground-Truthed Data Set of Zenon Papyri for Handwritten Text Recognition \cite{marthotsantaniello2022zenon}}\\
\textbf{Description:} \\
\textbf{Link:} \url{https://zenodo.org/records/6565706}\\
\textbf{License:} CC BY 4.0

\vspace{1em}
\textbf{HN2021-ChateauChavigny \cite{pascual2022chavigny}}\\
\textbf{Description:}  Testimony of the looting of the Château de Chavigny in Lerné in Touraine in 1468 by the Protestants\\
\textbf{Link:} \url{https://github.com/PSL-Chartes-HTR-Students/HN2021-ChateauChavigny}\\
\textbf{License:} CC BY 4.0

\vspace{1em}
\textbf{HN2021-Kovalewsky-1893 \cite{leveque2022kovaleswky}}\\
\textbf{Description:} Transcription of the chapter "On the regime of property among the Ossetians", in: Kovalewsky M. Contemporary custom and ancient law: Ossetian customary law, enlightened by comparative history. Paris, 1893\\
\textbf{Link:} \url{https://github.com/PSL-Chartes-HTR-Students/HN2021-Kovalewsky-1893}\\
\textbf{License:} CC BY 4.0

\vspace{1em}
\textbf{Memorials for Jane Lathrop Stanford \cite{guimaraes2022memorials}}\\
\textbf{Description:} \\
\textbf{Link:} \url{https://github.com/PSL-Chartes-HTR-Students/HN2021-Memorials_Jane_Lathrop_Stanford}\\
\textbf{License:} CC BY 4.0

\vspace{1em}
\textbf{HN2021-OCR-Poesie-Corse \cite{sarbachpulicani2022corse}}\\
\textbf{Description:} the collection of poems "Pontenôvu", written by Petru Rocca and published by the "Stamparia di a Muvra" in 1927 in the name of the Corsican Action Party.\\
\textbf{Link:} \url{https://github.com/PSL-Chartes-HTR-Students/HN2021-OCR-Poesie-Corse}\\
\textbf{License:} CC BY 4.0

\vspace{1em}
\textbf{Handwritten Paleographic Greek Text Recognition: A Century-Based
Approach \cite{platanou2022paleogreek}}\\
\textbf{Description:} The Barocci collection
consists of 244 volumes and it is the largest acquisition of the Bodleian collection. The dates of these
manuscripts range from the 8th century AD to the 17th
century AD.
folios serve as characteristic examples
of the writing style they represent. They include both
Greek minuscule script and the cursive style of the minuscule script. In this way, our work involves examination of different styles and is not limited to one style of
script.\\
\textbf{Link:} \url{https://github.com/vivianpl/hpgtr}\\
\textbf{License:} CC BY-NC-SA 3.0

\vspace{1em}
\textbf{15th century manuscript HTR data \cite{pinche2022htr15c}}\\
\textbf{Description:} French manuscripts from the 15th century.\\
\textbf{Link:} \url{https://github.com/Gallicorpora/HTR-MSS-15e-Siecle}\\
\textbf{License:} CC0-1.0 license

\vspace{1em}
\textbf{ \cite{humbel2024sloanelab}}\\
\textbf{Description:} handwriting of Hans Sloane (1660-1753)
Sloane’s Catalogue of Miscallanies (folio 3-152, recto and verso). The Catalogue of Miscallanies was chosen for the training data creation because it is known to be predominantly written by Sloane.\\
\textbf{Link:} \url{https://github.com/sloanelab-org/HTR-Model}\\
\textbf{License:} CC BY-NC-SA 4.0

\vspace{1em}
\textbf{ The SETAF project \cite{solfrini2023setaf}}\\
\textbf{Description:} works published by Jean Michel, master printer in Geneva from 1538 to 1544, who bought Pierre de Vingle's typographical equipment. The texts are sixteenth-century French prints in Gothic characters and the list of texts with more details can be found in the CSV table of the repository.

This repository contains OCR data from works published by Pierre de Vingle, master printer in Lyon from 1525 to 1532, in Geneva in 1532-1533 and in Neuchâtel in 1533-1535. The texts are sixteenth-century French prints in Gothic characters and the list of texts with more details can be found in the CSV table of the repository.

This repository contains OCR data from various editions of the Facts of Jesus Christ and the Pope. It is a singular work of religious polemic, the only illustrated book of the French-speaking Reformation preserved for the first half of the sixteenth century. It was published anonymously, in three successive editions, in [Neuchâtel], [Geneva] and [Lyon], between the years 1530 and 1560.
\\
\textbf{Link:} \url{https://github.com/SETAFDH/HTR-SETAF-Jean-Michel} \url{https://github.com/SETAFDH/HTR-SETAF-Pierre-de-Vingle} \url{https://github.com/SETAFDH/HTR-SETAF-LesFaictzJCH}\\
\textbf{License:}  CC BY 4.0. 

\vspace{1em}
\textbf{ \cite{gabay2022htr16e}}\\
\textbf{Description:} \\
\textbf{Link:} \url{https://github.com/Gallicorpora/HTR-imprime-16e-siecle}\\
\textbf{License:} CC0-1.0 license

\vspace{1em}
\textbf{ \cite{imprime-17e-siecle}}\\
\textbf{Description:} \\
\textbf{Link:} \url{https://github.com/Gallicorpora/HTR-imprime-17e-siecle}\\
\textbf{License:}  CC-BY 4.

\vspace{1em}
\textbf{ \cite{imprime-18e-siecle}}\\
\textbf{Description:} \\
\textbf{Link:} \url{https://github.com/Gallicorpora/HTR-imprime-18e-siecle}\\
\textbf{License:}  CC-BY 4.

\vspace{1em}
\textbf{ \cite{htr-incunable-15e}}\\
\textbf{Description:} \\
\textbf{Link:} \url{https://github.com/Gallicorpora/HTR-incunable-15e-siecle}\\
\textbf{License:}  CC-BY 4.

\vspace{1em}
\textbf{JosephHookerHTR \cite{schaefer2023hooker}}\\
\textbf{Description:} the correspondence of Joseph Dalton Hooker (1817-1911), primarily letters to William Turner Thiselton-Dyer (1843-1928) during the late-19th/early-20th century. Many transcriptions in this dataset were generated by a small team of anonymous volunteers as part of the Joseph Hooker Correspondence Project based at Kew Gardens. \\
\textbf{Link:} \url{https://github.com/jschaefer738b/JosephHookerHTR.git}\\
\textbf{License:} CC BY 4.0

\vspace{1em}
\textbf{LiDi 1.0 Project \cite{agostini2024lidi}}\\
\textbf{Description:} 16th antiquarian Pirro Ligorio
time: notBefore: '1568' notAfter: '1580'

hands: count: '1' precision: estimated\\
\textbf{Link:} \url{https://github.com/Giorgiaagostini/LiDi1.0-project}\\
\textbf{License:} CC-BY-SA 4.0

\vspace{1em}
\textbf{Liber \cite{aruta2023liber}}\\
\textbf{Description:} \\
\textbf{Link:} \url{https://github.com/CIHAM-HTR/Liber}\\
\textbf{License:} CC-BY 4.0

\vspace{1em}
\textbf{Dresdner Hofdiarium 1665 (Mscr.Dresd.K.80) - 17th century Kurrent manuscript \cite{beckert2024hofdiarium}}\\
\textbf{Description:} \\
\textbf{Link:} \url{https://zenodo.org/records/14356190}\\
\textbf{License:} CC BY-NC-SA 4.0

\vspace{1em}
\textbf{NuBIS OCR \cite{nubis2024ocr}}\\
\textbf{Description:} a sample of 3 pages taken from 19 printed books, making a total of 57 pages. These books, in French and Latin, cover a period from 1602 to 1989, with roughly one book every 20 years. They are thus representative of the printed documents available in the digital library.\\
\textbf{Link:} \url{https://github.com/ksefil/NuBIS-OCR}\\
\textbf{License:} CC-BY 4.0

\vspace{1em}
\textbf{TNAH-2021-ArgusDesBrevets \cite{decraene2022argus}}\\
\textbf{Description:} The 1910 patent argus is presented in the form of a contemporary print, organized into sections grouping chronologically and then thematically the patents filed in France. This list and brief presentation of the patents is divided into two columns and presents standardized abbreviations.\\
\textbf{Link:} \url{https://github.com/PSL-Chartes-HTR-Students/TNAH-2021-ArgusDesBrevets}\\
\textbf{License:} CC-BY 4.0

\vspace{1em}
\textbf{TNAH-2021-DecameronFR \cite{biay2022decameronfr}}\\
\textbf{Description:} The project aims to create ground truth data for training HTR (Handwritten Text Recognition) models based on a French manuscript from the years 1430--1455: manuscript 5070 from the Bibliothèque de l'Arsenal (available on Gallica). This manuscript contains the French translation of Boccaccio's Decameron by Laurent de Premierfait. Our ground truth data covers the description of the plague in Florence, located in the prologue of the work.\\
\textbf{Link:} \url{https://github.com/PSL-Chartes-HTR-Students/TNAH-2021-DecameronFR}\\
\textbf{License:} CC-BY 4.0

\vspace{1em}
\textbf{Notre-Dame Project \cite{doat2022notredame}}\\
\textbf{Description:} the transcription of the daily journals from the year 1860, documenting the restoration work carried out between 1844 and 1865 at the Notre-Dame Cathedral in Paris under the direction of Eugène Viollet-le-Duc and Jean-Baptiste Lassus. \\
\textbf{Link:} \url{https://github.com/PSL-Chartes-HTR-Students/TNAH-2021-Projet-Notre-Dame}\\
\textbf{License:} CC-BY 4.0

\vspace{1em}
\textbf{TranscriboQuest 2024 Medieval Literary \cite{vandyck2024transcriboquest}}\\
\textbf{Description:}  The aim of this dataset was to contribute to underrepresented aspects of the manuscripts used in the CATMuS project. We opted to focus on medieval scientific documents that are damaged, in several different languages. \\
\textbf{Link:} \url{https://zenodo.org/records/13757440}\\
\textbf{License:} CC BY 4.0

\vspace{1em}
\textbf{Verard Corpus \cite{hoeben2024verard}}\\
\textbf{Description:} Parts (10 pages) of Vérard's Editions princeps of the following texts:

Premier Volume de Tristan (1489) shelfmark: IFN-8600174, folios 12r-16v
Second Volume de Tristan (1489) shelfmark: IFN-8600175, folios 12r-16v
Premier Volume de Merlin [1498/1503] shelfmark: Res Y2 26, folios 12v-17r
Second Volume de Merlin [1498/1503] shelfmark: Res Y2 26, folios 12v-17r
Prophecies de Merlin [1498/1503] shelfmark: Res y2 27, folios 12r-16v
Gyron le Gourtoys [1501/1503] shelfmark: RES-Fol-Bl-922, folios 12r-16v\\
\textbf{Link:} \url{https://github.com/LaurieHoeben/Verard-corpus}\\
\textbf{License:} Etalab Open License 2.0

\vspace{1em}
\textbf{EPARCHOS - Historical Greek handwritten document dataset \cite{papazoglou2020eparchos}}\\
\textbf{Description:} The dataset originates from a Greek handwritten codex that dates from around 1500-1530. This is the subset of the codex British Museum Addit. 6791, written by two hands, one by Antonius Eparchos and the other by Camillos Zanettus (ff. 104r-174v) and delivers texts by Hierocles (In Aureum carmen), Matthaeus Blastares (Collectio alphabetica) and, notably, texts by Michael Psellos (De omnifaria doctrina). The writing delivers the most important abbreviations, logograms and conjunctions, which are cited in virtually every Greek minuscule handwritten codex from the years of the manuscript transliteration and the prevalence of the minuscule script (9th century) to the post-Byzantine years.\\
\textbf{Link:} \url{https://zenodo.org/records/4095301}\\
\textbf{License:} CC BY 4.0

\vspace{1em}
\textbf{Episearch HTR \cite{tommasi2024episearch}}\\
\textbf{Description:} ``Diplomatic trascription of Giovanni Antonio Astori’s letters to Ludovico Antonio Muratori by Tatiana Tommasi;
images of Astori’s letters to Muratori (from Internet Culturale)''\\
\textbf{Link:} \url{https://github.com/vedph/episearch-htr}\\
\textbf{License:} CC BY-SA 4.0

\vspace{1em}
\textbf{Ground Truth Dataset Medieval Greek Manuscripts -- Codex heidelbergensis palatinus graecus 23 \cite{htr_united_https-gitlabhuma-numfr-ecrinum-anthologia-htr-cpgr23}}\\
\textbf{Description:} \\
\textbf{Link:} \url{https://gitlab.huma-num.fr/ecrinum/anthologia/htr_cpgr23}\\
\textbf{License:} CC-BY 4.0

\vspace{1em}
\textbf{The project iForal: Portuguese municipal charters in the Middle Ages: an historical and linguistic approach in the digital era \cite{htr_united_https-githubcom-arch-w-iforal-dataset}}\\
\textbf{Description:} \\
\textbf{Link:} \url{https://github.com/Arch-W/iForal-Dataset}\\
\textbf{License:} CC-BY 4.0

\vspace{1em}
\textbf{BiblIA \cite{stokl_ben_ezra_2021_5167263}}\\
\textbf{Description:} ``Medieval Hebrew manuscripts from the Bibliothèque nationale de France (BnF, National Library of France) and the Biblioteca Apostolica Vaticana (BAV, Vatican Library)''\\
\textbf{Link:} \url{https://zenodo.org/records/5167263}\\
\textbf{License:}  CC-BY-NC-SA 4.0

\vspace{1em}
\textbf{Transcription corpora for training HTR models for medieval manuscripts from the 12th to the 15th century. \cite{cremma-medieval}}\\
\textbf{Description:} \\
\textbf{Link:} \url{https://github.com/HTR-United/cremma-medieval}\\
\textbf{License:} CC-BY 4.0

\vspace{1em}
\textbf{CREMMA HTR GT for medieval latin manuscripts \cite{cremma-medieval-lat}}\\
\textbf{Description:} \\
\textbf{Link:} \url{https://github.com/HTR-United/CREMMA-Medieval-LAT}\\
\textbf{License:}  CC-BY 4.0

\vspace{1em}
\textbf{CREMMA - A repository of 20th century manuscripts \cite{cremma-mss20}}\\
\textbf{Description:} \\
\textbf{Link:} \url{https://github.com/HTR-United/CREMMA-MSS-20}\\
\textbf{License:}  CC-BY 4.0

\vspace{1em}
\textbf{CREMMA - A repository of 18th century manuscripts \cite{cremma-mss18}}\\
\textbf{Description:} \\
\textbf{Link:} \url{https://github.com/HTR-United/CREMMA-MSS-18}\\
\textbf{License:}  CC-BY 4.0

\vspace{1em}
\textbf{CREMMA - A repository of 19th century manuscripts \cite{cremma-mss19}}\\
\textbf{Description:} \\
\textbf{Link:} \url{https://github.com/HTR-United/CREMMA-MSS-19}\\
\textbf{License:} CC-BY 4.0

\vspace{1em}
\textbf{CREMMA - A repository of 16th century manuscripts \cite{cremma-mss16}}\\
\textbf{Description:} \\
\textbf{Link:} \url{https://github.com/HTR-United/CREMMA-MSS-16}\\
\textbf{License:}  CC-BY 4.0

\vspace{1em}
\textbf{CREMMA - A repository of 17th century manuscripts \cite{cremma-mss17}}\\
\textbf{Description:} \\
\textbf{Link:} \url{https://github.com/HTR-United/CREMMA-MSS-17}\\
\textbf{License:}  CC-BY 4.0

\vspace{1em}
\textbf{CREMMA Testament De Poilus \cite{cremma-testament}}\\
\textbf{Description:} Each document (1 or more images sharing on the same side) corresponds to a hand. As these are the wills of soldiers who died for France during the First World War, all the documents were written between 1898 and 1918.\\
\textbf{Link:} \url{https://github.com/HTR-United/CREMMA-AN-TestamentsDePoilus}\\
\textbf{License:}  CC-BY 4.0

\vspace{1em}
\textbf{CREMMA Early Modern Books \cite{cremma-16-17-print}}\\
\textbf{Description:} \\
\textbf{Link:} \url{https://github.com/HTR-United/cremma-16-17-print}\\
\textbf{License:}  CC-BY 4.0

\vspace{1em}
\textbf{DAHN Corpus \cite{dahn}}\\
\textbf{Description:} Ground Truth dataset for French 20th typewritten OCR\\
\textbf{Link:} \url{https://github.com/HTR-United/dahncorpus}\\
\textbf{License:}  CC-BY 4.0

\vspace{1em}
\textbf{Celestine Doniau-Danest \cite{celestine}}\\
\textbf{Description:} The dataset contains pages randomly selected from the digitization of the "Journal de Célestine Doniau-Danest sur les débuts de la Guerre 1914-1918" (Diary of Célestine Doniau-Danest on the beginning of the 1914--1918 war)
The dataset features only one handwriting style, with little variation, for a text written between 1914 and 1915. The digitizations are double-page spreads.\\
\textbf{Link:} \url{https://github.com/alix-tz/dataset-celestine-doniau-danest}\\
\textbf{License:} CC-BY 4.0

\vspace{1em}
\textbf{Assisi, Fondo Antico presso la Biblioteca del Sacro Convento, 408 \cite{assisi_fondo_antico_sacro_convento}}\\
\textbf{Description:} \\
\textbf{Link:} \url{https://www.internetculturale.it/it/1175/assisi-fondo-antico-del-sacro-convento-mediatheca-franciscana}\\
\textbf{License:} CC BY-NC-SA 4.0

\vspace{1em}
\textbf{La Correspondance Jacques Doucet - René Jean \cite{doucet}}\\
\textbf{Description:} This collection, donated by René-Jean in 1946, is one of the main sources on the relationship between Doucet and the man he hired as librarian on 2 June 1908. The letters and documents that make up the book evoke various episodes relating to the formation and development of the Art and Archaeology Library; they also provide more general information on the life, functioning and ideas of Jacques Doucet.\\
\textbf{Link:} \url{https://gitlab.inha.fr/snr/LaCorrespondanceDoucetReneJean}\\
\textbf{License:} Etalab Open License 2.0

\vspace{1em}
\textbf{The school of Salamanca \cite{salamanca}}\\
\textbf{Description:} ``The digital collection of sources of the project "The School of Salamanca" comprises 107 works in total,''\\
\textbf{Link:} \url{https://www.salamanca.school/works.html}\\
\textbf{License:} CC BY 4.0

\vspace{1em}
\textbf{A Human-Annotated Dataset of Scanned Images and OCR Texts from Medieval Documents \cite{ahisto}}\\
\textbf{Description:} This is an open dataset of scanned images and OCR texts from 19th and 20th century letterpress reprints of documents from the Hussite era.\\
\textbf{Link:} \url{https://nlp.fi.muni.cz/trac/ahisto/wiki/OcrDataset}\\
\textbf{License:} CC0 1.0

\vspace{1em}
\textbf{Old Books Dataset \cite{books}}\\
\textbf{Description:} ``built with Project Gutenberg ebooks. They were selected among the following books:

-Betrayed Armenia, de Diana Agabeg Apcar
-The Boy Apprenticed to an Enchanter, de Padraic Colum
-The Child of the Moat, de Stoughton Holborn
-The Corset and the Crinoline, de W.B.L
-Engraving of Lions, Tigers, Panthers, Leopards, Dogs, \&C., de Thomas Landseer
-Half-Hours with Highwaymen, de Charles G. Harper
-Historical Sketches of Colonial Florida, de Richard L. Campbell
-Horton Genealogy, de Geo. F. Horton
-The Lusitania's Last Voyage, de Charles E. Lauriat
-Seat Weaving, de L. Day Perry''\\
\textbf{Link:} \url{https://github.com/PedroBarcha/old-books-dataset}\\
\textbf{License:} Project Gutenberg License for annotations

\vspace{1em}
\textbf{An Historical Handwritten Arabic Dataset for Segmentation-Free Word Spotting - HADARA80P
 \cite{hadara}}\\
\textbf{Description:} \\
\textbf{Link:} \url{https://cloud.tu-braunschweig.de/s/anM8gnrLC66aTAf}\\
\textbf{License:} permission granted for research use

\vspace{1em}
\textbf{SleukRith Set \cite{sleuk}}\\
\textbf{Description:}  the first dataset specifically created for Khmer palm leaf manuscripts
Unlike the writing of most Latin languages where
characters are sequenced from left to right, in Khmer writing,
vowels are positioned either on the left, on the right, below, or
above the consonant they are spelled with. Two or more
consonants can also be merged together by transforming into
different shapes called low-consonant or subscript form which
are placed below the main consonant. These variations of how
Khmer letters are formed produce multiple levels (sometimes
more than three) of characters. Moreover, some writers tend to
exaggerate their writing by elongating the upper or lower part of
a character which makes it go far out of its main line, touch, or
overlap with other characters from adjacent lines.
In addition, the ambiguity of certain characters in Khmer
alphabet is a big challenge as well for character recognition
problem. Some groups of characters can only be distinguishable
by a mere difference of a small hole or a short stroke. Some
types of symbols contain multiple parts whose shapes are
identical or very similar to other characters.\\
\textbf{Link:} \url{https://github.com/donavaly/SleukRith-Set}\\
\textbf{License:} permission granted for research use

\vspace{1em}
\textbf{annotationdb \cite{annotationdb}}\\
\textbf{Description:} \\
\textbf{Link:} \url{https://tc11.cvc.uab.es/datasets/AnnotationDB_1}\\
\textbf{License:} permission granted for research use

\vspace{1em}
\textbf{Ground Truth Model for Pracalit for Sanskrit and Newar MSS 16th to 19th C.
 \cite{pracalit}}\\
\textbf{Description:} \\
\textbf{Link:} \url{https://heidata.uni-heidelberg.de/dataset.xhtml?persistentId=doi:10.11588/data/WI9184}\\
\textbf{License:} CC BY 4.0

\vspace{1em}
\textbf{ICDAR 2019 Historical Document Reading Challenge on Large Structured Chinese Family Records (ICDAR2019HDRC) \cite{icdar}}\\
\textbf{Description:} mainly written in Chinese traditional Han script. The document images have been taken from different books.\\
\textbf{Link:} \url{https://tc11.cvc.uab.es/datasets/ICDAR2019HDRC_1} \url{https://tc11.cvc.uab.es/datasets/ICDAR2019HDRC_1/gt_1_1}\\
\textbf{License:} CC BY-NC-SA 3.0

\vspace{1em}
\textbf{ \cite{ocr-data}}\\
\textbf{Description:} historical prints from around 1830\\
\textbf{Link:} \url{https://github.com/millawell/ocr-data}\\
\textbf{License:} CC BY 4.0

\vspace{1em}
\textbf{TariMa dataset \cite{tarima}}\\
\textbf{Description:} This new dataset is focused on specific contents or vocabulary lacking in previous RASAM models.\\
\textbf{Link:} \url{https://github.com/calfa-co/tarima}\\
\textbf{License:} Apache-2.0 license

\vspace{1em}
\textbf{ \cite{finlam}}\\
\textbf{Description:} The Finlam dataset includes 149 French newspapers from the 19th to 20th centuries.\\
\textbf{Link:} \url{https://huggingface.co/datasets/Teklia/Newspapers-finlam}\\
\textbf{License:} MIT License

\vspace{1em}
\textbf{ \cite{laliberte}}\\
\textbf{Description:} The Finlam La Liberté dataset includes 1500 issues from La Liberté, a French newspaper, from 1925 to 1928.\\
\textbf{Link:} \url{https://huggingface.co/datasets/Teklia/Newspapers-finlam-La-Liberte}\\
\textbf{License:} MIT License

\vspace{1em}
\textbf{The Osman Agha dataset \cite{osmanagha}}\\
\textbf{Description:} The Osman Agha dataset consists of 243 images of an 18th-century Ottoman manuscript and its Arabic-script transcription, aligned at the page level by the authors of this paper. The manuscript contains the memoirs of Osman Agha, who was a prisoner of war in Austria between 1688 and 1699. The memoir was completed on May 18, 1724. The manuscript survives as a single authorial copy, currently held at the British Library (MS. Or. 3213). The manuscript is commonly known in English as Prisoner of the Infidels: The Memoirs of Osman Agha of Timişoara, the title of its 2021 English translation (Casale 2021). A transcription of the manuscript was published in 1980 (Kreutel, 1980).\\
\textbf{Link:} \\
\textbf{License:} permission granted for research use

\vspace{1em}
\textbf{American Stories~\cite{stories}}\\
\textbf{Description:} ``newspaper scans from Library of Congress's Chronicling America collection''\\
\textbf{Link:} \url{https://huggingface.co/datasets/dell-research-harvard/AmericanStoriesTraining/blob/main/gold_data/gold_hand_transcription.json}\\
\textbf{License:} Apache license 2.0

\end{document}